%% file: main.tex
\documentclass[twoside,11pt]{article}
\usepackage[preprint]{jmlr2e}
\usepackage{amsmath}
\usepackage{booktabs}
\usepackage{array}
\usepackage{algorithm}
\usepackage{algpseudocode}
\usepackage{tikz}
\usetikzlibrary{arrows.meta,positioning,fit}
\usepackage{lastpage}
\usepackage{xcolor}

\graphicspath{{figures/}}
\input{numbers.tex}

\jmlrheading{}{2026}{1-\pageref{LastPage}}{}{}{}{Sai Praveen Chettupalli and Narahara C. Dingari}
\ShortHeadings{Anytime-Valid Skill Admission}{Chettupalli and Dingari}
\firstpageno{1}

\begin{document}

\title{Certifying That an Agent Skill Helps: Anytime-Valid, Resubmission-Proof and Verifiable Admission for Skill Registries}

\author{\name Sai Praveen Chettupalli \email saichettupalli@gmail.com\\
       \addr Independent researcher
       \AND
       \name Narahara C. Dingari \email ndingari@wpi.edu\\
       \addr Department of Data Science\\
       Worcester Polytechnic Institute\\
       100 Institute Road, Worcester, MA 01609, USA\\
       ORCID 0009-0006-9351-3375. Corresponding author}

\editor{}

\maketitle

\begin{abstract}%
\input{\jmlrdir sections/abstract_jmlr.tex}
\end{abstract}


\input{sections/intro}
\input{sections/related}
\input{sections/setup}
\input{sections/gate}
\input{sections/theory}
\input{sections/evaluation}
\input{sections/results}
\input{sections/discussion}

\section*{Broader impact}
\input{sections/impact}

\section*{Code and data availability}
Code (the \texttt{nixigate} package, all experiment scripts, and the scripts that generate every number and figure in this paper) are public at \url{https://github.com/ncdingari/skill-admission-gate}, with the ledger, transcripts and a verifier for every live study; the raw SkillsBench job folders of live study~1 are in a separate public data archive linked from the repository's \texttt{DATA.md}. The benchmark data reanalyzed here are public: SWE-Skills-Bench per-cell results (MIT license) and the SkillsBench results registry (AGPL-3.0), at the commits listed in Appendix~\ref{app:data}.

\acks{This work received no funding. The authors declare no competing interests.}

\appendix
\input{sections/appendix_notation}
\input{sections/appendix_related}
\input{sections/appendix_results}
\input{sections/appendix_theory}
\input{sections/appendix}

\vskip 0.2in
\bibliography{refs}

\end{document}

%% file: numbers.tex
\newcommand{\PAlpha}{0.05}
\newcommand{\PAlphaF}{0.10}
\newcommand{\PNmax}{600}
\newcommand{\PCap}{0.50}
\newcommand{\PCapAlt}{0.75}
\newcommand{\PLiftOne}{10}
\newcommand{\PFutM}{5}
\newcommand{\PFutMFrac}{0.05}
\newcommand{\PBase}{33.9}
\newcommand{\PCurated}{16.6}
\newcommand{\PSdTheta}{1.5}
\newcommand{\PSdLift}{0.5}
\newcommand{\PRho}{0.5}
\newcommand{\PWindow}{20}
\newcommand{\PWindows}{10}
\newcommand{\PNullShare}{80}
\newcommand{\PResub}{20}
\newcommand{\PReps}{20{,}000}
\newcommand{\PRepsPower}{4{,}000}
\newcommand{\PDrift}{0.5}
\newcommand{\PGenSpace}{200}
\newcommand{\PCoverMax}{1}
\newcommand{\PTranscriptKB}{40}
\newcommand{\NEOneFixed}{4.9}

\newcommand{\NEOneValidMax}{5.1}
\newcommand{\NEOneValidMaxSE}{0.15}
\newcommand{\NEOnePeekTMin}{20.0}
\newcommand{\NEOnePeekTMax}{34.5}
\newcommand{\NEOnePeekCIMin}{9.7}
\newcommand{\NEOnePeekCIMax}{19.9}
\newcommand{\NEOneStaleMin}{65}
\newcommand{\NEOneStaleMax}{99}
\newcommand{\NEOneStaleBet}{66}
\newcommand{\NFixedN}{137}
\newcommand{\NEOnePeekRatio}{3.6}
\newcommand{\NEEightNullPeekT}{30.3}
\newcommand{\NEEightNullPeekCI}{15.4}

\newcommand{\NSweTies}{69}
\newcommand{\NSweSD}{0.20}
\newcommand{\NETwoPowFixed}{81}
\newcommand{\NETwoPowBet}{99}

\newcommand{\NETwoNBetLOne}{174}

\newcommand{\NETwoNBetCur}{68}

\newcommand{\NETwoSaveCur}{50}
\newcommand{\NETwoNBetZero}{570}
\newcommand{\NETwoNBetFutZero}{362}
\newcommand{\NETwoNBetFutNeg}{139}
\newcommand{\NETwoNSPRTZero}{77}
\newcommand{\NETwoPowBetHalf}{47}

\newcommand{\NETwoPowBetThree}{81}
\newcommand{\NETwoRunsBetThree}{311}
\newcommand{\NETwoRunsFixed}{274}

\newcommand{\NETwoPowBetHalfPrpl}{55}
\newcommand{\NETwoNullBet}{2.8}
\newcommand{\NETwoNullBetPrpl}{1.7}
\newcommand{\NEThreeFdrPerExam}{10.1}

\newcommand{\NEThreeNStopm}{345}

\newcommand{\NEThreeNRunMax}{421}

\newcommand{\NEThreePowELond}{82}

\newcommand{\NEThreeFdrELondH}{1.0}
\newcommand{\NEThreePowELondH}{88}

\newcommand{\NResubPerExam}{42.4}
\newcommand{\NResubFivePerExam}{12.7}
\newcommand{\NResubImpliedSize}{2.7}
\newcommand{\NResubPeekCI}{98.1}
\newcommand{\NResubFivePeekCI}{62.5}
\newcommand{\NResubLineage}{2.4}

\newcommand{\NEFourRelBiasFive}{67}
\newcommand{\NEFourRelBiasTen}{19}
\newcommand{\NEFourBiasFixedFive}{79}
\newcommand{\NEFourCoverMin}{98.1}
\newcommand{\NEFourLcbMax}{1.0}
\newcommand{\NEFiveSweFalseFixed}{52}

\newcommand{\NEFiveSweFalsePeekCI}{75}

\newcommand{\NEFiveSweFalseBet}{35}
\newcommand{\NEFiveSweRunsBet}{12{,}956}
\newcommand{\NEFiveSweFalseELond}{0}
\newcommand{\NEFiveSweRunsELond}{23{,}702}

\newcommand{\NEsixFPLiteral}{0}

\newcommand{\NEsixCleanCover}{0.018}
\newcommand{\NESevenN}{600}
\newcommand{\NESevenVerify}{0.05}
\newcommand{\NESevenProctor}{0.32}
\newcommand{\NESevenMB}{48}
\newcommand{\NESevenLedger}{936}
\newcommand{\NSweCells}{392}
\newcommand{\NSweGolden}{5}
\newcommand{\NSweGoldenSkills}{4}
\newcommand{\NSweHarmful}{1}
\newcommand{\NSweOneSided}{7}
\newcommand{\NSweBH}{0}

\newcommand{\NSweRepLift}{7.9}
\newcommand{\NSweRepPos}{80}
\newcommand{\NSweAllPos}{48}
\newcommand{\NSwePooled}{10}
\newcommand{\NSwePooledBH}{1}
\newcommand{\NSwePooledBet}{1}
\newcommand{\NSwePooledBetPrpl}{0}

\newcommand{\NSweMaxWealth}{9.3}
\newcommand{\NSweMaxWealthHi}{24.2}
\newcommand{\NSweObsWealth}{4.0}

\newcommand{\NSweObsWealthGrid}{7.6}
\newcommand{\NSweGridMinAllOnes}{7}
\newcommand{\NShopPer}{0.9}
\newcommand{\NShopAny}{3.5}
\newcommand{\NShopRatio}{3.9}

\newcommand{\NSbT}{27}
\newcommand{\NSbTBH}{21}
\newcommand{\NSbBet}{17}
\newcommand{\NSbBetPrpl}{2}
\newcommand{\NSbBetHi}{10}
\newcommand{\NSbEBH}{10}
\newcommand{\NSbEBHPrpl}{0}
\newcommand{\NSbTasks}{87}
\newcommand{\NSbSplitAdm}{22.8}
\newcommand{\NSbSplitLift}{45}
\newcommand{\NSbSplitPos}{92}
\newcommand{\NSbSplitNot}{6}

\newcommand{\NSbSplitBHAdm}{14.8}
\newcommand{\NSbSplitBHLift}{57}
\newcommand{\NSbSplitBHPos}{99}
\newcommand{\NSbLift}{15.8}
\newcommand{\NSbSD}{0.40}
\newcommand{\NSbTies}{54}
\newcommand{\NSbModelSD}{0.30}
\newcommand{\NSbModelTies}{47}
\newcommand{\NSbPos}{60}
\newcommand{\NSbMedNAlpha}{56}
\newcommand{\NSbMedNBonf}{135}

\newcommand{\NSbCertSix}{81}

\newcommand{\NSbELond}{2.4}
\newcommand{\NSbELondLo}{0}
\newcommand{\NSbELondHi}{5}
\newcommand{\NSbBonfBet}{2}

\newcommand{\NFnTBH}{21}
\newcommand{\NFnTBY}{17}
\newcommand{\NFnTBonf}{16}
\newcommand{\NFnSfBH}{19}
\newcommand{\NFnSfBY}{17}
\newcommand{\NFnSfBonf}{15}

\newcommand{\NFnSplitTBY}{6.4}

\newcommand{\NFnSplitSfBH}{2.2}

\newcommand{\NThHelpful}{60}

\newcommand{\NThOracleRatio}{1.07}
\newcommand{\NThOracleRatioMax}{1.13}
\newcommand{\NThPrplRatio}{2.78}
\newcommand{\NThPrplRatioMax}{3.69}
\newcommand{\NThAgrapaRatio}{1.82}

\newcommand{\NThAgrapaHiRatio}{1.43}

\newcommand{\NThAllIn}{62}
\newcommand{\NThMedTask}{fix\mbox{-}visual\mbox{-}stability}
\newcommand{\NThMedKinf}{0.061}
\newcommand{\NThMedIdeal}{49}
\newcommand{\NThExtraTwoSim}{18.4}
\newcommand{\NThExtraTwoTh}{18.3}

\newcommand{\NThExtraTenSim}{60.7}
\newcommand{\NThExtraTenTh}{60.7}
\newcommand{\NThExtraTwentySim}{78.9}
\newcommand{\NThExtraTwentyTh}{78.9}
\newcommand{\NThPriceOne}{13.6}
\newcommand{\NThPriceGeoThree}{35.4}
\newcommand{\NThFloorGeoThree}{31.4}
\newcommand{\NThPriceZipf}{28.2}
\newcommand{\NThFloorZipf}{26.3}
\newcommand{\NThGeoOnZipfLoss}{12.7}
\newcommand{\NScLogTwenty}{3.00}
\newcommand{\NScZetaWorstTwenty}{5.62}
\newcommand{\NScMonoLoThousand}{1.2625}
\newcommand{\NScMonoHiThousand}{1.2626}
\newcommand{\NScLogHThousand}{2.01}
\newcommand{\NScLogThousand}{6.91}
\newcommand{\NScKLZero}{1.09}
\newcommand{\NScKLHundred}{0.119}
\newcommand{\NScKLThousand}{0.025}
\newcommand{\NScKLPoisonHundred}{0.45}
\newcommand{\NScBoundPoisonHundred}{0.57}
\newcommand{\NScVarHundredSim}{2.1}
\newcommand{\NScVarHundredTh}{2.0}
\newcommand{\NScVarFiveHundredSim}{0.63}
\newcommand{\NScVarFiveHundredTh}{0.58}
\newcommand{\NScVarZetaSim}{1.9}
\newcommand{\NScSlope}{16.3}
\newcommand{\NScOneOverG}{16.5}
\newcommand{\NPilotTasks}{6}
\newcommand{\NPilotScored}{4}

\newcommand{\NPilotWealth}{1.0}
\newcommand{\NPilotDbar}{0.00}
\newcommand{\NPilotRound}{33966073}
\newcommand{\NRvMarginPowBet}{44}
\newcommand{\NRvMarginPowFixed}{81}
\newcommand{\NRvMarginNFixed}{545}
\newcommand{\NRvMarginNBet}{447}
\newcommand{\NRvRThreeNBet}{70}
\newcommand{\NRvRThreeNFixed}{50}
\newcommand{\NRvRThreePowFixed}{80}
\newcommand{\NRvSprtPowLow}{83}
\newcommand{\NRvSprtNLow}{96}
\newcommand{\NRvBetPowLow}{81}
\newcommand{\NRvBetNLow}{157}
\newcommand{\NRvSprtPowHigh}{99}
\newcommand{\NRvSprtNHigh}{139}

\newcommand{\NRvBetNHigh}{178}

\newcommand{\NRvFixedNHigh}{348}
\newcommand{\NRvBetNMaxLow}{300}
\newcommand{\NRvFitSdTheta}{2.25}
\newcommand{\NRvFitSdLift}{2.25}
\newcommand{\NRvFitRho}{0.95}
\newcommand{\NRvFitTie}{52}
\newcommand{\NRvFitSD}{0.37}
\newcommand{\NRvFitBig}{21}
\newcommand{\NRvObsTie}{54}
\newcommand{\NRvObsSD}{0.33}
\newcommand{\NRvObsBig}{23}
\newcommand{\NRvSensNBetLo}{124}
\newcommand{\NRvSensNBetHi}{213}
\newcommand{\NRvSensNFixedLo}{105}
\newcommand{\NRvSensNFixedHi}{160}
\newcommand{\NRvSensPowBetLo}{96}
\newcommand{\NRvSensNCurLo}{50}
\newcommand{\NRvSensNCurHi}{76}
\newcommand{\NRvFitRThreeNBet}{95}
\newcommand{\NRvDefRThreeNBet}{71}

\newcommand{\NRvSensNullMax}{3.1}
\newcommand{\NRvSensSaveLo}{50}
\newcommand{\NRvSensSaveHi}{53}
\newcommand{\NRvLibELond}{127}
\newcommand{\NRvLibELondCert}{82}
\newcommand{\NRvLibHorizon}{88}
\newcommand{\NRvLibHorizonCert}{85}
\newcommand{\NRvLibBonf}{140}
\newcommand{\NRvLibPerExam}{58}
\newcommand{\NRvSelSame}{75}
\newcommand{\NRvSelHeld}{109}
\newcommand{\NRvSelSameFin}{96}
\newcommand{\NRvSelHeldFin}{72}
\newcommand{\NRvSelN}{55}
\newcommand{\NSbPZero}{36.6}
\newcommand{\NAgGridQone}{81}
\newcommand{\NAgKellyQone}{66}
\newcommand{\NAgAgrapaQone}{98}
\newcommand{\NAgPrplQone}{139}
\newcommand{\NAgBoundQone}{109}
\newcommand{\NAgGridBeatsAgrapa}{83}
\newcommand{\NAgExcessMin}{0.24}
\newcommand{\NAgExcessMax}{0.82}
\newcommand{\NAgFixedMin}{1.84}
\newcommand{\NAgFixedMax}{6.08}
\newcommand{\NAgKellyExMax}{0.43}
\newcommand{\NAgMaxReject}{2.1}
\newcommand{\NAgMeanReject}{1.4}
\newcommand{\NAgLevelQone}{2.19}
\newcommand{\NCuValue}{0.474}
\newcommand{\NObUnifQoneNaive}{0.063}
\newcommand{\NObUnifQoneDelay}{0.050}
\newcommand{\NObUnifPiOne}{0.050}
\newcommand{\NObIncNaive}{27.03}
\newcommand{\NObIncDelay}{25.44}
\newcommand{\NObGeoNaive}{48.26}
\newcommand{\NObGeoDelay}{69.32}
\newcommand{\NObBound}{197}
\newcommand{\NEbT}{2000}
\newcommand{\NEbInfFdr}{0.000}
\newcommand{\NEbInfPow}{78.3}
\newcommand{\NEbHorFdr}{0.011}
\newcommand{\NEbHorPow}{87.8}
\newcommand{\NEbPerFdr}{0.105}
\newcommand{\NEbPerPow}{95.2}
\newcommand{\NEbInfPowFirst}{82.0}
\newcommand{\NEbInfPowLast}{76.9}
\newcommand{\NEbElondBinds}{99.7}
\newcommand{\NSecSkills}{232}
\newcommand{\NSecSafe}{185}
\newcommand{\NSecCaution}{39}
\newcommand{\NSecDNI}{8}
\newcommand{\NSecTasksFlag}{8}
\newcommand{\NSecFlagHelpful}{6}

\newcommand{\NSecRho}{-0.16}
\newcommand{\NSecP}{0.14}
\newcommand{\NSecGenuine}{1}

\newcommand{\NSecCatAgree}{6}
\newcommand{\NSecFlagN}{8}
\newcommand{\NLsPlacN}{209}

\newcommand{\NLsPlacExcl}{39}

\newcommand{\NLsPlacLevel}{0.0219}
\newcommand{\NLsLeakN}{191}

\newcommand{\NLsLeakWith}{12}
\newcommand{\NLsLeakWithout}{12}

\newcommand{\NLsLeakLevel}{0.0072}
\newcommand{\NLsHarmN}{150}

\newcommand{\NLsHarmLevel}{0.0038}
\newcommand{\NLsHelpN}{28}

\newcommand{\NLsHelpLevel}{0.0024}
\newcommand{\NLsCost}{2.33}
\newcommand{\NLsRound}{34012203}
\newcommand{\NLsPairs}{624}
\newcommand{\NLsCalCost}{0.08}
\newcommand{\NLsBasePct}{6}
\newcommand{\NLoTasks}{87}
\newcommand{\NLoScored}{29}
\newcommand{\NLoExcluded}{56}
\newcommand{\NLoMissing}{2}
\newcommand{\NLoDbar}{0.13}
\newcommand{\NLoPos}{9}
\newcommand{\NLoNeg}{2}
\newcommand{\NLoWealth}{2.25}
\newcommand{\NLoLcb}{-0.32}
\newcommand{\NLoCost}{75}
\newcommand{\NLoRound}{33980345}
\newcommand{\NLoRuns}{881}
\newcommand{\NLsBPlacN}{209}

\newcommand{\NLsBPlacLevel}{0.0219}

\newcommand{\NLsBHarmN}{26}

\newcommand{\NLsBHarmRun}{30}

\newcommand{\NLsBHarmLevel}{0.0072}

\newcommand{\NLsBHelpN}{52}
\newcommand{\NLsBHelpWith}{54}
\newcommand{\NLsBHelpWithout}{28}
\newcommand{\NLsBHelpRun}{54}

\newcommand{\NLsBHelpLevel}{0.0038}

\newcommand{\NLsBHarmUcb}{0.166}
\newcommand{\NLsResubAdmit}{0}
\newcommand{\NLsResubAdmitPer}{0}
\newcommand{\NLsResubNmin}{186}

\newcommand{\NLsResubLevelFirst}{0.0048}
\newcommand{\NLsResubLevelLast}{0.00157}
\newcommand{\NLsThreeCost}{3.69}
\newcommand{\NLsThreeRound}{34014158}
\newcommand{\NLsBBasePct}{47}
\newcommand{\NLnModel}{nvidia/nemotron-3.5-super-vl}
\newcommand{\NLnMaxTok}{16{,}000}
\newcommand{\NLnPlacN}{232}

\newcommand{\NLnPlacDec}{futility}
\newcommand{\NLnLeakN}{168}

\newcommand{\NLnLeakDec}{futility}
\newcommand{\NLnHarmN}{111}

\newcommand{\NLnHarmDec}{futility}
\newcommand{\NLnHelpN}{29}

\newcommand{\NLnHelpDec}{admit}
\newcommand{\NLnBPlacN}{168}

\newcommand{\NLnBPlacDec}{futility}
\newcommand{\NLnBHarmN}{18}

\newcommand{\NLnBHarmDec}{futility}
\newcommand{\NLnBHelpN}{50}

\newcommand{\NLnBHelpDec}{admit}

\newcommand{\NLnResubNmin}{168}
\newcommand{\NLnResubNmax}{209}

\newcommand{\NLnResubDisc}{5}
\newcommand{\NLnBHarmUcb}{0.195}

\newcommand{\NLnBHarmDisc}{10}
\newcommand{\NLnBHarmSignP}{0.0010}
\newcommand{\NLnCalls}{3{,}348}
\newcommand{\NLnCut}{268}
\newcommand{\NLnCutPct}{8}
\newcommand{\NLnFailed}{2}

\newcommand{\NLnPairs}{1674}

\newcommand{\NLnTotalCost}{9.95}
\newcommand{\NLnRound}{34046500}

\newcommand{\NGbCurGrid}{66}
\newcommand{\NGbCurPrpl}{94}

\newcommand{\NAdSkills}{3}

\newcommand{\NAdClairBelowMax}{39}

\newcommand{\NLsPlacTauWith}{12}
\newcommand{\NLsPlacTauWithout}{11}
\newcommand{\NLsPlacExtraWith}{0}
\newcommand{\NLsPlacExtraWithout}{0}

\newcommand{\NLsPlacGridN}{209}
\newcommand{\NLsLeakTauWith}{12}
\newcommand{\NLsLeakTauWithout}{12}
\newcommand{\NLsLeakExtraWith}{0}
\newcommand{\NLsLeakExtraWithout}{0}

\newcommand{\NLsLeakGridN}{191}
\newcommand{\NLsHarmTauWith}{9}
\newcommand{\NLsHarmTauWithout}{11}
\newcommand{\NLsHarmExtraWith}{0}
\newcommand{\NLsHarmExtraWithout}{0}

\newcommand{\NLsHarmGridN}{150}
\newcommand{\NLsHelpTauWith}{28}
\newcommand{\NLsHelpTauWithout}{0}
\newcommand{\NLsHelpExtraWith}{2}
\newcommand{\NLsHelpExtraWithout}{1}

\newcommand{\NLsHelpGridN}{11}
\newcommand{\NLsBPlacTauWith}{98}
\newcommand{\NLsBPlacTauWithout}{97}
\newcommand{\NLsBPlacExtraWith}{0}
\newcommand{\NLsBPlacExtraWithout}{0}

\newcommand{\NLsBPlacGridN}{209}
\newcommand{\NLsBHarmTauWith}{1}
\newcommand{\NLsBHarmTauWithout}{10}
\newcommand{\NLsBHarmExtraWith}{0}
\newcommand{\NLsBHarmExtraWithout}{3}

\newcommand{\NLsBHarmGridN}{26}
\newcommand{\NLsBHelpTauWith}{52}
\newcommand{\NLsBHelpTauWithout}{26}
\newcommand{\NLsBHelpExtraWith}{2}
\newcommand{\NLsBHelpExtraWithout}{2}

\newcommand{\NLsBHelpGridN}{18}
\newcommand{\NLnPlacTauWith}{12}
\newcommand{\NLnPlacTauWithout}{10}
\newcommand{\NLnPlacExtraWith}{0}
\newcommand{\NLnPlacExtraWithout}{0}

\newcommand{\NLnPlacGridN}{232}
\newcommand{\NLnLeakTauWith}{7}
\newcommand{\NLnLeakTauWithout}{8}
\newcommand{\NLnLeakExtraWith}{0}
\newcommand{\NLnLeakExtraWithout}{0}

\newcommand{\NLnLeakGridN}{168}
\newcommand{\NLnHarmTauWith}{0}
\newcommand{\NLnHarmTauWithout}{5}
\newcommand{\NLnHarmExtraWith}{0}
\newcommand{\NLnHarmExtraWithout}{0}

\newcommand{\NLnHarmGridN}{111}
\newcommand{\NLnHelpTauWith}{29}
\newcommand{\NLnHelpTauWithout}{1}
\newcommand{\NLnHelpExtraWith}{1}
\newcommand{\NLnHelpExtraWithout}{0}

\newcommand{\NLnHelpGridN}{12}
\newcommand{\NLnBPlacTauWith}{91}
\newcommand{\NLnBPlacTauWithout}{92}
\newcommand{\NLnBPlacExtraWith}{1}
\newcommand{\NLnBPlacExtraWithout}{1}

\newcommand{\NLnBPlacGridN}{168}
\newcommand{\NLnBHarmTauWith}{0}
\newcommand{\NLnBHarmTauWithout}{10}
\newcommand{\NLnBHarmExtraWith}{0}
\newcommand{\NLnBHarmExtraWithout}{0}

\newcommand{\NLnBHarmGridN}{18}
\newcommand{\NLnBHelpTauWith}{50}
\newcommand{\NLnBHelpTauWithout}{24}
\newcommand{\NLnBHelpExtraWith}{4}
\newcommand{\NLnBHelpExtraWithout}{1}

\newcommand{\NLnBHelpGridN}{15}
\newcommand{\NEbhGridElondFdr}{0.1}
\newcommand{\NEbhGridElondPow}{81}
\newcommand{\NEbhGridElondN}{356}
\newcommand{\NEbhGridSebhFdr}{1.7}
\newcommand{\NEbhGridSebhPow}{91}
\newcommand{\NEbhGridSebhN}{338}

\newcommand{\NEbhGridSupPow}{81}
\newcommand{\NEbhGridSupN}{361}

\newcommand{\NEbhGridNmaxPow}{89}
\newcommand{\NEbhGridNmaxN}{422}

\newcommand{\NEbhPrplSupPow}{59}

\newcommand{\NEThreePrplPowELond}{80}
\newcommand{\NEThreePrplNELond}{369}

\newcommand{\NCalFourRealWith}{19}
\newcommand{\NCalFourRealWithout}{15}
\newcommand{\NCalFourPlacWith}{19}
\newcommand{\NCalFourPlacWithout}{17}

\newcommand{\NCalFourLift}{0.20}

\newcommand{\NLsFourPlacDec}{futility}
\newcommand{\NLsFourPlacN}{69}
\newcommand{\NLsFourPlacLevel}{0.0219}
\newcommand{\NLsFourPlacTauWith}{57}
\newcommand{\NLsFourPlacTauWithout}{65}
\newcommand{\NLsFourPlacExtraWith}{3}
\newcommand{\NLsFourPlacExtraWithout}{3}

\newcommand{\NLsFourRealDec}{admit}
\newcommand{\NLsFourRealN}{72}
\newcommand{\NLsFourRealLevel}{0.0072}
\newcommand{\NLsFourRealTauWith}{71}
\newcommand{\NLsFourRealTauWithout}{61}
\newcommand{\NLsFourRealExtraWith}{0}
\newcommand{\NLsFourRealExtraWithout}{0}

\newcommand{\NLnFourCost}{1.43}

\newcommand{\NLnFourPlacDec}{cap}
\newcommand{\NLnFourPlacN}{600}

\newcommand{\NLnFourPlacTauWith}{471}
\newcommand{\NLnFourPlacTauWithout}{457}
\newcommand{\NLnFourPlacExtraWith}{2}
\newcommand{\NLnFourPlacExtraWithout}{1}
\newcommand{\NLnFourPlacExcl}{171}
\newcommand{\NLnFourPlacRun}{774}
\newcommand{\NLnFourPlacFailWith}{108}
\newcommand{\NLnFourPlacFailWithout}{119}
\newcommand{\NLnFourRealDec}{admit}
\newcommand{\NLnFourRealN}{64}

\newcommand{\NLnFourRealTauWith}{62}
\newcommand{\NLnFourRealTauWithout}{49}
\newcommand{\NLnFourRealExtraWith}{0}
\newcommand{\NLnFourRealExtraWithout}{0}
\newcommand{\NLnFourRealExcl}{230}
\newcommand{\NLnFourRealRun}{294}
\newcommand{\NLnFourRealFailWith}{168}
\newcommand{\NLnFourRealFailWithout}{180}
\newcommand{\NLnFourCut}{4}
\newcommand{\NCalFourBRealWith}{16}
\newcommand{\NCalFourBRealWithout}{0}
\newcommand{\NCalFourBPlacWith}{2}
\newcommand{\NCalFourBPlacWithout}{0}

\newcommand{\NCalFourBLift}{0.80}

\newcommand{\NLsFourBPlacDec}{admit}
\newcommand{\NLsFourBPlacN}{44}
\newcommand{\NLsFourBPlacLevel}{0.0219}
\newcommand{\NLsFourBPlacTauWith}{9}
\newcommand{\NLsFourBPlacTauWithout}{1}
\newcommand{\NLsFourBPlacExtraWith}{2}
\newcommand{\NLsFourBPlacExtraWithout}{0}

\newcommand{\NLsFourBRealDec}{admit}
\newcommand{\NLsFourBRealN}{13}
\newcommand{\NLsFourBRealLevel}{0.0072}
\newcommand{\NLsFourBRealTauWith}{11}
\newcommand{\NLsFourBRealTauWithout}{1}
\newcommand{\NLsFourBRealExtraWith}{4}
\newcommand{\NLsFourBRealExtraWithout}{0}

\newcommand{\NLnFourBPlacDec}{admit}
\newcommand{\NLnFourBPlacN}{68}

\newcommand{\NLnFourBPlacTauWith}{8}
\newcommand{\NLnFourBPlacTauWithout}{0}
\newcommand{\NLnFourBPlacExtraWith}{0}
\newcommand{\NLnFourBPlacExtraWithout}{0}

\newcommand{\NLnFourBRealDec}{admit}
\newcommand{\NLnFourBRealN}{16}

\newcommand{\NLnFourBRealTauWith}{11}
\newcommand{\NLnFourBRealTauWithout}{1}
\newcommand{\NLnFourBRealExtraWith}{1}
\newcommand{\NLnFourBRealExtraWithout}{0}

\newcommand{\NCalFourCRealWith}{20}
\newcommand{\NCalFourCRealWithout}{2}
\newcommand{\NCalFourCPlacWith}{1}
\newcommand{\NCalFourCPlacWithout}{1}

\newcommand{\NCalFourCLift}{0.90}

\newcommand{\NLsFourCPlacDec}{futility}
\newcommand{\NLsFourCPlacN}{162}
\newcommand{\NLsFourCPlacLevel}{0.0219}
\newcommand{\NLsFourCPlacTauWith}{11}
\newcommand{\NLsFourCPlacTauWithout}{15}
\newcommand{\NLsFourCPlacExtraWith}{0}
\newcommand{\NLsFourCPlacExtraWithout}{0}

\newcommand{\NLsFourCRealDec}{admit}
\newcommand{\NLsFourCRealN}{11}
\newcommand{\NLsFourCRealLevel}{0.0072}
\newcommand{\NLsFourCRealTauWith}{11}
\newcommand{\NLsFourCRealTauWithout}{1}
\newcommand{\NLsFourCRealExtraWith}{1}
\newcommand{\NLsFourCRealExtraWithout}{0}

\newcommand{\NLnFourCPlacDec}{futility}
\newcommand{\NLnFourCPlacN}{186}

\newcommand{\NLnFourCPlacTauWith}{0}
\newcommand{\NLnFourCPlacTauWithout}{0}
\newcommand{\NLnFourCPlacExtraWith}{0}
\newcommand{\NLnFourCPlacExtraWithout}{0}

\newcommand{\NLnFourCRealDec}{admit}
\newcommand{\NLnFourCRealN}{12}

\newcommand{\NLnFourCRealTauWith}{11}
\newcommand{\NLnFourCRealTauWithout}{1}
\newcommand{\NLnFourCRealExtraWith}{0}
\newcommand{\NLnFourCRealExtraWithout}{0}

\newcommand{\NCalNFourRealWith}{18}
\newcommand{\NCalNFourRealWithout}{15}
\newcommand{\NCalNFourPlacWith}{17}
\newcommand{\NCalNFourPlacWithout}{16}
\newcommand{\NLsFourRealLiftTau}{0.14}
\newcommand{\NLnFourRealLiftTau}{0.20}
\newcommand{\NLFourBCMinN}{11}
\newcommand{\NLFourBCMaxN}{16}
\newcommand{\NPowCons}{0.86}
\newcommand{\NPowNmax}{600}
\newcommand{\NPowNoise}{0.10}
\newcommand{\NPowConsVone}{1.00}
\newcommand{\NLFourTotalCost}{2.06}
\newcommand{\NLsFourBPlacOverN}{31}
\newcommand{\NLsFourBPlacOverWith}{12}
\newcommand{\NLsFourBPlacOverWithout}{0}
\newcommand{\NLsFourBPlacMoneyN}{26}
\newcommand{\NLsFourBPlacMoneyWith}{10}
\newcommand{\NLsFourBPlacMoneyWithout}{0}

\newcommand{\NLnFourBPlacOverN}{43}
\newcommand{\NLnFourBPlacOverWith}{15}
\newcommand{\NLnFourBPlacOverWithout}{1}
\newcommand{\NLnFourBPlacMoneyN}{42}
\newcommand{\NLnFourBPlacMoneyWith}{11}
\newcommand{\NLnFourBPlacMoneyWithout}{1}

\newcommand{\NThPrplRatioLo}{2.25}
\newcommand{\NThPrplRatioHi}{3.22}
\newcommand{\NSbMedNAlphaLo}{31}
\newcommand{\NSbMedNAlphaHi}{75}
\newcommand{\NSbMedNBonfLo}{62}
\newcommand{\NSbMedNBonfHi}{178}

\newcommand{\NIvEbInfFdr}{0.03}
\newcommand{\NIvEbInfFdrHi}{0.04}

\newcommand{\NNsControl}{2.9}
\newcommand{\NNsMildMax}{5.1}
\newcommand{\NNsSkillArm}{15.5}
\newcommand{\NNsBlock}{11.5}
\newcommand{\NNsRepeat}{7.4}
\newcommand{\NNsExclude}{9.9}
\newcommand{\NNsCoin}{22.2}
\newcommand{\NFuFdr}{0.12}
\newcommand{\NFuNullLin}{0.06}
\newcommand{\NFuHonFalse}{0.02}
\newcommand{\NFuHonAdm}{74}
\newcommand{\NFuAdm}{25.8}
\newcommand{\NMpDesEightyFixed}{137}
\newcommand{\NMpDesEightyFixedNull}{137}
\newcommand{\NMpDesEightyGrid}{148}
\newcommand{\NMpDesEightyGridNull}{227}

\newcommand{\NMpDesNinetyFiveFixed}{239}

\newcommand{\NMpDesNinetyFiveGrid}{167}

\newcommand{\NMpCurEightyFixed}{55}

\newcommand{\NMpCurEightyGrid}{57}

\newcommand{\NMpCurNinetyFiveFixed}{96}

\newcommand{\NMpCurNinetyFiveGrid}{64}

\newcommand{\NMpNullMaxGrid}{2.9}
\newcommand{\NLcPredErrMean}{1.5}
\newcommand{\NLcPredErrMax}{6}
\newcommand{\NLcBindMax}{0.52}
\newcommand{\NLcNullMax}{0.3}
\newcommand{\NLcHonFalseMax}{0.25}
\newcommand{\NLcGeoUseless}{758}
\newcommand{\NLcGeoHelp}{470}
\newcommand{\NLcUniUseless}{3580}
\newcommand{\NLcUniHelp}{483}
\newcommand{\NLcHelpAgeFifty}{437}
\newcommand{\NLcHelpAgeTwoK}{498}
\newcommand{\NLcAdmAgeFifty}{72}
\newcommand{\NLcAdmAgeTwoK}{55}
\newcommand{\NLcSchedSpread}{0.5}
\newcommand{\NLcUselessNeg}{591}
\newcommand{\NLcUselessZero}{1603}
\newcommand{\NLsFiveCost}{1.29}
\newcommand{\NLsFiveRound}{34127468}
\newcommand{\NLsFiveTeachN}{96}
\newcommand{\NLsFiveTeachDec}{futility}
\newcommand{\NLsFiveTeachLift}{$-$7.3}
\newcommand{\NLsFiveTeachLevel}{0.0219}

\newcommand{\NLsFiveHarmN}{29}
\newcommand{\NLsFiveHarmDec}{futility}
\newcommand{\NLsFiveHarmLift}{$-$31.0}
\newcommand{\NLsFiveHarmLevel}{0.0072}

\newcommand{\NLsFivePlacN}{144}
\newcommand{\NLsFivePlacDec}{futility}
\newcommand{\NLsFivePlacLift}{$-$5.6}
\newcommand{\NLsFivePlacLevel}{0.0038}

\newcommand{\NLsFiveHelpN}{161}
\newcommand{\NLsFiveHelpDec}{futility}
\newcommand{\NLsFiveHelpLift}{$-$5.0}
\newcommand{\NLsFiveHelpLevel}{0.0024}

\newcommand{\NLsFiveSkillN}{48}
\newcommand{\NLsFiveSkillDec}{futility}
\newcommand{\NLsFiveSkillLift}{$-$18.8}
\newcommand{\NLsFiveSkillLevel}{0.0017}
\newcommand{\NLsFiveSkillCutWith}{6}

\newcommand{\NLsFiveSkillRun}{48}
\newcommand{\NLsFiveCutPct}{2}
\newcommand{\NLnFiveCost}{5.30}
\newcommand{\NLnFiveRound}{34128169}
\newcommand{\NLnFiveTeachN}{76}
\newcommand{\NLnFiveTeachDec}{futility}
\newcommand{\NLnFiveTeachLift}{$-$9.2}

\newcommand{\NLnFiveHarmN}{600}
\newcommand{\NLnFiveHarmDec}{cap}
\newcommand{\NLnFiveHarmLift}{+5.7}
\newcommand{\NLnFiveHarmLevel}{0.0072}

\newcommand{\NLnFiveHarmCutPctWith}{1}

\newcommand{\NLnFiveHarmCutPctWithout}{35}

\newcommand{\NLnFiveHarmLcb}{$-$2.1}
\newcommand{\NLnFivePlacN}{62}
\newcommand{\NLnFivePlacDec}{futility}
\newcommand{\NLnFivePlacLift}{$-$11.3}

\newcommand{\NLnFiveHelpN}{139}
\newcommand{\NLnFiveHelpDec}{futility}
\newcommand{\NLnFiveHelpLift}{$-$2.9}

\newcommand{\NLnFiveSkillN}{46}
\newcommand{\NLnFiveSkillDec}{futility}
\newcommand{\NLnFiveSkillLift}{$-$19.6}

\newcommand{\NLnFiveCutPct}{25}
\newcommand{\NLFiveTotalCost}{6.58}
\newcommand{\NLnFiveHarmBootSix}{18}
\newcommand{\NLnFiveHarmBootMed}{1{,}205}
\newcommand{\NLnFiveHarmBootTwoK}{81}
\newcommand{\NLnFiveHarmFourTwoK}{33}
\newcommand{\NLnFiveHarmValid}{602}

%% file: sections/abstract_jmlr.tex
Skill registries for LLM agents decide which skills to list by running an agent with and without each skill. These decisions are sequential, span many skills, and face resubmission. We assemble a gate that stays valid under all three: a paired exam on commit-reveal hidden variants, a betting test with a futility margin, online e-value false-discovery control with alpha-spending over a skill's resubmissions, and a credential from which anyone can recompute the decision. For gates whose exams are fixed in advance, we show that resubmission protection costs an honest author, up to additive constants, at least the entropy of their attempt count in units of the information per variant; spending the error budget according to the attempt-count law attains this, and a level-tuned mixture of bets attains it to first order without knowing the effect. In simulation, in a registry past its first 50 exams that also controls its false-discovery rate online, that control sets the level in more than 99\% of exams, so resubmission protection is nearly free. Current per-skill evidence is too thin for sequential certification: on two public benchmarks, no betting test admits a skill from 10 paired units, and online control admits \NSbELond{} of \NSbTasks{} SkillsBench skills from 18. In committed live studies on generated task families with Claude Haiku~5.5 and the open-weight Nemotron~3.5 Super VL, each costing under US\$10, the gate admitted every skill whose lift could be certified within the committed budget, including both skills we wrote to help and all three published skills it tested, and refused a planted answer leak, skills with wrong instructions and every placebo that did not itself raise the pass rate. On a task family built by Google DeepMind no published math skill helped, and the gate refused them all; with Nemotron a skill we wrote to hurt helped slightly and reached the variant cap unadmitted. Every decision can be recomputed from the released ledger and transcripts.

%% file: sections/intro.tex
\section{Introduction}

An agent skill is a directory with a \texttt{SKILL.md} file, optional scripts and reference material that an LLM agent reads when a task calls for it \citep{anthropic2025introducingskills,agentskills2026spec}.\footnote{The authors used an LLM-based assistant (Claude, Anthropic) for code, analysis scripts and drafting. The authors checked every claim, number and citation, and take full responsibility for the content.} Public directories now index skills at large scale (one advertises more than 276{,}000 \citep{agentskill2026}), and registries have started to gate listing or signing on a with-and-without evaluation \citep{nvidia2026trustpipeline,skillmill2026}. Whether a skill helps is not obvious. SkillsBench found that curated skills raised the average pass rate from 33.9\% to 50.5\%, while skills an agent wrote for itself scored 8.1 to 11.5 points below no skill in the configurations tested \citep{li2026skillsbench}. SWE-Skills-Bench found no significant effect for 92 of 98 skills \citep{han2026sweskillsbench}, and WebDev-Skills-Bench found that injecting skills lowered mean Pass@2 by 1.3\% to 4.2\% \citep{yang2026webdevskills}.

A registry that admits skills is therefore running a hypothesis test, many times, on data that arrive sequentially, for authors who can resubmit. Each of these features defeats a familiar procedure used in the obvious way. A confidence interval re-checked as results arrive loses its coverage; a test applied to hundreds of skills without correction produces false discoveries at the nominal rate; and an author who resubmits a lightly edited skill until it passes will eventually succeed. Prespecified fixed-sample tests, multiplicity corrections and group-sequential designs remain valid when used as designed; the difficulty is that a registry needs all three protections at once, on exams that it wants to stop early. The registry's own record also has to be trustworthy. A reader should be able to check that the exam was not chosen after seeing the skill, that it was not stopped at a lucky moment, and that the reported decision follows from the transcripts.

The problem is not specific to skills. Model leaderboards, benchmark submission servers and any certification that lets a submitter retry face the same combination of sequential data, multiplicity and resubmission; skills are our motivating case and the source of our data. This paper asks two questions. First, can the per-skill evidence that current benchmarks collect support such decisions? Second, what does a decision procedure that is valid under optional stopping, multiplicity and resubmission cost, and how can a third party check it?

\paragraph{Findings.}
\begin{enumerate}
\item \textbf{Current per-skill evidence is too thin for sequential certification at library scale} (Section~\ref{sec:res-real}). In the released paired outcomes of SWE-Skills-Bench, \NSweGolden{} cells are ``golden'' under the source paper's uncorrected rule, \NSweBH{} survive Benjamini-Hochberg control, and no betting test admits any cell with 10 units. In SkillsBench, a one-off Benjamini-Hochberg analysis of fixed-sample $t$-tests certifies \NSbTBH{} of \NSbTasks{} skills, and they replicate on held-out configurations, so the signal is real. But a registry tests skills as they arrive: with 18 units, a per-exam betting test admits \NSbBet{} skills, and online control over skills in arrival order (e-LOND) keeps \NSbELond{} on average.
\item \textbf{Common decision rules do not hold their error rate, and per-exam validity is not enough} (Sections~\ref{sec:res-validity} and~\ref{sec:res-library}). A $t$-test re-run after each variant admits useless skills \NEOnePeekTMin\% to \NEOnePeekTMax\% of the time, and a 95\% interval checked every ten variants \NEOnePeekCIMin\% to \NEOnePeekCIMax\% of the time, against a nominal 5\%; a stale stored baseline breaks every rule, valid or not (\NEOneStaleMin\% to \NEOneStaleMax\%). A useless skill resubmitted \PResub{} times passes a per-exam gate \NResubPerExam\% of the time, and alpha-spending over its lineage holds this to \NResubLineage\%. Stopping each exam at $1/\alpha$ and then applying e-BH admits nothing, because every stopped e-value is close to $1/\alpha$ (Remark~\ref{prop:ebh}).
\item \textbf{Resubmission-proofness has a price set by an entropy, and in a running registry it is nearly free} (Section~\ref{sec:theory}). One exam at level $a$ needs about $\log(1/a)/K_{\inf}$ hidden variants, where $K_{\inf}$ is the information per variant against the closest useless skill \citep{agrawal2025stopping}. Controlling error over a skill's resubmissions costs an honest author at least $H(\pi)/K_{\inf}$ extra variants, where $H(\pi)$ is the entropy of their attempt count; spending according to $\pi$ attains this, and a schedule $q\neq\pi$ adds $\mathrm{KL}(\pi\|q)/K_{\inf}$. So choosing a schedule is a coding problem, with a minimax schedule over a class of attempt-count laws and a learned schedule whose excess falls like $N/m$ after $m$ lineages. A level-tuned mixture of bets keeps the price, with leading coefficient one, when the gate does not know the effect. The price is what the lineage guarantee costs on its own: in a registry past 50 exams with online false-discovery control, the online level is the stricter one in all but \NLcBindMax\% of exams, and the schedule moves the cost by under \NLcSchedSpread\% (E19). It binds for a new registry and for one that budgets lineages separately.
\item \textbf{Validity has a cost that can be planned} (Section~\ref{sec:res-cost}). Resampling the released SkillsBench outcomes, certifying the median helpful curated skill needs \NSbMedNAlpha{} hidden variants per exam, or \NSbMedNBonf{} with library-wide control (95\% bootstrap intervals over skills \NSbMedNAlphaLo--\NSbMedNAlphaHi{} and \NSbMedNBonfLo--\NSbMedNBonfHi), against the 18 paired units available today; the standard capped plug-in bet needs a median \NThPrplRatio{} times (\NThPrplRatioLo--\NThPrplRatioHi) the information lower bound. These figures treat the benchmark's configurations as stand-ins for hidden variants, an exchangeability the released data cannot test, so they are planning estimates.
\item \textbf{The gate works on live models and published skills} (Section~\ref{sec:res-live}). With Claude Haiku~5.5 and the open-weight Nemotron~3.5 Super VL, in two task families we designed, it admitted both helpful skills (after \NLsHelpN{} and \NLsBHelpN{} variants with Haiku) and refused placebos, a skill carrying answers to public examples and skills with wrong instructions. On three published skills by other authors it admitted every skill on both models, and also a placebo that itself raised the pass rate. On a task family built by Google DeepMind, no published math skill showed a positive lift on either model and the gate refused all five skills, for US\$\NLFiveTotalCost{} in all; with Nemotron a skill we wrote to hurt helped slightly and ran to the cap without admission. On the SkillsBench tasks, where harness failures left \NLoScored{} of \NLoTasks{} tasks scorable, it admitted nothing. Every decision can be recomputed from the ledger and transcripts, released with the code.
\end{enumerate}

\paragraph{Contributions.} (i) A statement of skill admission as a sequential, multiple and resubmittable testing problem, with four requirements a registry decision should meet (Section~\ref{sec:setup}). (ii) A gate that meets them, assembled from existing tools: a paired exam on commit-reveal hidden variants, the betting test of \citet{waudbysmith2024betting} with a futility margin, e-LOND \citep{xu2024online} with lineage alpha-spending, and a dual-arm credential whose decision anyone can recompute from published transcripts (Section~\ref{sec:gate}). The statistical components are not new. Lineage spending is classical online family-wise control within one skill's resubmissions \citep{tian2021online}, and the failure of ``stop at $1/\alpha$, then e-BH'' for any bet cap is an elementary observation (Remark~\ref{prop:ebh}). What is new is their combination for admission under resubmission. (iii) A characterization of what resubmission-proofness costs an honest author, namely an entropy lower bound over lineage-valid gates whose exams are fixed in advance, attained by spending according to the law of attempt counts, with the mismatch cost of any other schedule, the minimax and learned schedules that follow from it, and an explicit finite-level version for a gate that does not know the effect (Section~\ref{sec:theory}). It builds on known bounds for the cost of one exam \citep{agrawal2025stopping,larsson2025numeraire}; the proof averages the one-exam bound over attempts, and what it adds is the explicit price of resubmission-proofness and the schedule that attains it. (iv) A reanalysis of released per-skill paired outcomes for admission decisions, a simulation study calibrated to them, and committed live studies on two models (Sections~\ref{sec:eval}--\ref{sec:results}). (v) An open reference implementation with a verifier and tests.

\paragraph{What we do not claim.} This work began as the design of a ``campus'' in which agents distill skills from shared traces and a skill enters a shared library only after an exam. Whether such peer-synthesized skills beat self-generated ones is not answered here (Appendix~\ref{app:campus} gives a study protocol). We also do not claim that per-skill testing or anytime-valid evaluation of LLMs is new: SWE-Skills-Bench already applies a per-skill significance test \citep{han2026sweskillsbench}, and anytime-valid procedures have been applied to self-evolving LLM systems and leaderboards \citep{yin2026enough,gao2026bbedge}.

%% file: sections/related.tex
\section{Related work}
\label{sec:related}

\paragraph{Evaluating skills.} SkillsBench \citep{li2026skillsbench}, SWE-Skills-Bench \citep{han2026sweskillsbench} and WebDev-Skills-Bench \citep{yang2026webdevskills} compare agents with and without a skill on fixed task sets. SWE-Skills-Bench classifies each skill with an exact paired sign-flip test on ten units at $p<0.05$, a per-skill decision rule with fixed sample size and no multiplicity correction. ACES \citep{kevin2026aces}, SkillTester \citep{wang2026skilltester}, the skillcheck tool \citep{sx4im2026skillcheck} and NVIDIA's signing pipeline \citep{nvidia2026trustpipeline} report intervals or verdicts for individual skills. Lift depends on the model: skill rankings transfer weakly across models \citep{yang2026webdevskills}, and a skill that helps one backbone can hurt another \citep{yu2026masa}. OpenSkillEval \citep{ying2026openskilleval} audits 30 community skills on generated tasks and finds that many popular skills do not consistently beat agents without them. PROMPT \citep{hu2026prompt} evaluates the components of clinical prompts in pre-registered randomized experiments with matched controls and fixed sample sizes. We take the paired design from this work and ask what a sequential registry decision built on it must satisfy.

\paragraph{Anytime-valid evaluation.} \citet{yin2026enough} use an anytime-valid detector on paired outcomes to decide when a self-evolving LLM system has stopped improving; BB-EDGE \citep{gao2026bbedge} and AV-AIVAT \citep{li2026avaivat} give anytime-valid leaderboards and agent comparisons. Our tools come from the same literature: betting confidence sequences \citep{waudbysmith2024betting}, e-values \citep{ramdas2023gametheoretic,grunwald2024safe}, e-BH \citep{wang2022ebh}, online testing with e-values \citep{xu2024online} and always-valid A/B tests \citep{johari2022always}. None of these addresses admission across a library with resubmissions. The reusable holdout and the Ladder leaderboard \citep{dwork2015reusable,blum2015ladder} protect one fixed test set that is queried again and again, by perturbing or coarsening its answers. Our setting differs: every exam draws fresh hidden variants, so reuse of test data is not the threat; repeated submission of new versions is, and lineage spending prices it (Section~\ref{sec:theory}).

\paragraph{Verifiable records.} Transparency logs \citep{laurie2013ct,newman2022sigstore} and public beacons \citep{drand2026beacon} provide the append-only record and randomness the protocol needs; any such log can serve, and a public blockchain is one option rather than a requirement. Agent evidence logs \citep{bromme2026blackbox,seyedkazemi2026novafabric} and patent filings on ledger-based agent credentials \citep{drivehealth2026credentialing,dayama2022ibm} record single evaluations; our credential binds both arms of a paired test and the sequential decision, so the decision itself can be recomputed. Appendix~\ref{app:related} covers this literature and skill synthesis in more detail.

%% file: sections/setup.tex
\section{The admission problem}
\label{sec:setup}

Table~\ref{tab:notation} in Appendix~\ref{app:notation} collects the notation.

\paragraph{Exam.} Fix an agent $a$ (model, harness and configuration), a candidate skill $s$ and a comparator $s_0$ (no skill, or the incumbent skill for the same task class). A task class has a generator $G$ that maps a seed $\sigma$ and an index $i$ to a task instance with a deterministic verifier. For variant $i$ the agent attempts the instance $r$ times in each arm, in identical sandboxes, with arm order randomized; $Y^{(1)}_{i\ell},Y^{(0)}_{i\ell}\in[0,1]$ are the verifier scores. The paired difference is
\begin{equation}
D_i=\frac1r\sum_{\ell=1}^r\bigl(Y^{(1)}_{i\ell}-Y^{(0)}_{i\ell}\bigr)\in[-1,1],
\label{eq:D}
\end{equation}
Let $\mathcal F_{i-1}$ hold everything fixed before variant $i$ is scored: the registry's history when the exam starts (earlier exams, their decisions and levels), the skill, the comparator, the commitment, any state fixed for the whole exam, and the first $i-1$ variants' transcripts, taken in index order. The lift of variant $i$ given the exam so far is $\Delta_i=\mathbb E[D_i\mid\mathcal F_{i-1}]$. For a margin $\delta\ge0$ fixed in advance the registry tests
\begin{equation}
H_0:\ \Delta_i\le\delta\ \text{for every } i\quad\text{against}\quad H_1:\ \Delta=\mathbb E[D_i]>\delta .
\label{eq:H}
\end{equation}
The null is stated conditionally because that is what sequential validity needs. A null on the marginal mean is not enough: if one fair coin decides, for a whole exam, whether every difference is $+1$ or $-1$, then $\mathbb E[D_i]=0$, yet a constant bet of $1/2$ in Eq.~\eqref{eq:wealth} reaches $1/\alpha=20$ after 14 differences, so the exam admits with probability $1/2$. The two nulls coincide under the design we require: variants drawn i.i.d.\ from $G$ with seeds $H(\sigma,i)$ that no one knows before the exam (Section~\ref{sec:gate}); agent calls independent given the variant, with no memory, caching or batching shared across variants; arm order randomized; outcomes consumed in index order, not in order of completion; no repeated seeds; and exclusion rules that do not depend on the arm. The $D_i$ are then i.i.d.\ given the exam-level state and $\Delta_i=\Delta$. State that is fixed for the whole exam, such as the model snapshot the provider serves, belongs to $\mathcal F_0$, so a certificate is conditional on it: it says that the skill helped this agent as it was served during the exam. Section~\ref{sec:discussion} lists the departures that break the condition. Pairing removes between-variant difficulty from $\operatorname{Var}(D_i)$, which is what makes small lifts detectable.

\paragraph{Requirements.} We ask four things of a registry's decision rule.
\begin{description}
\item[R1, validity under optional stopping.] $\Pr_{H_0}(\text{admit})\le\alpha$ whatever rule the proctor uses to stop the exam, including stopping as soon as the evidence looks good.
\item[R2, validity across the library and over time.] Each exam tests one null, Eq.~\eqref{eq:H} for its version against its comparator, and an admission is a false discovery when that null is true. The expected share of false admissions among all admissions (the false discovery rate) is at most $\alpha$ over the registry's history, under any dependence between exams, such as shared agents and generators, that leaves each null exam's conditional null intact given the registry's history. Section~\ref{sec:gate} fixes the convention under which exams and admitted skills are the same unit.
\item[R3, resubmission.] An author who resubmits versions of a skill any number of times gets a version with $\Delta\le\delta$ admitted with probability at most $\alpha$ over the skill's lineage, whatever helpful versions the lineage also contains.
\item[R4, recomputability.] A third party can check, from published material, that the variants were fixed after the skill, that the stopping rule was the registered one, and that the reported decision follows from the transcripts.
\end{description}
R1 rules out re-checking a fixed-$n$ test or interval; R2 rules out per-skill testing at $\alpha$ without correction; R3 rules out any rule whose per-exam level does not shrink with resubmissions; R4 requires that the exam seed and stopping rule be committed before the evidence exists. Section~\ref{sec:results} measures how far current practice is from each requirement.

%% file: sections/gate.tex
\section{An admission gate that meets R1 to R4}
\label{sec:gate}

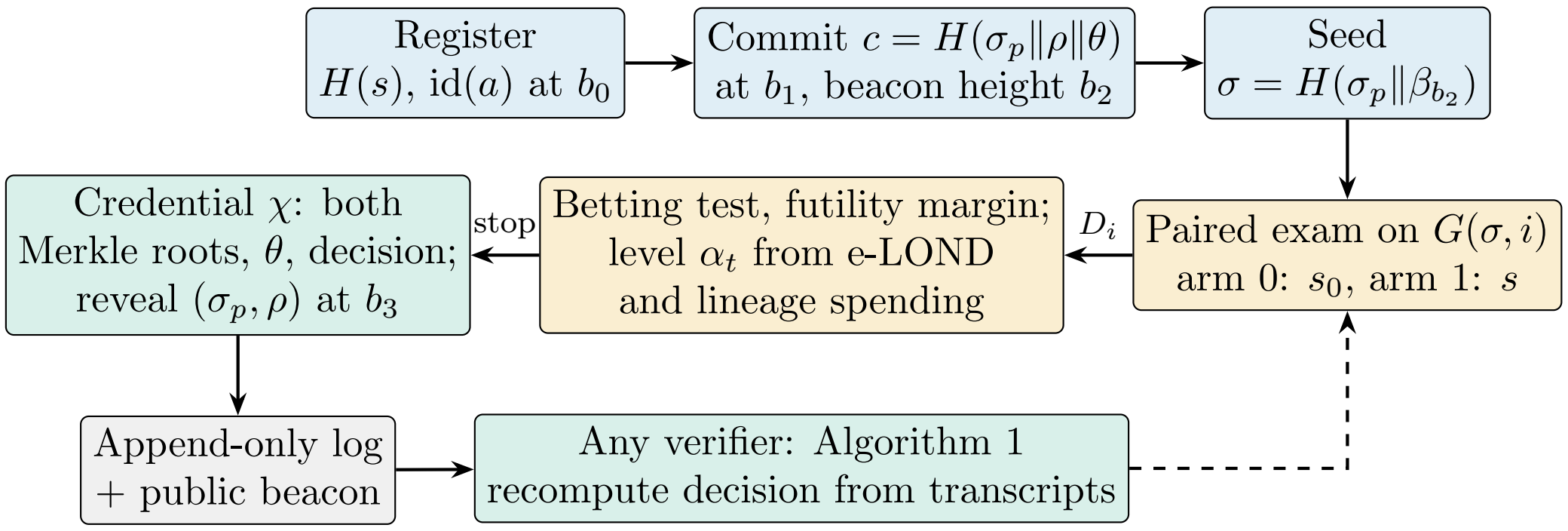
\begin{figure}[t]
\centering
\input{figures/F1_protocol.tex}
\caption{Admission protocol. The skill digest is logged before the proctor commits to the exam parameters $\theta$ and a secret $\sigma_p$; the variant seed mixes $\sigma_p$ with a public beacon value that does not yet exist at commit time. The exam runs paired arms on variants $G(\sigma,i)$ and stops by the registered rule. The credential binds both arms' transcript Merkle roots and the decision, and any verifier can recompute the decision (Algorithm~\ref{alg:verify}). Any append-only log with a public beacon can serve (a public chain, or a transparency log plus drand).}
\label{fig:protocol}
\end{figure}

Figure~\ref{fig:protocol} shows the protocol. Each component below answers one requirement.

\subsection{Hidden variants by commit and reveal (R4)}
The registry logs the skill digest $H(s)$ and agent identity at log height $b_0$. The proctor then draws a secret $\sigma_p$ and salt $\rho$ and logs $c=H(\sigma_p\,\|\,\rho\,\|\,\theta)$ at $b_1>b_0$, where $\theta$ is the registered parameter block (test, $\alpha$, margins, $r$, $N_{\max}$, bet rule, generator version, beacon height $b_2>b_1$). The exam seed is $\sigma=H(\sigma_p\,\|\,\beta_{b_2})$, where $\beta_{b_2}$ is the beacon value at $b_2$. After the exam the proctor reveals $(\sigma_p,\rho)$. Because $H(s)$ precedes $c$, the skill cannot depend on the variants; because $\sigma_p$ is secret, the agent cannot predict them; because $c$ precedes $\beta_{b_2}$, the proctor cannot choose variants after seeing the skill. Commitments without credentials stay visible, so abandoned exams cannot be hidden. A colluding proctor who leaks $\sigma_p$ is not stopped by this construction; threshold proctors or seed release inside a trusted execution environment address that case (Section~\ref{sec:discussion}).

With an unpredictable seed a skill can carry answers only by covering a large share of the generator's space, either with a lookup table or with a copy of the generator or its solver; the coverage statistic below measures the first, and Section~\ref{sec:res-leak} discusses the second. The registry estimates that share, $\lambda_{\text{cover}}$, on a Monte Carlo sample of the space and rejects skills above a threshold (Section~\ref{sec:res-leak}).

\subsection{The betting test with a futility margin (R1)}
Let $X_i=(1+D_i)/2\in[0,1]$ and $m_0=(1+\delta)/2$, so $H_0$ of Eq.~\eqref{eq:H} reads $\mathbb{E}[X_i\mid\mathcal F_{i-1}]\le m_0$ for every $i$. The wealth process of \citet{waudbysmith2024betting} is
\begin{equation}
K_n=\prod_{i=1}^n\bigl(1+\lambda_i(X_i-m_0)\bigr),\qquad
\lambda_i=\min\Bigl\{c,\ \sqrt{\tfrac{2\log(1/\alpha)}{\hat v_{i-1}\, i\log(1+i)}}\Bigr\},
\label{eq:wealth}
\end{equation}
with $\hat v_{i-1}$ the predictable plug-in variance ($\hat v_0=1/4$) and cap $c\in\{1/2,3/4\}$. Here $\alpha$ is the exam's own level: under e-LOND and spending it is the level $\alpha_t$ fixed when the exam starts, and the bet is tuned to it. The deployed gate uses instead the level-tuned grid mixture of Theorem~\ref{thm:adaptive}, a mixture of such processes, which Proposition~\ref{prop:valid} also covers. The exam admits at the first $n$ with $K_n\ge1/\alpha$. For futility it runs a second, downward bet $K^{\mathrm f}_n=\prod_i(1-\lambda^{\mathrm f}_i(X_i-m_{\mathrm f}))$ against $m_{\mathrm f}=(1+\delta_{\mathrm f})/2$ with $\delta_{\mathrm f}\ge\delta$, and rejects when $K^{\mathrm f}_n\ge1/\alpha_{\mathrm f}$; otherwise it rejects at $N_{\max}$.

\begin{proposition}[Validity]\label{prop:valid}
If $\mathbb{E}[X_i\mid\mathcal F_{i-1}]\le m_0$ for every $i$, then for any stopping rule, including the futility rule and any futility margin $\delta_{\mathrm f}$, $\Pr(\text{admit})\le\Pr(\sup_n K_n\ge1/\alpha)\le\alpha$, and the stopped wealth $K_\tau$ is an e-value: $\mathbb{E}[K_\tau]\le1$.
\end{proposition}
The proof (Appendix~\ref{app:proofs}) is Ville's inequality for the nonnegative supermartingale $K_n$. Two remarks matter in practice. First, the condition is on conditional means, not on identically distributed variants, so drift in a provider's model during an exam does not break validity as long as the arms run concurrently and the skill's conditional lift stays at or below $\delta$; a comparator taken from an earlier run does break it (Section~\ref{sec:res-validity}). Second, the futility margin is free in type-I error: it only removes admissions. Testing futility at $\delta_{\mathrm f}=\delta$, the obvious default, almost never fires when $\Delta=\delta$ exactly, so useless skills run to $N_{\max}$; we use $\delta_{\mathrm f}=\PFutMFrac$ (\PFutM{} percentage points), half the design lift. The simpler cap $\lambda_i\le m_0$ is close to $c=1/2$; we report it as a separate variant.

For pass/fail verifiers with $r=1$ and $\delta=0$, Wald's SPRT on discordant pairs tests $p=\Pr(D_i=1\mid D_i\ne0)\le1/2$; its likelihood ratio is also a test supermartingale, so it satisfies Proposition~\ref{prop:valid} and gives an e-value. It cannot express a margin $\delta>0$ or graded scores, and it needs a pre-specified alternative.

\subsection{Library and resubmissions (R2, R3)}
A natural design stops each exam at $K_n\ge1/\alpha$ and then applies e-BH \citep{wang2022ebh} to the $m$ exams of an admission window. This fails.
\begin{remark}[Stopped e-values are too small for e-BH]\label{prop:ebh}
If each exam stops at the first $n$ with $K_n\ge1/\alpha$ and every bet satisfies $\lambda_i\le\bar\lambda<1/m_0$, then $K_\tau<(1+\bar\lambda(1-m_0))/\alpha$. e-BH at level $\alpha$ over $m$ such exams admits nothing unless more than $m/(1+\bar\lambda(1-m_0))$ of them crossed. At $\delta=0$ this is $80\%$ of the window for PrPl with $c=1/2$ ($\bar\lambda=1/2$), $73\%$ for $c=3/4$, and $67\%$ for aGRAPA with $c=1/2$ ($\bar\lambda=c/m_0=1$). Every admissible bet has $\bar\lambda\le1/m_0$, and $K_\tau<1/(m_0\alpha)$ holds even at equality, so at least a fraction $m_0$ of the window must cross, half of it at $\delta=0$.
\end{remark}
\citet{wang2022ebh} already observe that the all-or-nothing e-value $\mathbf 1\{p\le\alpha\}/\alpha$ carries little information for later analyses, unlike a generic e-value; the remark makes explicit what this means for e-BH over exams stopped at $1/\alpha$. Valid alternatives either stop at the strictest e-BH level, $m/\alpha$, or run every exam to $N_{\max}$ before applying e-BH; both cost variants. Stopped e-BH \citep{wang2025stopped} and e-BH on suprema of e-processes \citep{tavyrikov2025carefree} are principled versions of these fixes. Exams in a registry arrive over time, which suits online control. e-LOND \citep{xu2024online} assigns exam $t$ the level $\alpha_t=\alpha\gamma_t(R_{t-1}+1)$, where $R_{t-1}$ counts earlier admissions and $\sum_t\gamma_t\le1$; the exam then stops at $K_n\ge1/\alpha_t$. FDR stays below $\alpha$ under arbitrary dependence among valid e-values; here that means any dependence between exams that leaves each null exam's conditional null intact given the registry's history when it starts. The threshold is known when the exam starts, so it composes with sequential stopping. Using only admissions known at the start of a window keeps $\alpha_t$ predictable.

\paragraph{The unit of error.} e-LOND counts exams and admissions; R2 is stated for admitted skills. They agree under one convention, which the gate and all our simulations use. An exam tests one version, for one agent, against one comparator, and ends with at most one admission; every commitment consumes an e-LOND index and an attempt index of its lineage, whether or not the exam finishes, so an author cannot abort and restart at the same level. A lineage is admitted at most once against a given comparator: once a version is admitted, later versions of the same lineage are replacements, tested against the admitted version as comparator $s_0$, and their exams continue the lineage's attempt count. A retry of the same version is a new exam. Exams that overlap in time are separate hypotheses whose levels interact only through e-LOND. Each admission therefore lists exactly one (version, agent, comparator) triple, and the FDR over exams is the FDR over admitted triples. A skill admitted for several agents counts once per agent; a registry that wants FDR over distinct skills must count admissions by skill in $R_{t-1}$ and test each skill once per comparator. For example, suppose version~1 of a skill is useless and stops for futility, version~2 is helpful and is admitted, and version~3 is an edit that adds nothing. Version~3 is tested against version~2; its null is true, so admitting it would be a false discovery, counted once in the FDR and once against the lineage's budget. E22 checks that the gate, the proof and the simulations count the same unit (Appendix~\ref{app:more}).

FDR control does not stop a determined author: each resubmission of a useless skill is a fresh exam with a fresh level. We therefore spend alpha over a skill's lineage.
\begin{proposition}[Lineage alpha-spending]\label{prop:lineage}
If the $k$-th exam of a lineage runs at level $\alpha_k^{\text{lin}}=\min(\alpha_t,\ \alpha\gamma^{\text{lin}}_k)$, fixed when it starts, with $\sum_k\gamma^{\text{lin}}_k\le1$, then the probability that any version with $\Delta\le\delta$ is admitted is at most $\alpha$, whatever helpful versions the lineage also contains and however the author chooses versions from earlier results, and the e-LOND FDR guarantee still holds.
\end{proposition}
The rule needs a lineage definition: the registry links resubmissions by author key, task class and skill similarity. Linking is a policy choice outside the statistics; an author who evades linking with new identities is a sybil problem that stake or identity verification must address.

\subsection{Dual-arm credential and verification (R4)}
Each arm's transcripts are hashed into a Merkle tree, and a signed credential binds the skill digest, the commitment and reveal, $\theta$, both roots and the statistics $(N,\hat\Delta,\mathrm{LCB},\text{decision})$ (Appendix~\ref{app:schema}). Algorithm~\ref{alg:verify} (Appendix~\ref{app:schema}) is what a verifier runs. It re-executes only the deterministic verifiers on recorded final states, not the agent, so it is cheap (Section~\ref{sec:res-leak}). It cannot show that the recorded behaviour came from the stated model; that needs attested capture, which we treat as an extension.

\paragraph{What the credential should report.} A credential issued at the stopping time reports the estimate $\hat\Delta=\bar D_N$, which is biased upward because exams stop when the estimate is high, and the anytime-valid lower bound obtained by inverting Eq.~\eqref{eq:wealth} over $m_0$, which keeps its stated time-uniform coverage under the conditions of Proposition~\ref{prop:valid} but sits just above $\delta$ at admission (Section~\ref{sec:res-report}). The credential should lead with the bound; a registry that wants an informative effect size can continue the exam after admission at no cost to validity.

%% file: figures/F1_protocol.tex
% F1: admission protocol (TikZ, included by paper/main.tex)
\begin{tikzpicture}[font=\small, node distance=4mm and 6mm,
  box/.style={draw, rounded corners=2pt, align=center, minimum height=9mm, inner sep=3pt, fill=#1},
  box/.default=white, ar/.style={-{Stealth[length=2mm]}, thick}]
\definecolor{okblue}{HTML}{0072B2}\definecolor{okorange}{HTML}{E69F00}\definecolor{okgreen}{HTML}{009E73}
\node[box=okblue!12] (reg) {Register\\$H(s)$, $\mathrm{id}(a)$ at $b_0$};
\node[box=okblue!12, right=of reg] (com) {Commit $c=H(\sigma_p\|\rho\|\theta)$\\at $b_1$, beacon height $b_2$};
\node[box=okblue!12, right=of com] (seed) {Seed\\$\sigma=H(\sigma_p\|\beta_{b_2})$};
\node[box=okorange!18, below=7mm of seed] (exam) {Paired exam on $G(\sigma,i)$\\arm 0: $s_0$, arm 1: $s$};
\node[box=okorange!18, left=of exam] (gate) {Betting test, futility margin;\\level $\alpha_t$ from e-LOND\\and lineage spending};
\node[box=okgreen!15, left=of gate] (cred) {Credential $\chi$: both\\Merkle roots, $\theta$, decision;\\reveal $(\sigma_p,\rho)$ at $b_3$};
\node[box=okgreen!15, below=7mm of gate] (ver) {Any verifier: Algorithm~\ref{alg:verify}\\recompute decision from transcripts};
\node[box=gray!12, below=7mm of cred] (log) {Append-only log\\+ public beacon};
\draw[ar] (reg) -- (com); \draw[ar] (com) -- (seed); \draw[ar] (seed) -- (exam);
\draw[ar] (exam) -- node[above, font=\scriptsize]{$D_i$} (gate); \draw[ar] (gate) -- node[above, font=\scriptsize]{stop} (cred);
\draw[ar] (cred) -- (log); \draw[ar] (log) -- (ver); \draw[ar, dashed] (ver.east) -| (exam.south);
\end{tikzpicture}

%% file: sections/theory.tex
\section{The price of resubmission-proof admission}\label{sec:theory}
Proposition~\ref{prop:lineage} shows that spending $\alpha$ over a lineage is sufficient. This section asks what that protection costs an honest author. Repeated submission until success is a known threat to evaluation: it motivates reliable leaderboards and the reusable holdout \citep{blum2015ladder,dwork2015reusable}, it has been documented in public LLM leaderboards \citep{singh2025leaderboard}, and strategic submitters are studied in principal-agent testing \citep{bates2022principal}. Two ingredients are known. Some form of spending is necessary (Proposition~\ref{thm:lineage}, an elementary observation alongside classical alpha-spending, e.g.\ \citealp{tian2021online}). One exam at level $a$ needs about $\log(1/a)/K_{\inf}$ variants, where $K_{\inf}$ is the information each variant carries against the closest useless skill (Proposition~\ref{thm:onexam}, from \citealp{agrawal2025stopping,honda2010asymptotically,larsson2025numeraire}). Our contribution is Theorem~\ref{thm:price}: combining the two, the extra cost of resubmission-proofness is the entropy of the honest attempt-count distribution divided by $K_{\inf}$, no lineage-valid gate whose exams are fixed in advance can charge less, and a mismatched spending schedule adds a Kullback-Leibler term in both the upper and the lower bound. Choosing the schedule is then a coding problem: Proposition~\ref{prop:minimax} gives the minimax schedule over a class of attempt-count laws, Theorem~\ref{thm:learned} a schedule learned from earlier lineages, and Theorem~\ref{thm:adaptive} shows that the price holds, to first order, for a gate that does not know the skill's effect. Proofs are in Appendix~\ref{app:theory}.

\paragraph{Scope, in brief.} Three points frame what follows. (i) The lower bounds concern gates whose exam $k$ is fixed before the lineage starts; we claim no lower bound for gates that carry evidence or levels from one version to the next, or across lineages. (ii) The cost $C$ counts the variants of the exam that admits the helpful version; the useless versions are stopped by the futility bet at its own fixed level, so the schedule controls only the entropy term of the total. (iii) The deployed level $\min(\alpha_t,\alpha\gamma^{\text{lin}}_k)$ composes e-LOND across the library with spending within a lineage (Proposition~\ref{prop:lineage}); the theorems price the lineage term. Each point is made precise in ``What the theorem covers'' below.

\paragraph{Setting.} Write $X_i=(1+D_i)/2\in[0,1]$ and $m_0=(1+\delta)/2$, as in Section~\ref{sec:gate}. An author submits versions $k=1,2,\dots$ of a skill, and version $k$ may depend on the whole history $\mathcal H_{k-1}$: earlier transcripts, decisions and the author's own randomness. Exam $k$ draws its variants after its commitment, so they are fresh given $\mathcal H_{k-1}$.
\begin{definition}[Conditional null across a lineage]\label{def:condnull}
A lineage is null if, for every $k$, conditional on $\mathcal H_{k-1}$, on version $k$ and on the registry's history $\mathcal G_{k-1}$ when exam $k$ starts (which contains $\mathcal H_{k-1}$ and fixes the deployed level), the scores of exam $k$ satisfy $\mathbb E[X_i\mid\mathcal F_{i-1},\mathcal G_{k-1}]\le m_0$.
\end{definition}
This is the assumption that Propositions~\ref{prop:valid} and~\ref{prop:lineage} use. It fails if an author can see exam $k$'s hidden variants before submitting version $k$, which is what the commitment before the beacon prevents. For a law $P$ on $[0,1]$ with mean above $m_0$, let
\begin{align*}
K_{\inf}(P,m_0)&=\inf\{\mathrm{KL}(P\,\|\,Q):\ Q \text{ a law on }[0,1],\ \mathbb E_QX\le m_0\},\\
g_P(\lambda)&=\mathbb E_P\log(1+\lambda(X-m_0)).
\end{align*}

\begin{proposition}[Lineage control: sufficient and tight]\label{thm:lineage}
\emph{(a)} If exam $k$ runs at a level $a_k$ fixed in advance and is valid under Definition~\ref{def:condnull}, then every adaptive author with a null lineage is ever admitted with probability at most $1-\prod_k(1-a_k)\le\sum_ka_k$.
\emph{(b)} Consider a gate whose exams are fixed in advance: the level, bet and stopping rule of exam $k$ do not depend on earlier exams' outcomes, and each exam draws its variants independently of earlier exams. Fix a null law $Q$ and let $\sigma_k(Q)$ be the probability that exam $k$ admits a version whose scores are i.i.d.\ $Q$. The author who resubmits such a version until it is admitted succeeds with probability $1-\prod_k(1-\sigma_k(Q))$. A gate that controls lineage error at $\alpha$ therefore needs $\sum_k\sigma_k(Q)\le\bar\alpha:=\log\frac{1}{1-\alpha}$ for every null $Q$. In particular, any gate whose exams all have size at least $\sigma>0$ under some null law admits a resubmitted useless skill with probability $1-(1-\sigma)^K\to1$.
\end{proposition}
Part (a) is the classical alpha-spending argument \citep{tian2021online} stated under Definition~\ref{def:condnull}. Part (b) is elementary; we state it because it shows that per-exam sizes must be summable, so no gate that leaves each exam at a fixed level can be resubmission-proof. The \NResubPerExam\% failure after \PResub{} resubmissions in Section~\ref{sec:res-library} is this effect: it equals a constant per-exam size of \NResubImpliedSize\%.

\begin{proposition}[Cost of one exam]\label{thm:onexam}
Let $X_1,X_2,\dots$ be i.i.d.\ $P$ with $\mathbb E_PX>m_0$.
\emph{(a)} Any exam that admits with probability at most $s$ under every i.i.d.\ null law and with probability at least $1-\beta$ under $P$ has $\mathbb E_P[\tau]\ge\{(1-\beta)\log(1/s)-\log2\}/K_{\inf}(P,m_0)$.
\emph{(b)} For any $\lambda\in[0,1/m_0]$ with $g_P(\lambda)>0$, the exam that stops when $\prod_{i\le n}(1+\lambda(X_i-m_0))\ge1/a$ is valid at level $a$, has power one, and has $\mathbb E_P[\tau]\le\{\log(1/a)+\log(1/m_0)\}/g_P(\lambda)$.
\emph{(c)} $\sup_{0\le\lambda<1/m_0}g_P(\lambda)=K_{\inf}(P,m_0)$. The supremum is attained by a least-favourable null law $Q^*$ with $Q^*(x)\propto P(x)/(1+\lambda^*(x-m_0))$ on the support of $P$, plus any leftover mass at $0$.
\end{proposition}
None of this is new in substance. For power-one tests, \citet[Theorem~3.1]{agrawal2025stopping} prove the lower bound $\log(1/\alpha)/\mathrm{KL}_{\inf}$ for general composite nulls, with asymptotically matching upper bounds, and \citet{waudbysmith2025instance} prove matching asymptotic bounds for portfolio e-processes, bounded means included. Part (c) is the bounded-support duality of \citet{honda2010asymptotically} \citep[see also][Proposition~1]{honda2015nonasymptotic}, mirrored for a null mean bounded above, and $Q^*$ is the reverse information projection of \citet{larsson2025numeraire}. Part (a) uses the change-of-measure inequality of \citet[Lemma~1]{kaufmann2016complexity} \citep[see also][]{garivier2019explore}. We state the bounded case with explicit finite-level constants, allowing power $1-\beta<1$ as Theorem~\ref{thm:price} needs, and give a short proof that exhibits $Q^*$. Together the parts say that the cheapest valid exam at level $a$ needs $\log(1/a)/K_{\inf}$ variants, up to an additive $\log(2/m_0)/K_{\inf}$, and that the fixed Kelly bet $\lambda^*=\arg\max_\lambda g_P(\lambda)$ attains this \citep[compare][]{grunwald2024safe,shekhar2024nonparametric}.

\begin{theorem}[Price of resubmission-proofness]\label{thm:price}
An honest author's versions $1,\dots,J-1$ are useless and version $J$ has i.i.d.\ scores from $P$, where the attempt index $J$ has law $\pi$ and each event $\{J=k\}$ is fixed before the hidden variants of exam $k$ are revealed; it may depend on the outcomes of earlier exams. Let $C=\mathbb E[\tau_J]$ be the expected number of variants spent on the helpful version's exam.
\emph{(a)} Suppose a gate's exam $k$ depends only on $k$, the gate controls lineage error at $\alpha$, and every exam has power at least $1-\beta\ge1/2$ against $P$. Let $\bar\alpha q_k$ be the probability that exam $k$ admits a version with i.i.d.\ scores from the least-favourable law $Q^*$ of Proposition~\ref{thm:onexam}(c), so that $\sum_kq_k\le1$ by Proposition~\ref{thm:lineage}(b). Then
\[
C\ \ge\ \frac{(1-\beta)\{\log(1/\bar\alpha)+H(\pi)+\mathrm{KL}(\pi\,\|\,q)\}-h(\beta)}{K_{\inf}(P,m_0)},
\]
where $h(\beta)=-\beta\log\beta-(1-\beta)\log(1-\beta)$ and $\mathrm{KL}(\pi\|q)=\sum_k\pi_k\log(\pi_k/q_k)\ge0$.
\emph{(b)} Spending $a_k=\alpha q_k$ with $\sum_kq_k\le1$ and the bet of Proposition~\ref{thm:onexam}(b) gives a lineage-valid gate with
\[
C\ \le\ \frac{\log(1/\alpha)+H(\pi)+\mathrm{KL}(\pi\,\|\,q)+\log(1/m_0)}{g_P(\lambda)}.
\]
\end{theorem}
Compared with an exam that ignores resubmission, the honest author pays $H(\pi)/K_{\inf}$ extra variants, and among lineage-valid gates whose exams are fixed in advance and have power one, none can charge less, up to the additive constants of the theorem. With power $1-\beta<1$ the lower bound loses the factor $1-\beta$ and the term $h(\beta)$, so the gap between the two parts grows by $\beta\{\log(1/\bar\alpha)+H(\pi)+\mathrm{KL}(\pi\|q)\}+h(\beta)$, which grows with $H(\pi)$. The two parts meet for the gate of part (b) with the Kelly bet $\lambda^*=\arg\max_\lambda g_P(\lambda)$, including the boundary case $\lambda^*=1/m_0$, which occurs for \NThAllIn\% of the helpful SkillsBench skills (E11). There a variant can shrink the wealth by an arbitrary factor, but the overshoot at the crossing is still at most $\log(1/m_0)$, so Proposition~\ref{thm:onexam}(b) holds at $\lambda=1/m_0$ (Appendix~\ref{app:theory}). Its exams have power one, so $\beta=h(\beta)=0$, and exam $k$ admits $Q^*$ with probability at most $\alpha q_k$. Hence
\begin{equation}\label{eq:sandwich}
\log\frac1\alpha+H(\pi)+\mathrm{KL}(\pi\,\|\,q)\ \le\ K_{\inf}\,C\ \le\ \log\frac1\alpha+H(\pi)+\mathrm{KL}(\pi\,\|\,q)+\log\frac1{m_0}.
\end{equation}
The schedule therefore changes the honest author's cost only through $\mathrm{KL}(\pi\|q)/K_{\inf}$, to within $\log(1/m_0)/K_{\inf}$ variants that depend on neither $q$ nor $\pi$. Reading a spending sequence as a prior is old: Occam-style bounds spend $\delta P(c)$ on hypothesis $c$ with $P$ called a prior \citep[Theorem~4.1]{langford2005tutorial}, the spending functions of \citet{howard2021timeuniform} correspond to mixture densities, and weighted multiple testing uses prior weights \citep{genovese2006weighting,roeder2009genome}. Optimal alpha spending for the expected time to signal has also been derived for group-sequential binomial designs \citep{silva2020optimal}. To our knowledge, the characterization of the honest author's expected sample cost by the entropy of the attempt-count law, with a matching lower bound over lineage-valid gates whose exams are fixed in advance, is new.

\paragraph{What the theorem covers.} Three restrictions matter. First, the lower bound concerns gates whose exam $k$ is fixed before the lineage starts. Gates that carry evidence or levels from one version to the next, or across lineages through e-LOND, are outside it (Appendix~\ref{app:theory}, Scope). Second, the author may react to earlier exams, for example by submitting the helpful version sooner after a near miss, but cannot see an exam's variants before choosing what to submit to it; commit-reveal enforces exactly this. The bound then holds with $\pi$ the marginal law of $J$. The condition cannot be dropped: in simulation (E16, \NAdSkills{} SkillsBench skills) both bounds hold for authors who react to past exams, with or without the $N_{\max}$ cap, while an author who could see the variants of the first three exams and submit to the fastest one spends up to \NAdClairBelowMax\% fewer variants than the lower bound allows. Third, $C$ counts only the helpful version's exam. The useless versions also spend variants, but each of their exams is stopped by the futility bet at its own fixed level $\alpha_{\mathrm f}$ (Section~\ref{sec:gate}), so its cost does not depend on the spending schedule except through an admission probability of at most $\alpha q_k$. The entropy term is therefore the part of the cost that the schedule controls. The gate of part (b) itself has neither cap nor futility bet, so under a useless version its exam may never end; the deployed gate adds both. Futility also runs in the helpful exam: it fires with probability at most $\alpha_{\mathrm f}$ when $\Delta\ge\delta_{\mathrm f}$, but with high probability when $\delta<\Delta<\delta_{\mathrm f}$, which is one more source of $\beta>0$. We take exam $J$ to run whatever happened before; if a useless version is admitted first, which has probability at most $\alpha$, $C$ is defined on the event that exam $J$ runs. With a cap, $C$ counts only the admitting exam, not missed helpful exams that ran to $N_{\max}$. Finally, if $J$ reacts to earlier outcomes its law may depend on the gate's levels; comparisons between gates at the same $\pi$ assume it does not, as when the author reacts only to futility outcomes. In live studies 2 and 3 with Haiku each placebo and the leak skill used 186 to 209 variants before futility, against 28 and 52 for the helpful skills, so the schedule governs a small share of a lineage's total cost when most versions are useless. Two further limits concern the gate we evaluate. Power at least $1-\beta$ in every exam is a property of the untruncated exam; a cap $N_{\max}$ lowers power, and the bound applies to exams that reach the stated power before the cap. And the deployed level is $\min(\alpha_t,\alpha\gamma^{\text{lin}}_k)$: over long histories the e-LOND term sets it in \NEbElondBinds\% of first attempts (E3b), and it set every level in the live resubmission study. The theorem prices the lineage term, which binds in short histories, for early lineages and after many admissions. E19 measures the whole cost of a lineage under the deployed rule: once a lineage starts 50 or more exams into the registry, the e-LOND term sets the level in all but at most \NLcBindMax\% of exams, the choice of lineage schedule changes the helpful exam's cost by under \NLcSchedSpread\%, and the useless versions cost more than the helpful one (Appendix~\ref{app:more}). The entropy price is therefore the price of the lineage guarantee taken alone. In a long-running registry with online FDR control, resubmission protection is nearly free, and the $\log(1/\text{level})$ form of the theorem, applied to the deployed level, predicts the cost of a helpful exam to within \NLcPredErrMean\% on average; the price itself is what a new registry, or one that controls each lineage separately, pays. The simulations and reanalyses in Section~\ref{sec:results} use the grid bet, which the theory assumes and which we recommend for deployment (E13); the PrPl bet with cap $1/2$, common in earlier work, is reported beside it. The grid bet is faster at moderate and large lifts but has less power at small ones (Appendix~\ref{app:more}). Two parts still use PrPl: live studies 1 to 3, which committed it before we adopted the grid bet (studies 4 and 5 committed the grid bet), and the estimation check (E4), whose confidence bound is built from PrPl bets.

\begin{corollary}[No schedule is cheap for every attempt]\label{cor:universal}
For any $q$ with $\sum_kq_k\le1$, $\log(1/q_k)>\log k+\log\log k$ for infinitely many $k$. The schedule $q_k=c_u/\{(k+1)\log^2(k+1)\}$, with $c_u=1/\sum_{k\ge1}\{(k+1)\log^2(k+1)\}^{-1}=\NCuValue$, pays $\log k+2\log\log k+O(1)$.
\end{corollary}
The corollary is classical \citep{elias1975universal,rissanen1983universal}, and \citet{foster2013risk} already use this universal shape to set alpha-investing rules; we include it to translate it into variants.
When nothing is known about $\pi$, some attempt index always costs about $(\log k+\log\log k)/K_{\inf}$ extra variants. The schedule used in this paper, $q_k=k^{-1.6}/\zeta(1.6)$, has the shape of the spending sequences of \citet{howard2021timeuniform} and pays $1.6\log k+0.83$ nats. That is more than the universal schedule for large $k$, but less for every $k$ from 2 to 20; only the first attempt is cheaper under the universal schedule (0.71 against 0.83 nats).

\paragraph{Choosing the schedule.} By Eq.~\eqref{eq:sandwich}, a registry that does not know $\pi$ faces a coding problem: each lineage pays $\mathrm{KL}(\pi\|q)/K_{\inf}$ variants above the entropy floor. The registry may know only a class $\Pi$ that contains $\pi$, or it may learn $\pi$ from lineages that have already closed. For a class $\Pi$ of laws on the attempt indices, let
\[
R(\Pi)=\inf_q\ \sup_{\pi\in\Pi}\ \mathrm{KL}(\pi\,\|\,q),
\]
with the infimum over probability laws $q$. Write $\Pi_N$ for all laws on $\{1,\dots,N\}$, $\Pi_N^{\downarrow}$ for the non-increasing ones, and $C(\pi)$ for the cost of Theorem~\ref{thm:price} when $J\sim\pi$.

\begin{proposition}[Minimax schedule]\label{prop:minimax}
\emph{(a)} Every gate as in Theorem~\ref{thm:price}(a) has
\[\sup_{\pi\in\Pi}\big\{K_{\inf}C(\pi)-(1-\beta)[\log(1/\bar\alpha)+H(\pi)]\big\}\ \ge\ (1-\beta)R(\Pi)-h(\beta).\]
\emph{(b)} If $\sup_{\pi\in\Pi}\mathrm{KL}(\pi\|q)\le R(\Pi)+\epsilon$, the gate of Theorem~\ref{thm:price}(b) with the Kelly bet $\lambda^*$ has
\[\sup_{\pi\in\Pi}\big\{K_{\inf}C(\pi)-\log(1/\alpha)-H(\pi)\big\}\ \le\ R(\Pi)+\epsilon+\log(1/m_0).\]
\emph{(c)} For any $\pi_1,\dots,\pi_M\in\Pi$ and weights $w$, $R(\Pi)\ge\sum_jw_j\,\mathrm{KL}(\pi_j\|\bar\pi_w)$, where $\bar\pi_w=\sum_jw_j\pi_j$.
\emph{(d)} $R(\Pi_N)=\log N$, attained by the uniform schedule, and $R(\Pi_N^{\downarrow})\le\log H_N$ with $H_N=\sum_{k\le N}1/k$, the worst-case excess of $q_k=1/(kH_N)$.
\end{proposition}
Part (c) is the easy half of the redundancy-capacity theorem, by which $R(\Pi)$ equals the capacity $\sup_w\sum_jw_j\mathrm{KL}(\pi_j\|\bar\pi_w)$ of the channel $j\mapsto\pi_j$ under general conditions, with the capacity-achieving mixture as the minimax schedule \citep{haussler1997general,merhav1995strong}. Parts (a) and (b) say that this capacity, divided by $K_{\inf}$, is the price in variants of knowing only the class of $\pi$. Over all laws on the positive integers $R=\infty$, which is Corollary~\ref{cor:universal} again; a finite price needs a restricted class, as in coding on countable alphabets \citep{boucheron2009coding}. For natural classes the price is small (Appendix~\ref{app:more}, E12). If honest authors make at most 20 attempts, the minimax excess is $\log20=\NScLogTwenty$ nats, against \NScZetaWorstTwenty{} nats in the worst case for the deployed $\zeta(1.6)$ schedule. If, in addition, later attempts are never more likely than earlier ones, the excess for $N=1{,}000$ lies in the certified interval $[\NScMonoLoThousand,\NScMonoHiThousand]$ nats, computed with the Blahut-Arimoto algorithm \citep{blahut1972computation,arimoto1972algorithm} on the extreme points of $\Pi_N^{\downarrow}$. That is against $\log H_N=\NScLogHThousand$ for the explicit schedule and $\log N=\NScLogThousand$ without the monotonicity.

\begin{theorem}[Learned schedule]\label{thm:learned}
Close every lineage after at most $N$ attempts. When lineage $i$ opens, the registry fixes
\[
\hat q^{(i)}_k=(1-\epsilon)\,\frac{c_k+1}{m+N}+\frac{\epsilon}{N},\qquad k=1,\dots,N,
\]
where $c_k$ is the number of counted lineages admitted at attempt $k$, and $m=\sum_kc_k$. The counted lineages must be closed when lineage $i$ opens and must be chosen without regard to their attempt indices. Counting every lineage that happens to have closed is not enough, because lineages that need fewer attempts close sooner and would be over-represented. The rule used below counts lineages $1,\dots,i-1-D$, which is valid when every lineage closes before $D$ further lineages open. Exam $k$ of lineage $i$ runs at level $\alpha\hat q^{(i)}_k$.
\emph{(a)} If lineage $i$ is null in the sense of Definition~\ref{def:condnull}, it is ever admitted with probability at most $\alpha$, whatever the other lineages do, including lineages run by the same author to move the counts.
\emph{(b)} Suppose that $m_h$ of the counted lineages are honest, with attempt indices i.i.d.\ $\pi\in\Pi_N$, and that at most $r$ other counts are arbitrary and may depend on them. Then
\[
\mathbb E\,\mathrm{KL}(\pi\,\|\,\hat q^{(i)})\ \le\ \log\frac1{1-\epsilon}+\log\Big(1+\frac{N-1+r}{m_h+1}\Big),
\]
and $\mathrm{KL}(\pi\|\hat q^{(i)})\le\log(N/\epsilon)-H(\pi)$ for every value of the counts.
\emph{(c)} Suppose lineages open one after another, all are honest and recorded at their attempt index, every lineage closes before $D$ further lineages open, and lineage $i$ counts lineages $1,\dots,i-1-D$. Then $\sum_{i\le n}\mathbb E\,\mathrm{KL}(\pi\|\hat q^{(i)})\le n\log\frac1{1-\epsilon}+(D+1)\log N+(N-1)\log n$. Conversely, for any rule that fixes the schedule of lineage $i$ from the attempt indices of lineages $1,\dots,i-1$,
\[\sup_{\pi\in\Pi_N}\sum_{i\le n}\mathbb E\,\mathrm{KL}(\pi\,\|\,q^{(i)})\ \ge\ R_n(N)=\frac{N-1}2\log\frac n{2\pi e}+\log\frac{\Gamma(1/2)^N}{\Gamma(N/2)}+o(1),\]
the minimax redundancy of $n$ i.i.d.\ draws from $\Pi_N$ \citep{xie1997minimax}.
\end{theorem}
\emph{Capped exams.} With a cap $N_{\max}$, a helpful version can miss admission; the lineage is then admitted at a later attempt or closes unadmitted. We define $c_k$ as the number of counted lineages \emph{admitted} at attempt $k$, so $\hat q$ learns $\tilde\pi$, the law of the admitting attempt among admitted lineages. This is also the law that matters for cost, because Theorem~\ref{thm:price} charges the exam that admits. Part (a) does not use the counts and is unchanged, and parts (b) and (c) hold with $\tilde\pi$ in place of $\pi$ (Appendix~\ref{app:theory}). If each helpful exam at attempts $k\le N$ admits before the cap with probability at least $1-\beta_{\max}$, then, with $\varepsilon=\beta_{\max}+\alpha$, $\tilde\pi$ and $\pi$ differ in total variation by at most $\varepsilon+\varepsilon/(1-\varepsilon)$. A uniform $\beta_{\max}$ is only plausible because lineages close after $N$ attempts: wealth after $N_{\max}$ variants is at most $m_0^{-N_{\max}}$, so no exam admits at a level below $m_0^{N_{\max}}$, and with unlimited attempts $q_k\to0$ would make late helpful exams powerless.

Learning the schedule costs nothing in validity. A null author who floods the registry with lineages of their own can raise the honest authors' cost, by at most $\log(N/\epsilon)$ nats per lineage, but cannot raise any lineage's error. Learning levels from earlier results while keeping validity is also the idea of contextual online FDR control \citep{chen2020contextual}; what Theorem~\ref{thm:learned} adds is the price. The excess per lineage falls like $(N-1)/m$, and with $\epsilon=1/n$ the cumulative excess is, as $n\to\infty$ with $N$ fixed, within a factor of two of the minimax redundancy, so no rule learns much faster. The factor is asymptotic: at $N=20$ and $n=400$ the bound is about four times $R_n(N)$. It comes from the add-one estimator; an add-one-half (Krichevsky-Trofimov) estimator would match the leading term. The cost converse assumes the same $P$, and so the same $K_{\inf}$, in every lineage. Lineages that overlap in time cost only the additive $(D+1)\log N$, provided the registry waits for a fixed number of later openings rather than counting whichever lineages have closed. In E12b, with one attempt per time step and $N=20$, the naive rule over-represents short lineages: with uniform $\pi$ it puts mass \NObUnifQoneNaive{} on the first attempt after 25 lineages, against \NObUnifQoneDelay{} for the fixed-delay rule and $\pi_1=\NObUnifPiOne$. Its cumulative excess after 400 lineages is higher when long lineages are common (\NObIncNaive{} against \NObIncDelay{} nats for $\pi_k\propto k$) and lower when short ones dominate (\NObGeoNaive{} against \NObGeoDelay{} for the geometric law), because it uses more lineages. Only the fixed-delay rule carries the guarantee, and it stays far below the bound of part (c) (\NObBound{} nats). With power-one exams, Theorem~\ref{thm:price}(a) applies to each lineage given the earlier ones with $\beta=0$, so the converse holds for the cost itself: some $\pi\in\Pi_N$ gives $\sum_{i\le n}\{K_{\inf}\mathbb E[C_i]-\log(1/\bar\alpha)-H(\pi)\}\ge R_n(N)$. In E12, with $N=20$, $\epsilon=0.05$ and geometric attempt counts with mean 3, the expected excess falls from \NScKLZero{} nats with no closed lineages to \NScKLHundred{} after 100 and \NScKLThousand{} after 1{,}000, below the bound at every $m$. With 50 poisoned counts placed where $\pi$ is smallest, it is \NScKLPoisonHundred{} nats after 100 honest lineages, against the bound \NScBoundPoisonHundred. For the median skill, simulated extra variants match $\mathrm{KL}(\pi\|\hat q)/g_P(\lambda^*)$: \NScVarHundredSim{} against \NScVarHundredTh{} after 100 closed lineages and \NScVarFiveHundredSim{} against \NScVarFiveHundredTh{} after 500. The deployed $\zeta(1.6)$ schedule, renormalized to 20 attempts, costs \NScVarZetaSim{} extra variants for this $\pi$, so the learned schedule overtakes it between 100 and 500 closed lineages.

\paragraph{Not knowing the skill.} Theorem~\ref{thm:price}(b) bets with a $\lambda$ chosen for $P$, which a registry does not know. A mixture of fixed bets removes this need, and spending makes a natural choice of grid available: each exam has its own level, fixed in advance, so the grid can be tuned to it. For an exam at level $a$, let $L=\log(1/a)$, $\rho=L/(L+1)$ and $\lambda_j=\rho^{j+1}/m_0$ for $j=0,1,\dots$, with the weights $w_j=c_u/\{(j+2)\log^2(j+2)\}$ of Corollary~\ref{cor:universal}, which sum to one. The exam admits at the first $n$ with
\[
M_n=\sum_{j\ge0}w_j\prod_{i\le n}\{1+\lambda_j(X_i-m_0)\}\ \ge\ 1/a .
\]
Let $\lambda^*\in(0,1/m_0]$ maximize $g_P$ on $[0,1/m_0]$, so that $g_P(\lambda^*)=K_{\inf}(P,m_0)$, and write $\Lambda=\log\{1/(m_0\lambda^*)\}\ge0$ for how far the Kelly bet lies below the largest admissible bet.
\begin{theorem}[Level-tuned grid]\label{thm:adaptive}
\emph{(a)} The exam does not depend on $P$, is valid at level $a$ under Definition~\ref{def:condnull}, and has power one against every $P$ with $\mathbb E_PX>m_0$.
\emph{(b)} For every such $P$, with $j^*=\lfloor\Lambda/\log(1/\rho)\rfloor$ and $\ell(j)=\log(1/w_j)$,
\[
K_{\inf}(P,m_0)\,\mathbb E_P[\tau]\ \le\ \Bigl(1+\frac1L\Bigr)\Bigl\{L+\log\frac1{m_0}+\ell(j^*)\Bigr\}.
\]
With $u=\log\{(L+1)(2+\Lambda)\}$, this is at most $L+1+(1+1/L)\{\log(L+1)+\log(2+\Lambda)+2\log u+\log(1/m_0)+\log(1/c_u)\}$.
\emph{(c)} Spending $a_k=\alpha q_k$ with this exam at each level gives a lineage-valid gate that does not depend on $P$. With $\bar L=\log(1/\alpha)+H(\pi)+\mathrm{KL}(\pi\|q)$ and $\bar u=\log\{(\bar L+1)(2+\Lambda)\}$,
\[
K_{\inf}C\ \le\ \bar L+1+\Bigl(1+\frac1{\log(1/\alpha)}\Bigr)\Bigl\{\log(\bar L+1)+\log(2+\Lambda)+2\log\bar u+\log\frac1{m_0}+\log\frac1{c_u}\Bigr\},
\]
while every lineage-valid gate whose exam $k$ depends only on $k$ and has power one has $K_{\inf}C\ge\log(1/\bar\alpha)+H(\pi)+\mathrm{KL}(\pi\|q')$, with $q'$ its own size schedule under $Q^*$.
\end{theorem}
In part (c) we replaced $1+1/\bar L$ by $1+1/\log(1/\alpha)$, which is valid because the braced term is nonnegative: $m_0\le1$, $c_u<1$, and each exam's $u_k=\log\{(L_k+1)(2+\Lambda)\}\ge1$ because $L_k\ge\log(1/\alpha)\ge\log2$ for every $\alpha\le0.5$, which covers every level used here. In part (c) the lower bound uses the gate's own schedule $q'$ and the upper bound uses $q$. For the grid gate, exam $k$ admits $Q^*$ with probability at most $\alpha q_k$, so $\bar\alpha q'_k\le\alpha q_k\le\bar\alpha q_k$ and $\mathrm{KL}(\pi\|q')\ge\mathrm{KL}(\pi\|q)$; the two bounds therefore share the leading terms $\log(1/\alpha)+H(\pi)+\mathrm{KL}(\pi\|q)$ up to $\log(\bar\alpha/\alpha)$. So the entropy price survives without knowing the skill: at every level the leading term of $K_{\inf}\mathbb E[\tau]$ is $L$ with coefficient one, and not knowing $P$ costs at most about $\log\bar L+\log\Lambda$ nats. The construction combines known parts, and its first-order behaviour is known. Mixing over a geometric grid whose resolution is tuned to the level is the device of \citet[proof of Theorem~33]{kaufmann2021mixture}, who apply it to grids over time in deviation inequalities; here we move it to the bet. For bounded payoffs, \citet[Theorem~2]{orabona2026optimal} characterizes the expected rejection time of portfolio tests exactly as $\alpha\to0$: with a prior whose log-scale density decays at the integrability border, which needs no knowledge of the level, $\mathbb E_P[\tau]$ is asymptotic to $\{\log(1/\alpha)+\log\log(1/|\lambda^*_P|)\}/K_{\inf}$, with $\lambda^*_P$ the Kelly bet under $P$, uniformly over alternatives, and his Theorem~5 shows that the iterated-logarithm term cannot be removed. Asymptotically his bound is sharper than ours, which pays an extra $\log L$. What Theorem~\ref{thm:adaptive} adds is a bound that holds at every finite level with explicit constants, which a registry needs for exams at levels such as $\alpha q_k$ that are small but fixed, and part (c), which combines it with the spending schedule; to our knowledge that combination is new. For exponential families and in an integrated (Bayes) sense, \citet{pollak1978optimality} shows that the second-order term in $L$ cannot be smaller than $\tfrac12\log L$, so our coefficient of about one on $\log L$ is not optimal. The term in $\Lambda$ has the order of the law of the iterated logarithm, which cannot be avoided when the effect is small \citep{farrell1964asymptotic,agrawal2025stopping,orabona2026optimal}.
In E13 (Appendix~\ref{app:more}) we ran the grid exam on the \NThHelpful{} SkillsBench skills with positive released lift. At the first exam's level $\alpha q_1$ it needs a median of \NAgGridQone{} variants, against \NAgKellyQone{} for the Kelly bet that knows $P$, \NAgAgrapaQone{} for aGRAPA and \NAgPrplQone{} for the PrPl bet with cap $1/2$; the bound of part (b) gives \NAgBoundQone. For the median skill, $K_{\inf}\mathbb E[\tau]-L$ stays between \NAgExcessMin{} and \NAgExcessMax{} nats as $L$ grows from $\log(1/\alpha)$ to $\log(1/\alpha)+16$, against \NAgFixedMin{} to \NAgFixedMax{} nats for the same grid with $\rho=1/2$ fixed. Under each skill's least-favourable null, the largest rejection rate is \NAgMaxReject\% at level \NAgLevelQone\%.

\paragraph{What the results say about the gate.} We checked each statement on the \NThHelpful{} SkillsBench skills with positive released lift, taking each skill's 18 per-configuration differences as $P$ (Appendix~\ref{app:more}, E11).
\begin{itemize}
\item \emph{Duality.} The least-favourable law $Q^*$ of Proposition~\ref{thm:onexam}(c) is a null law with $\mathrm{KL}(P\|Q^*)=K_{\inf}$ for every skill, to numerical precision.
\item \emph{One exam.} No simulated cost falls below the lower bound. Relative to $\log(1/\alpha)/K_{\inf}$, the Kelly bet needs a median $\NThOracleRatio\times$ (at most $\NThOracleRatioMax\times$), aGRAPA $\NThAgrapaRatio\times$, and the PrPl bet with $c=1/2$ (used in live studies 1 to 3) needs $\NThPrplRatio\times$ (at most $\NThPrplRatioMax\times$). For \NThAllIn\% of these skills the Kelly bet is $\lambda^*=1/m_0$, and the cap is what costs the most variants. This is why the deployed gate uses the level-tuned grid bet.
\item \emph{Lineage.} For the median skill ($K_{\inf}=\NThMedKinf$, about \NThMedIdeal{} variants at level $\alpha$), the simulated extra cost of the $k$-th attempt under our schedule matches $\log(q_1/q_k)/g_P(\lambda^*)$: \NThExtraTwoSim{} against \NThExtraTwoTh{} variants at $k=2$, \NThExtraTenSim{} against \NThExtraTenTh{} at $k=10$, and \NThExtraTwentySim{} against \NThExtraTwentyTh{} at $k=20$.
\item \emph{Honest authors.} An author who always succeeds on the first attempt pays \NThPriceOne{} extra variants, because the first exam runs at $\alpha q_1=0.022$. If honest attempt counts are geometric with mean 3, the expected price is \NThPriceGeoThree{} variants against the entropy floor $H(\pi)/K_{\inf}=\NThFloorGeoThree$. If they follow a Zipf law with exponent 2, it is \NThPriceZipf{} against \NThFloorZipf{}. A geometric schedule fits geometric authors better, but costs \NThGeoOnZipfLoss{} more variants under the Zipf law.
\end{itemize}

%% file: sections/evaluation.tex
\section{Evaluation design}
\label{sec:eval}

\paragraph{Released benchmark data.} SWE-Skills-Bench released pass rates with and without each skill for 98 skills, four harness-model configurations and ten requirement batches per skill, \NSweCells{} skill-by-configuration cells in all \citep{han2026sweskillsbench}. We take $D$ as the difference in batch pass rate. \NSweTies\% of the differences are exactly zero and their standard deviation is \NSweSD. SkillsBench released, for each of \NSbTasks{} tasks and 18 configurations, the number of passes in three trials with and without the task's curated skill \citep{li2026skillsbench}; we take $D=(\text{passes with}-\text{passes without})/3$, which has mean \NSbLift{} points, standard deviation \NSbSD{} and \NSbTies\% zeros. The source paper reports a lift of \PCurated{} points and a no-skill pass rate of \PBase\% over its full evaluation; the released registry gives \NSbLift{} points and \NSbPZero\%. The simulation model uses the published figures, and the reanalyses use the registry. The self-generated condition is not in the released registry. We use these data three ways: to apply admission rules to real per-skill evidence (E8, E9), to build a resampled null with the real tie and magnitude structure by randomizing the signs of the SWE-Skills-Bench differences (E8), and to calibrate and check the simulation model.

\paragraph{Simulation model.} Variant $i$ has difficulty $\theta_i\sim N(0,s_\theta^2)$ on the logit scale; the comparator passes with probability $\operatorname{logit}^{-1}(b_0+\theta_i)$ and the skill arm with $\operatorname{logit}^{-1}(b_0+g_i+\theta_i)$, $g_i\sim N(g,s_g^2)$. With probability $\rho$ both arms share one sampling uniform (common random numbers). $b_0$ is set so the comparator pass rate is the SkillsBench no-skill average, \PBase\%, and $g$ is solved by quadrature so that $\Delta$ equals a target exactly. The defaults $s_\theta=\PSdTheta$, $s_g=\PSdLift$ and $\rho=\PRho$ have no published values; we sweep each. At the SkillsBench average lift with $r=3$ the model gives $\operatorname{sd}(D)=\NSbModelSD$ and \NSbModelTies\% zeros, against \NSbSD{} and \NSbTies\% in the released data, so the model is somewhat less noisy than the benchmark and our cost estimates from it are optimistic. To bound the effect, E14 fits $(s_\theta,s_g,\rho)$ at $r=3$ to three within-skill moments of the released differences: the tie share, the standard deviation around each skill's mean, and the share with $|D|\ge2/3$ (\NRvObsTie\%, \NRvObsSD{} and \NRvObsBig\%). The best fit, $s_\theta=\NRvFitSdTheta$, $s_g=\NRvFitSdLift$ and $\rho=\NRvFitRho$, gives \NRvFitTie\%, \NRvFitSD{} and \NRvFitBig\%; it needs far more heterogeneous skill effects and stronger common random numbers than the defaults. E14 also lets a variant's easiness $\theta_i$ and its skill effect $g_i$ correlate, with correlation $\kappa\in\{-0.5,0,0.5\}$, and reruns the cost claims under each model. Section~\ref{sec:res-cost} also resamples the released data directly.

\paragraph{Gates compared.} We compare seven gates. Two are what a careful benchmark would use today: a one-sided paired $t$-test at the fixed $n=\NFixedN$ that the normal approximation $n=\lceil((z_{1-\alpha}+z_{1-\beta})\,\sigma_D/(\Delta_1-\delta))^2\rceil$ gives for $\Delta_1=\PLiftOne$ points at 80\% power (G1), and the same test re-run after every variant from $n=10$ (G2). G3 admits when the lower end of a two-sided 95\% normal interval exceeds $\delta$, checked every ten variants; this is how ACES- or skillcheck-style intervals behave if a registry reads them as a decision rule. The valid alternatives are the discordant-pair SPRT with its alternative set at the design lift (G4), the normal-mixture mSPRT of \citet{johari2022always} (G5), the betting test with the level-tuned grid bet of Theorem~\ref{thm:adaptive} (G7, the deployed gate, with and without a futility margin), and the betting test with the plug-in bet of Eq.~\eqref{eq:wealth} (G6). Unless stated, ``the betting gate'' below is G7. For G6 we vary the cap ($c=\PCap$, $\PCapAlt$, and the simpler $\lambda_i\le m_0$), the bet (predictable plug-in or aGRAPA, \citealp{waudbysmith2024betting}) and the futility margin. Every sequential gate uses $\alpha=\PAlpha$, $\alpha_{\mathrm f}=\PAlphaF$ and $N_{\max}=\PNmax$. Null cells use \PReps{} replicates, which puts the standard error of a 5\% rate at 0.15 points; power cells use \PRepsPower.

\paragraph{Experiments.} E1 measures false admission under nulls; E2 power and expected variants; E3 library-level FDR over \PWindows{} windows of \PWindow{} candidates with \PNullShare\% nulls, and a resubmission attack; E4 the bias of reported lifts; E5 libraries whose lifts follow the published benchmark summaries; E6 leakage checks; E7 credential cost; E8 and E9 the benchmark reanalyses; E20 compares e-LOND with stopped e-BH and e-BH on adjusted suprema on the E3 histories; E24 gives uncertainty on headline numbers (a bootstrap over SkillsBench skills with 10{,}000 resamples, and Monte Carlo intervals for error rates, with the exact one-sided 95\% bound $1-0.05^{1/n}$ when no error occurs in $n$ runs). E10 to E19 and E21 to E25 are described where they are used and in Appendix~\ref{app:more}. Every parameter, with its source or the reason for its value, is in the released \texttt{params.yaml}; every number in Sections~\ref{sec:results} and~\ref{sec:discussion} is generated from the result files.

%% file: sections/results.tex
\section{Results}
\label{sec:results}

\subsection{Released per-skill evidence certifies few skills sequentially}
\label{sec:res-real}

\paragraph{SWE-Skills-Bench (E8).} The source paper's rule (exact two-sided sign-flip test on the ten batch differences, $p<0.05$) marks \NSweGolden{} of the \NSweCells{} skill-by-configuration cells, covering \NSweGoldenSkills{} skills, as golden and \NSweHarmful{} as harmful. None of the one-sided $p$-values survives Benjamini-Hochberg control at 5\% (\NSweBH{} admissions), and no betting test admits any cell. With ten units admission needs a nearly perfect record: the grid bet admits after \NSweGridMinAllOnes{} straight wins, but no cell's wealth passed \NSweObsWealthGrid{} of the 20 needed (Appendix~\ref{app:more}). Pooling a skill's 40 units across configurations, \NSwePooled{} skills pass an uncorrected one-sided test, \NSwePooledBH{} after correction, and \NSwePooledBet{} the grid betting test (\NSwePooledBetPrpl{} with PrPl). The uncorrected calls are not noise: cells flagged on one configuration show a mean lift of \NSweRepLift{} points on the other three, positive for \NSweRepPos\% of them against \NSweAllPos\% of all cells. Admitting a skill that passes on any of the four configurations raises the false-admission rate on the sign-flip null from \NShopPer\% to \NShopAny\%.

\paragraph{SkillsBench (E9).} With 18 paired units per curated skill, a one-sided $t$-test admits \NSbT{} of \NSbTasks{} skills, \NSbTBH{} after Benjamini-Hochberg correction; the grid betting test admits \NSbBet{} (PrPl: \NSbBetPrpl{} with cap $1/2$, \NSbBetHi{} with cap $3/4$), and e-BH on the final grid wealth admits \NSbEBH{} (PrPl: \NSbEBHPrpl). Splitting the configurations into random halves 500 times, skills admitted on one half (\NSbSplitAdm{} on average) show \NSbSplitLift{} points of lift on the other, positive in \NSbSplitPos\% of cases, against \NSbSplitNot{} points for skills not admitted. Benjamini-Hochberg admissions replicate even better (\NSbSplitBHAdm{} on average, \NSbSplitBHLift{} points, \NSbSplitBHPos\% positive). Other valid fixed-sample procedures agree (E10, Appendix~\ref{app:more}). A one-off fixed-sample analysis of a completed benchmark can therefore certify a share of skills. A registry cannot use it as its decision rule, because skills arrive over time, exams are stopped when results look good, and authors resubmit (Section~\ref{sec:res-validity}); the anytime-valid procedure that survives those conditions, e-LOND over skills in arrival order with the grid bet, certifies \NSbELond{} skills on average with 18 units (\NSbELondLo{} to \NSbELondHi{} over 90\% of random orders; Bonferroni at $\alpha/87$: \NSbBonfBet; none with PrPl), against \NSbTBH{} for the one-off analysis. The obstacle is sample size (Section~\ref{sec:res-cost}).

\subsection{Validity under optional stopping}
\label{sec:res-validity}

Table~\ref{tab:false} gives false-admission rates under nulls (Figure~\ref{fig:false} in Appendix~\ref{app:more} plots them). Re-checking a $t$-test after every variant admits up to \NEOnePeekTMax\% of useless skills, and re-checking a 95\% interval every ten variants up to \NEOnePeekCIMax\% (the lowest rates, \NEOnePeekTMin\% and \NEOnePeekCIMin\%, are for a slightly harmful placebo skill not shown in the table); at the null boundary the interval rule runs at \NEOnePeekRatio{} times 5\%, and twice that relative to its one-sided level of 2.5\%; on the resampled SWE-Skills-Bench null the rates are \NEEightNullPeekT\% and \NEEightNullPeekCI\%. The fixed-$n$ test holds its level (\NEOneFixed\% at the boundary). The SPRT, the mSPRT and every betting variant stay within Monte Carlo error of 5\% or below in every null configuration, including the worst case in which every pair is discordant. The largest rate among them is \NEOneValidMax\% (standard error \NEOneValidMaxSE), for the SPRT with common random numbers at $\rho=0.9$ (Appendix~\ref{app:more}); the betting gates stay below 3.5\%.

The last column changes the comparison rather than the test: the comparator arm comes from a stored run and the provider's model has since improved by \PDrift{} logit units, while the skill has no effect. Every gate then admits \NEOneStaleMin\% to \NEOneStaleMax\% of the time, the betting gate included (\NEOneStaleBet\%), because a stale baseline violates the condition of Proposition~\ref{prop:valid}. Running the arms concurrently matters more than the choice of test.

\begin{table}[t]
\centering\small
\caption{False-admission rate (\%) by gate and null configuration (E1, E8). Boundary: $\Delta=\delta=0$ under the default model. Worst case: every pair discordant. Stale baseline: the comparator arm is from an earlier model snapshot. SWE null: the SWE-Skills-Bench differences with random signs. Dashes: the discordant-pair SPRT is defined only for one pass/fail attempt per arm at $\delta=0$, so it is not run with $r=3$, with a margin, or on batch pass rates.}
\label{tab:false}
\resizebox{\linewidth}{!}{\input{tables/t_false_admission.tex}}
\end{table}

\subsection{What validity costs}
\label{sec:res-cost}

\paragraph{Matched power (E18).} Comparing gates at different power mixes cost with power, so E18 tunes each gate to the same power: the fixed-$n$ test by its sample size, the SPRT by $\beta$, and the truncated sequential gates by the smallest $N_{\max}$ that reaches the target (Table~\ref{tab:matched}). At 80\% power the deployed grid gate costs about what a fixed-$n$ test costs when the skill helps (\NMpDesEightyGrid{} against \NMpDesEightyFixed{} variants at the design lift, \NMpCurEightyGrid{} against \NMpCurEightyFixed{} at the curated lift); at 95\% it is cheaper (\NMpDesNinetyFiveGrid{} against \NMpDesNinetyFiveFixed, and \NMpCurNinetyFiveGrid{} against \NMpCurNinetyFiveFixed), because early stopping pays more when the fixed design must cover the tail. The price of anytime validity falls on useless skills: at the 80\% design the grid gate spends \NMpDesEightyGridNull{} variants on a useless skill against \NMpDesEightyFixedNull{} for the fixed design. The discordant-pair SPRT is cheapest where it applies (pass/fail scores, $r=1$, no margin), and PrPl with cap $1/2$ is the most expensive betting rule. The same ordering holds for $r=3$ and for the moment-matched model (Appendix~\ref{app:more}); the tuned grid gate never admitted a useless skill more than \NMpNullMaxGrid\% of the time.
\begin{table}[t]
\centering\small
\caption{Expected variants at matched power (E18; default model, $r=1$, 4{,}000 exams per cell): when the skill has the stated lift, and in parentheses for a useless skill ($\Delta=0$) with the same tuning. The SPRT assumes pass/fail scores and no margin.}
\label{tab:matched}
\input{tables/t_matched_power.tex}
\end{table}

\begin{figure}[t]
\centering
\includegraphics[width=0.78\linewidth]{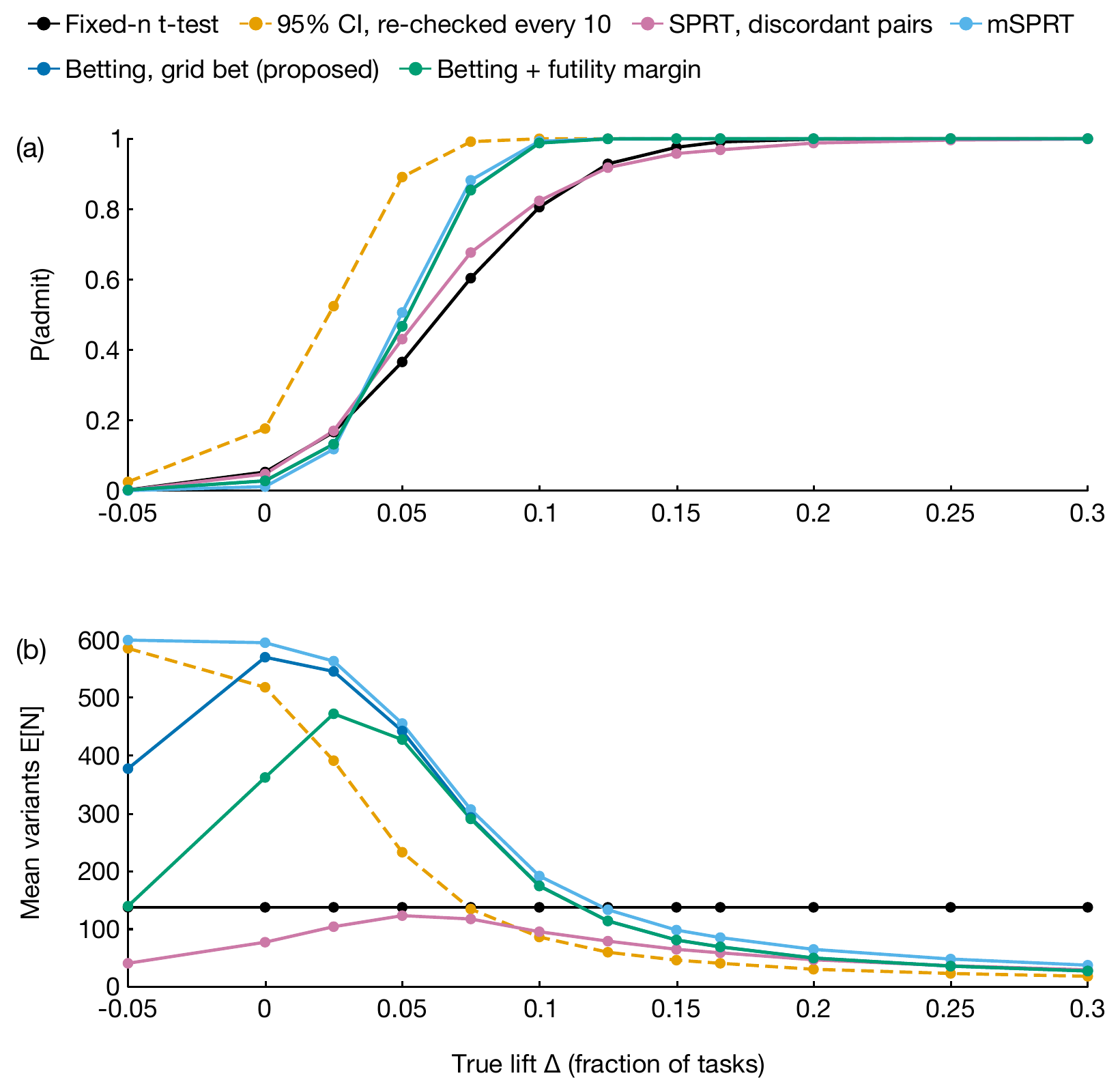}
\caption{(a) Admission probability and (b) mean variants per exam against the true lift (E2, \PRepsPower{} replicates per point, $N_{\max}=\PNmax$). The re-checked interval (dashed) is fast because it is invalid.}
\label{fig:power}
\end{figure}

Figure~\ref{fig:power} shows power and expected variants. At the design lift of \PLiftOne{} points the fixed-$n$ test has \NETwoPowFixed\% power with \NFixedN{} variants; the betting gate with futility margin has \NETwoPowBet\% power with \NETwoNBetLOne{} variants on average, and at the SkillsBench curated average (\PCurated{} points) it stops after \NETwoNBetCur{}, \NETwoSaveCur\% fewer than the fixed design. Anytime validity costs variants near the design lift and saves them above it. The futility margin cuts the variants a useless skill consumes from \NETwoNBetZero{} to \NETwoNBetFutZero. Across the default and moment-matched models with $\kappa\in\{-0.5,0,0.5\}$ (E14, $r=1$), the fixed design needs \NRvSensNFixedLo{} to \NRvSensNFixedHi{} variants, and the betting gate uses \NRvSensNBetLo{} to \NRvSensNBetHi{} at the design lift (power at least \NRvSensPowBetLo\%) and \NRvSensNCurLo{} to \NRvSensNCurHi{} at the curated lift, \NRvSensSaveLo\% to \NRvSensSaveHi\% fewer than the fixed design. No model raises the betting gate's false-admission rate above \NRvSensNullMax\%. At $r=3$ the moment-matched model raises the betting gate's variants at the design lift from \NRvDefRThreeNBet{} to \NRvFitRThreeNBet.

\paragraph{Where the betting gate loses.} Three regimes favour other designs. With a margin $\delta=0.05$, the betting gate has \NRvMarginPowBet\% power at the design lift against \NRvMarginPowFixed\% for a fixed-$n$ test planned for the margin (\NRvMarginNFixed{} variants, against \NRvMarginNBet{} on average for the betting gate), because only five points of lift lie above the margin and the bets are capped. With $r=3$ attempts per arm the differences are less noisy: the fixed design needs only \NRvRThreeNFixed{} variants for \NRvRThreePowFixed\% power, while the betting gate, with higher power, uses \NRvRThreeNBet{} on average. For pass/fail scores at $\delta=0$, the discordant-pair SPRT is cheaper at matched power (E14). At the design lift it reaches \NRvSprtPowLow\% power with \NRvSprtNLow{} variants, where the betting gate with futility, truncated at \NRvBetNMaxLow{} variants, reaches \NRvBetPowLow\% with \NRvBetNLow. At \NRvSprtPowHigh\% power the SPRT needs \NRvSprtNHigh{} variants, the betting gate \NRvBetNHigh{} and a fixed-$n$ test \NRvFixedNHigh (here the cap is fixed; Table~\ref{tab:matched}, from E18, instead tunes each gate's cap to the target power). The SPRT gains by assuming the form of the alternative, and it does not apply to graded scores, to $r>1$ or to a margin. Anytime validity is therefore not free. It pays when lifts are well above the design lift and when exams are monitored, stopped early or resumed, and it costs power when the margin is close to the lift.

In libraries whose lifts follow the published benchmark summaries (E5, Table~\ref{tab:scen} in Appendix~\ref{app:more}), the cost is dominated by useless candidates. In a SWE-like library, where 94\% of skills have no effect, \NEFiveSweFalseFixed\% of the skills admitted by an uncorrected fixed-$n$ test are useless, \NEFiveSweFalsePeekCI\% for the re-checked interval, \NEFiveSweFalseBet\% for the betting gate at a per-exam $\alpha$, and \NEFiveSweFalseELond\% with e-LOND, at \NEFiveSweRunsELond{} agent runs per truly helpful admission.

Resampling each SkillsBench curated skill's 18 per-configuration differences as if they were exam variants, certifying the median helpful skill takes \NSbMedNAlpha{} variants at a per-exam $\alpha$ and \NSbMedNBonf{} at $\alpha/87$ (95\% bootstrap intervals over the skills, \NSbMedNAlphaLo--\NSbMedNAlphaHi{} and \NSbMedNBonfLo--\NSbMedNBonfHi; E24) (Figure~\ref{fig:need} in Appendix~\ref{app:more}); only \NSbCertSix\% are certified within $N_{\max}=600$ at $\alpha/87$. Under the library procedure the gate actually uses, e-LOND with $\gamma_t\propto t^{-1.6}$ over random arrival orders of the 87 skills, the median is \NRvLibELond{} variants (\NRvLibELondCert\% certified within 600), and \NRvLibHorizon{} (\NRvLibHorizonCert\%) when $\gamma_t$ is spread evenly over a planned horizon of 87 (E14). Two assumptions sit behind these numbers. First, the configurations are treated as exchangeable draws of exam variants. They are harness-model pairs, not variants of one task, and we cannot test the assumption. Second, the skills are selected on positive observed lift, which flatters them. Selecting skills on nine configurations and resampling from the same nine gives a median of \NRvSelSame{} variants at $\alpha/87$, against \NRvSelHeld{} when the cost is resampled from the other nine (\NRvSelSameFin\% and \NRvSelHeldFin\% of selected skills certified within 2{,}000; 100 splits, about \NRvSelN{} skills selected). The full-data figures are therefore optimistic. Fixed benchmarks with tens of tasks cannot supply this many paired units per skill; parametric generators can, which is a second reason for hidden variants beyond leakage control.

\subsection{Library error control and resubmission}
\label{sec:res-library}

Table~\ref{tab:lib} reports FDR and power over histories of \PWindows{} windows of \PWindow{} candidates, \PNullShare\% of them useless. Testing each exam at $\alpha$ without correction gives \NEThreeFdrPerExam\% FDR, a figure that grows with the null share. Stopping at $1/\alpha$ and then applying e-BH per window admits nothing (Remark~\ref{prop:ebh}); stopping at $m/\alpha$ or running to $N_{\max}$ restores power at \NEThreeNStopm{} and \NEThreeNRunMax{} variants per exam. e-LOND reaches \NEThreePowELondH\% power at \NEThreeFdrELondH\% FDR with $\gamma_t$ spread over a planned horizon. Over a longer history of \NEbT{} exams (E3b, 60 histories), the deployed $t^{-1.6}$ sequence keeps FDR at \NEbInfFdr{} with \NEbInfPow\% power (rerun with 600 histories: \NIvEbInfFdr\%, 95\% interval up to \NIvEbInfFdrHi\%; E24); power falls from \NEbInfPowFirst\% in the first tenth of the history to \NEbInfPowLast\% in the last; the level $\alpha\gamma_t(R_{t-1}+1)$ grows with the number of earlier admissions, which offsets part of the decay of $\gamma_t$. A known horizon of \NEbT{} gives \NEbHorPow\% power at FDR \NEbHorFdr{}, and testing each exam at $\alpha$ gives \NEbPerPow\% power at FDR \NEbPerFdr{}, above the target. For a first attempt the e-LOND term, not the lineage term, sets the level in \NEbElondBinds\% of exams. With the PrPl bet (cap $1/2$) in place of the grid bet, e-LOND reaches \NEThreePrplPowELond\% power with \NEThreePrplNELond{} variants per exam. A useless skill resubmitted to a per-exam betting gate is admitted within five attempts with probability \NResubFivePerExam\% and within \PResub{} with probability \NResubPerExam\% (\NResubPeekCI\% with a re-checked interval); lineage alpha-spending holds this at \NResubLineage\% (Figure~\ref{fig:resub} in Appendix~\ref{app:more}).

\paragraph{Against stopped e-BH (E20).} Two recent procedures apply e-BH to e-processes directly. Stopped e-BH \citep{wang2025stopped} monitors the e-BH set continuously and controls FDR at any stopping time when the streams are independent; e-BH on suprema adjusted by $A(M)=\sqrt M-1$ \citep{tavyrikov2025carefree} controls FDR under any dependence. We ran both on the E3 histories with the grid bet, letting each exam stop as soon as it enters the e-BH set of its window or at futility. Stopped e-BH reached \NEbhGridSebhPow\% power at \NEbhGridSebhFdr\% FDR with \NEbhGridSebhN{} variants per exam, against \NEbhGridElondPow\% power, \NEbhGridElondFdr\% FDR and \NEbhGridElondN{} variants for the deployed e-LOND in E20's own histories (Table~\ref{tab:lib} reports E3's run of the same rule) (adjusted suprema and the PrPl bet: Appendix~\ref{app:more}). So when candidates arrive in windows whose exams can run side by side, and their data are independent, stopped e-BH is the better library rule. e-LOND remains the rule for a registry that decides each exam as it closes, because it fixes the level when the exam starts and holds under arbitrary dependence; that is the setting of Theorem~\ref{thm:price}, whose levels must be fixed in advance.

\begin{table}[t]
\centering\small
\caption{Library-level control (E3), with the grid bet. Mean over 400 simulated registry histories.}
\label{tab:lib}
\input{tables/t_library.tex}
\end{table}

\subsection{Reporting and verification}
\label{sec:res-report}
\label{sec:res-leak}

Because exams stop when the evidence is strong, the estimate at admission overstates the lift, by \NEFourRelBiasFive\% of the true value at $\Delta=0.05$ (E4, Figure~\ref{fig:wc}). The anytime-valid lower bound keeps time-uniform coverage of at least \NEFourCoverMin\%, but at admission it sits just above zero: it certifies that a skill helps, not by how much. Planted lookup tables are caught by normalized and decoded matching, and no detector flagged a clean skill (E6, Table~\ref{tab:leak}); a skill that ships a copy of the generator is not caught by any literal check. Verifying a \NESevenN-variant credential takes \NESevenVerify{}~s on a laptop, and the log holds \NESevenLedger{} bytes per exam, independent of transcript size (E7, Table~\ref{tab:verify}).

A first live pilot (OpenHands with Claude Haiku~5.5 on \NPilotTasks{} SkillsBench tasks) exercised the whole protocol, including a verifier that rejects a changed reward, and admitted nothing (Appendix~\ref{app:more}). A static security scan of the 87 SkillsBench skills (E15) found risk unrelated to released lift (Spearman $\rho_S=\NSecRho$) and one skill with a real concern among \NSecFlagN{} flags, so a helpfulness certificate is no evidence of safety and the scan belongs in the credential as a separate field (Appendix~\ref{app:more}).

\subsection{Live studies}
\label{sec:res-live}
\paragraph{On the benchmark.} We first ran, as study~1, the committed protocol on all \NLoTasks{} SkillsBench tasks with OpenHands and Claude Haiku~5.5, three trials per arm. Harness failures left \NLoScored{} tasks scorable, the final wealth was \NLoWealth, short of $1/\alpha=20$, and nothing was admitted. One paired unit per task gives too few units for any anytime-valid admission (Section~\ref{sec:res-cost}); Appendix~\ref{app:live-detail} gives the exclusions and amendments.

\paragraph{On generated variants.} We therefore ran the complete gate on parametric task families with live models (studies~2 and~3). In the first family a variant is an access log generated from a 32-byte seed that follows two house conventions which cannot be inferred without documentation; in the second it is an order export in which a channel code marks internal test orders. For each family we registered a helpful skill that documents the conventions, a placebo of similar length and a harmful skill that states wrong conventions; the first family also had a planted leak, the placebo plus a table of answers for 300 public example seeds. The helpful skill was the second attempt of the placebo's lineage, and the second study also resubmitted the first study's placebo five times. Exam $t$ ran at level $\min(\alpha_t,\alpha\gamma^{\text{lin}}_k)$ with the gate of Section~\ref{sec:gate}, on variants seeded from a drand round that did not exist when the commitment was logged. We ran both studies with Claude Haiku~5.5 and, as separate registry histories with everything else fixed by hash, with the open-weight Nemotron~3.5 Super VL. Appendix~\ref{app:live-detail} gives the families, model settings and exclusions.

\begin{table}[t]
\centering\small
\caption{Live studies with generated variants, with Claude Haiku~5.5 and with Nemotron~3.5 Super VL; each study and each model is its own registry history, with the same committed exam order and levels. Level: the exam's significance level. $\tau$: paired variants at the decision. With / without at $\tau$: passes with and without the skill among those $\tau$ variants; in parentheses, passes among the variants completed in the same batch after the decision, which the decision did not use. Grid: variants the grid bet of Theorem~\ref{thm:adaptive} would have needed on the same transcripts at the same level (a post hoc analysis; the committed bet was PrPl with cap $1/2$); it reaches the same decision in every exam. In study 4 (published skills) the grid bet was the committed bet, so its Grid column repeats $\tau$. Cap: the exam reached $N_{\max}$ without admission.}
\label{tab:live}
\resizebox{\linewidth}{!}{\input{tables/t_live_study.tex}}
\end{table}

Table~\ref{tab:live} gives the decisions. On both models the gate admitted the helpful skills, after \NLsHelpN{} and \NLsBHelpN{} variants with Haiku and \NLnHelpN{} and \NLnBHelpN{} with Nemotron, and stopped the placebos, the leak and the harmful skills for futility. The leak gave no lift, because the exam variants came from seeds that did not exist when it was registered; this checks commit and reveal, not the coverage detector of E6. The harmful skill of the second family lowered the pass rate in every discordant variant, but at the decision the anytime-valid upper bound on $\Delta$ was still above zero (\NLsBHarmUcb{} with Haiku), so the gate refused it without certifying harm; flagging harmful skills needs a separate test with its own error budget. The five resubmissions ran at levels falling from \NLsResubLevelFirst{} to \NLsResubLevelLast{} and were all refused. The verifier recomputed every grade and decision from the logged responses, and changing one recorded grade made it fail. The Haiku studies cost US\$\NLsCost{} and US\$\NLsThreeCost, and the Nemotron repetitions US\$\NLnTotalCost{} in all.

\paragraph{Published third-party skills.} We wrote every skill above ourselves, for families built so that the skill helps a lot. Study~4 ran the gate on skills that other people published, used byte for byte as prompt text. Before any model call we ranked a shortlist of instruction-only skills whose rules a program can check, and fixed how one would be picked: the first whose lift, in a public calibration with Haiku on 20 variants per arm, fell in $[0.05,0.40]$. Joe Amditis's AP news-style skill (MIT licence) ranked first but lifted Haiku by \NCalFourBLift, and Sentry's commit-message skill (Apache-2.0), second, by \NCalFourCLift. The APA~7 skill of Kalaris Labs (MIT), third, gave \NCalFourLift{} and so became study~4. We did not drop the first two: they ran under the same design as studies 4b and 4c, outside the rule. Each skill got its own task family. In each study a matched-length placebo opened a lineage in a new registry history, and the published skill was its second attempt. The bet was the grid bet of Theorem~\ref{thm:adaptive}, and a simulation from the calibration, run before commitment, set $N_{\max}=\NPowNmax$ (power \NPowCons{} under a cautious reading; Appendix~\ref{app:live-extra}). Three protocol deviations, a wrong label, a failed first run and an early reveal of the committed secret, are explained there; none changes a decision.

Table~\ref{tab:live} gives the decisions. The gate admitted all three published skills on both models. The APA skill was admitted after \NLsFourRealN{} variants with Haiku (\NLsFourRealTauWith{} passes with it, \NLsFourRealTauWithout{} without) and after \NLnFourRealN{} with Nemotron (\NLnFourRealTauWith{} and \NLnFourRealTauWithout), lifts at the decision of \NLsFourRealLiftTau{} and \NLnFourRealLiftTau, inside the window of the selection rule; the AP and commit skills, whose lifts were large, were admitted after \NLFourBCMinN{} to \NLFourBCMaxN{} variants. The APA placebo was stopped for futility with Haiku after \NLsFourPlacN{} variants. With Nemotron its pass rate stayed slightly above the rate without it (\NLnFourPlacTauWith{} against \NLnFourPlacTauWithout{} passes over \NLnFourPlacN{} variants), a difference too small either to admit it or to stop it for futility, so the exam ran to $N_{\max}$ and ended without admission; this took most of the arm's US\$\NLnFourCost. The commit placebo was stopped for futility on both models. In the Nemotron APA arm, rate-limit errors (HTTP 429) excluded \NLnFourPlacExcl{} of \NLnFourPlacRun{} pairs in the placebo exam and \NLnFourRealExcl{} of \NLnFourRealRun{} in the skill exam, as the committed rule requires; such errors come back before any answer and struck both arms alike, so they cost time, not validity (Appendix~\ref{app:live-extra}). Study~4 cost about US\$\NLFourTotalCost{} in API fees, including all calibrations, and every credential verified.

\emph{A placebo that helped.} In study 4b the gate also admitted the placebo, on both models (after \NLsFourBPlacN{} variants with Haiku and \NLnFourBPlacN{} with Nemotron; at the decision \NLsFourBPlacTauWith{} and \NLnFourBPlacTauWith{} passes with it, \NLsFourBPlacTauWithout{} and \NLnFourBPlacTauWithout{} without). This is not a false admission. The placebo states none of the graded rules: it is general advice on news writing, verification and corrections. But it frames the task as newsroom work, and we think that with it both models applied conventions they already knew; the paired design shows the effect on the pass rate, not its cause. With it Haiku wrote ``over'' for ``more than'' in \NLsFourBPlacOverWith{} of \NLsFourBPlacOverN{} briefs that needed it, against \NLsFourBPlacOverWithout{} without (Appendix~\ref{app:live-extra}). The gate certifies that a text raises the pass rate, which this one did; it does not certify why. A placebo of matched length controls for the length of a skill and for the presence of one, not for what advice on the same activity reminds a model of. The placebo was the first attempt of its lineage, so the published skill, its second attempt, was tested at the lower level; it was still admitted, after fewer variants.

\paragraph{A family built by someone else.} Every family above was built by us. Study~5 (E25) ran the gate on Google DeepMind's \texttt{mathematics\_dataset} generator \citep{saxton2019analysing}, installed unmodified; we did not write its questions, answers or difficulty levels. Before any skill was run we fixed a rule that picks the generator's modules on which Haiku without a skill passes 30\% to 60\% of public-seed questions, at most one per area. It picked primality testing (hardest training regime) and division of large numbers (extrapolation regime), and each exam variant is a question from one of the two, drawn by a committed seed and graded by exact answer. We registered three math skills published by other authors, used byte for byte, a placebo of general work habits and a harmful skill that states wrong shortcuts for both modules, each as its own lineage in a new registry history. In the public-seed calibration every skill lowered Haiku's pass rate, so the plan predicted no admission and fixed $N_{\max}=600$ by the rule set in advance.

\begin{table}[t]
\centering\small
\caption{Live study 5 on Google DeepMind's \texttt{mathematics\_dataset} (primality and large division). One registry history per model, same committed exam order, levels and seeds. $\tau$: paired variants at the decision; lift: pass rate with the skill minus without, in points, over those $\tau$ variants. Cap: the exam reached $N_{\max}=600$ without crossing either boundary.}
\label{tab:live5}
\resizebox{\linewidth}{!}{%
\begin{tabular}{rllrrlrrlr}
\toprule
& & & \multicolumn{3}{c}{Claude Haiku~5.5} & & \multicolumn{3}{c}{Nemotron~3.5 Super VL}\\
\cmidrule{4-6}\cmidrule{8-10}
$t$ & Skill & Level & $\tau$ & Decision & Lift & & $\tau$ & Decision & Lift\\
\midrule
1 & \texttt{math-teacher} (published) & \NLsFiveTeachLevel & \NLsFiveTeachN & \NLsFiveTeachDec & \NLsFiveTeachLift & & \NLnFiveTeachN & \NLnFiveTeachDec & \NLnFiveTeachLift\\
2 & harmful (ours) & \NLsFiveHarmLevel & \NLsFiveHarmN & \NLsFiveHarmDec & \NLsFiveHarmLift & & \NLnFiveHarmN & \NLnFiveHarmDec & \NLnFiveHarmLift\\
3 & placebo (ours) & \NLsFivePlacLevel & \NLsFivePlacN & \NLsFivePlacDec & \NLsFivePlacLift & & \NLnFivePlacN & \NLnFivePlacDec & \NLnFivePlacLift\\
4 & \texttt{math-help} (published) & \NLsFiveHelpLevel & \NLsFiveHelpN & \NLsFiveHelpDec & \NLsFiveHelpLift & & \NLnFiveHelpN & \NLnFiveHelpDec & \NLnFiveHelpLift\\
5 & \texttt{Math.Skill} (published) & \NLsFiveSkillLevel & \NLsFiveSkillN & \NLsFiveSkillDec & \NLsFiveSkillLift & & \NLnFiveSkillN & \NLnFiveSkillDec & \NLnFiveSkillLift\\
\midrule
& API cost (US\$) & & \multicolumn{3}{r}{\NLsFiveCost} & & \multicolumn{3}{r}{\NLnFiveCost}\\
\bottomrule
\end{tabular}}
\end{table}

Table~\ref{tab:live5} gives the decisions. No published math skill helped either model: each had a negative observed lift, from \NLsFiveHelpLift{} to \NLsFiveSkillLift{} points with Haiku and from \NLnFiveHelpLift{} to \NLnFiveSkillLift{} with Nemotron, and the gate stopped each for futility (lift confidently below 5 points) within \NLsFiveHelpN{} variants. It stopped the placebo the same way on both models and the harmful skill with Haiku after \NLsFiveHarmN{} variants. With Nemotron the harmful skill \emph{raised} the pass rate, with a lift of \NLnFiveHarmLift{} points over 600 variants (the anytime-valid lower bound of the PrPl confidence sequence at the exam's level, which the credential reports, is \NLnFiveHarmLcb{} points), and the exam ran to $N_{\max}$ without admitting it. The skill tells the model to trust short rules instead of long calculations; without it Nemotron hit its 16{,}000-token limit in \NLnFiveHarmCutPctWithout\% of answers, with it in \NLnFiveHarmCutPctWith\%. So a skill written to be wrong helped a model whose failures were of another kind, as the placebo of study 4b did, and the gate, which measures lift rather than intent, should admit such a skill once the evidence reaches the exam's level. Here 600 variants were too few. Resampling the exam's \NLnFiveHarmValid{} valid pairs, which treats the observed lift as the true one and is therefore optimistic, the gate admits by 600 variants in \NLnFiveHarmBootSix\% of replays, with a median of \NLnFiveHarmBootMed{} variants and \NLnFiveHarmBootTwoK\% by 2{,}000, at this exam's level of \NLnFiveHarmLevel; at a true lift of 4 points the share by 2{,}000 falls to \NLnFiveHarmFourTwoK\%. The refusal is therefore consistent with a miss forced by the cap, which the plan set from a Haiku calibration in which this skill lowered the pass rate by 30 points. It is not evidence that the skill does not help Nemotron. The whole study cost US\$\NLFiveTotalCost{} on both models; Appendix~\ref{app:study5} gives the selection, a grader change made before any skill was run, and two restarts.

So the gate admitted every skill whose lift could be certified within the committed budget, ours and published ones, and in studies 2, 3 and 5 refused every text that did not, with the same decisions on two models. Study~4 is less tidy, with one placebo admitted and one run to $N_{\max}$, and so is study 5 on Nemotron, where a skill we wrote to hurt helped a little and ran to $N_{\max}$. Study 5 also shows the other side of the gate: on a family built by others, published skills did not help, and refusing all five skills cost US\$\NLsFiveCost{} with Haiku. Because the study-4 skills were chosen for checkable rules and ran on families we built, the studies show that the protocol works on real skills, not how often published skills help; an audit of 30 community skills found that many popular ones do not consistently beat an agent without them \citep{ying2026openskilleval}.

%% file: tables/t_false_admission.tex
\begin{tabular}{lrrrrrrrr}
\toprule
Gate & Boundary & Worst case & $s_\theta=3$ & $\rho=0$ & $r=3$ & $\delta=0.05$ & Stale baseline & SWE null \\
\midrule
Fixed-$n$ paired $t$ & 4.9 & 4.3 & 5.1 & 5.2 & 5.0 & 5.1 & 65.4 & 3.3 \\
$t$-test re-checked each variant & 32.0 & 34.5 & 30.2 & 33.9 & 33.8 & 30.7 & 98.7 & 30.3 \\
95\% CI re-checked every 10 & 18.1 & 18.9 & 16.2 & 19.9 & 19.2 & 16.7 & 96.8 & 15.4 \\
SPRT, discordant pairs & 4.3 & 4.2 & 4.9 & 4.6 & -- & -- & 69.7 & -- \\
mSPRT & 1.3 & 1.3 & 0.9 & 1.6 & 2.2 & 1.4 & 75.6 & 0.5 \\
Betting, cap $\lambda\le m_0$ & 1.5 & 2.7 & 0.9 & 2.1 & 0.5 & 1.9 & 75.4 & 0.2 \\
Betting, $c=1/2$ & 1.5 & 2.7 & 0.9 & 2.1 & 0.5 & 1.8 & 75.4 & 0.2 \\
Betting, $c=3/4$ & 2.2 & 3.3 & 1.8 & 2.8 & 1.3 & 2.5 & 69.7 & 0.8 \\
Betting, aGRAPA & 2.1 & 2.6 & 1.9 & 2.4 & 1.7 & 2.3 & 73.0 & 1.3 \\
Betting, $c=1/2$, futility margin & 1.5 & 2.7 & 0.9 & 2.1 & 0.5 & 1.8 & 75.4 & 0.2 \\
Betting, grid (deployed) & 2.7 & 2.8 & 2.6 & 2.9 & 3.0 & 3.3 & 66.3 & 2.8 \\
Betting, grid, futility margin & 2.7 & 2.8 & 2.6 & 2.9 & 3.0 & 3.3 & 66.2 & 2.2 \\
\bottomrule
\end{tabular}

%% file: tables/t_matched_power.tex
\begin{tabular}{llrrrrr}
\toprule
Lift & Power & Fixed-$n$ & SPRT & mSPRT & PrPl & Grid \\
\midrule
0.1 & 80\% & 137 (137) & 96 (76) & 167 (278) & 167 (219) & 148 (227) \\
0.1 & 95\% & 239 (239) & 124 (141) & 183 (427) & 183 (297) & 167 (313) \\
0.166 & 80\% & 55 (55) & 41 (38) & 76 (120) & 87 (116) & 57 (97) \\
0.166 & 95\% & 96 (96) & 57 (70) & 81 (164) & 91 (149) & 64 (147) \\
\bottomrule
\end{tabular}

%% file: tables/t_library.tex
\begin{tabular}{lrrr}
\toprule
Rule & FDR (\%) & Power (\%) & Mean variants \\
\midrule
Betting at $\alpha$ per exam, no multiplicity control & 10.1 & 95 & 317 \\
Stop at $1/\alpha$, then e-BH per window & 0.0 & 0 & 317 \\
Stop at $m/\alpha$, then e-BH per window & 0.6 & 88 & 345 \\
No early admission, e-BH per window & 0.0 & 89 & 421 \\
e-LOND, $\gamma_t\propto t^{-1.6}$ & 0.1 & 82 & 356 \\
e-LOND, uniform $\gamma_t$ over planned horizon & 1.0 & 88 & 343 \\
\bottomrule
\end{tabular}

%% file: tables/t_live_study.tex
\begin{tabular}{llrllr@{\hskip 8pt}lrlr@{\hskip 8pt}lrlr}
\toprule
 & & & & & & \multicolumn{4}{c}{Claude Haiku~5.5} & \multicolumn{4}{c}{Nemotron~3.5 Super VL} \\
\cmidrule(lr){7-10}\cmidrule(lr){11-14}
Study & Family & $t$ & Lineage, attempt & Skill & Level & Decision & $\tau$ & With / without at $\tau$ & Grid & Decision & $\tau$ & With / without at $\tau$ & Grid \\
\midrule
2 & logs & 1 & A, 1 & placebo & \NLsPlacLevel & futility & \NLsPlacN & \NLsPlacTauWith{} / \NLsPlacTauWithout\,{\scriptsize(+\NLsPlacExtraWith/\NLsPlacExtraWithout)} & \NLsPlacGridN & \NLnPlacDec & \NLnPlacN & \NLnPlacTauWith{} / \NLnPlacTauWithout\,{\scriptsize(+\NLnPlacExtraWith/\NLnPlacExtraWithout)} & \NLnPlacGridN \\
2 & logs & 2 & B, 1 & planted leak & \NLsLeakLevel & futility & \NLsLeakN & \NLsLeakTauWith{} / \NLsLeakTauWithout\,{\scriptsize(+\NLsLeakExtraWith/\NLsLeakExtraWithout)} & \NLsLeakGridN & \NLnLeakDec & \NLnLeakN & \NLnLeakTauWith{} / \NLnLeakTauWithout\,{\scriptsize(+\NLnLeakExtraWith/\NLnLeakExtraWithout)} & \NLnLeakGridN \\
2 & logs & 3 & C, 1 & harmful & \NLsHarmLevel & futility & \NLsHarmN & \NLsHarmTauWith{} / \NLsHarmTauWithout\,{\scriptsize(+\NLsHarmExtraWith/\NLsHarmExtraWithout)} & \NLsHarmGridN & \NLnHarmDec & \NLnHarmN & \NLnHarmTauWith{} / \NLnHarmTauWithout\,{\scriptsize(+\NLnHarmExtraWith/\NLnHarmExtraWithout)} & \NLnHarmGridN \\
2 & logs & 4 & A, 2 & helpful & \NLsHelpLevel & admit & \NLsHelpN & \NLsHelpTauWith{} / \NLsHelpTauWithout\,{\scriptsize(+\NLsHelpExtraWith/\NLsHelpExtraWithout)} & \NLsHelpGridN & \NLnHelpDec & \NLnHelpN & \NLnHelpTauWith{} / \NLnHelpTauWithout\,{\scriptsize(+\NLnHelpExtraWith/\NLnHelpExtraWithout)} & \NLnHelpGridN \\
\midrule
3 & orders & 1 & D, 1 & placebo & \NLsBPlacLevel & futility & \NLsBPlacN & \NLsBPlacTauWith{} / \NLsBPlacTauWithout\,{\scriptsize(+\NLsBPlacExtraWith/\NLsBPlacExtraWithout)} & \NLsBPlacGridN & \NLnBPlacDec & \NLnBPlacN & \NLnBPlacTauWith{} / \NLnBPlacTauWithout\,{\scriptsize(+\NLnBPlacExtraWith/\NLnBPlacExtraWithout)} & \NLnBPlacGridN \\
3 & orders & 2 & E, 1 & harmful & \NLsBHarmLevel & futility & \NLsBHarmN & \NLsBHarmTauWith{} / \NLsBHarmTauWithout\,{\scriptsize(+\NLsBHarmExtraWith/\NLsBHarmExtraWithout)} & \NLsBHarmGridN & \NLnBHarmDec & \NLnBHarmN & \NLnBHarmTauWith{} / \NLnBHarmTauWithout\,{\scriptsize(+\NLnBHarmExtraWith/\NLnBHarmExtraWithout)} & \NLnBHarmGridN \\
3 & orders & 3 & D, 2 & helpful & \NLsBHelpLevel & admit & \NLsBHelpN & \NLsBHelpTauWith{} / \NLsBHelpTauWithout\,{\scriptsize(+\NLsBHelpExtraWith/\NLsBHelpExtraWithout)} & \NLsBHelpGridN & \NLnBHelpDec & \NLnBHelpN & \NLnBHelpTauWith{} / \NLnBHelpTauWithout\,{\scriptsize(+\NLnBHelpExtraWith/\NLnBHelpExtraWithout)} & \NLnBHelpGridN \\
3 & logs & 4--8 & G, 1--5 & placebo & \NLsResubLevelFirst--\NLsResubLevelLast & futility (all) & \NLsResubNmin{} each & equal in every variant & same & futility (all) & \NLnResubNmin--\NLnResubNmax & \NLnResubDisc{} discordant & same \\
\midrule
4 & APA refs & 1 & S, 1 & placebo & \NLsFourPlacLevel & \NLsFourPlacDec & \NLsFourPlacN & \NLsFourPlacTauWith{} / \NLsFourPlacTauWithout\,{\scriptsize(+\NLsFourPlacExtraWith/\NLsFourPlacExtraWithout)} & \NLsFourPlacN & \NLnFourPlacDec & \NLnFourPlacN & \NLnFourPlacTauWith{} / \NLnFourPlacTauWithout\,{\scriptsize(+\NLnFourPlacExtraWith/\NLnFourPlacExtraWithout)} & \NLnFourPlacN \\
4 & APA refs & 2 & S, 2 & published & \NLsFourRealLevel & \NLsFourRealDec & \NLsFourRealN & \NLsFourRealTauWith{} / \NLsFourRealTauWithout\,{\scriptsize(+\NLsFourRealExtraWith/\NLsFourRealExtraWithout)} & \NLsFourRealN & \NLnFourRealDec & \NLnFourRealN & \NLnFourRealTauWith{} / \NLnFourRealTauWithout\,{\scriptsize(+\NLnFourRealExtraWith/\NLnFourRealExtraWithout)} & \NLnFourRealN \\
4b & AP style & 1 & S, 1 & placebo & \NLsFourBPlacLevel & \NLsFourBPlacDec & \NLsFourBPlacN & \NLsFourBPlacTauWith{} / \NLsFourBPlacTauWithout\,{\scriptsize(+\NLsFourBPlacExtraWith/\NLsFourBPlacExtraWithout)} & \NLsFourBPlacN & \NLnFourBPlacDec & \NLnFourBPlacN & \NLnFourBPlacTauWith{} / \NLnFourBPlacTauWithout\,{\scriptsize(+\NLnFourBPlacExtraWith/\NLnFourBPlacExtraWithout)} & \NLnFourBPlacN \\
4b & AP style & 2 & S, 2 & published & \NLsFourBRealLevel & \NLsFourBRealDec & \NLsFourBRealN & \NLsFourBRealTauWith{} / \NLsFourBRealTauWithout\,{\scriptsize(+\NLsFourBRealExtraWith/\NLsFourBRealExtraWithout)} & \NLsFourBRealN & \NLnFourBRealDec & \NLnFourBRealN & \NLnFourBRealTauWith{} / \NLnFourBRealTauWithout\,{\scriptsize(+\NLnFourBRealExtraWith/\NLnFourBRealExtraWithout)} & \NLnFourBRealN \\
4c & commits & 1 & S, 1 & placebo & \NLsFourCPlacLevel & \NLsFourCPlacDec & \NLsFourCPlacN & \NLsFourCPlacTauWith{} / \NLsFourCPlacTauWithout\,{\scriptsize(+\NLsFourCPlacExtraWith/\NLsFourCPlacExtraWithout)} & \NLsFourCPlacN & \NLnFourCPlacDec & \NLnFourCPlacN & \NLnFourCPlacTauWith{} / \NLnFourCPlacTauWithout\,{\scriptsize(+\NLnFourCPlacExtraWith/\NLnFourCPlacExtraWithout)} & \NLnFourCPlacN \\
4c & commits & 2 & S, 2 & published & \NLsFourCRealLevel & \NLsFourCRealDec & \NLsFourCRealN & \NLsFourCRealTauWith{} / \NLsFourCRealTauWithout\,{\scriptsize(+\NLsFourCRealExtraWith/\NLsFourCRealExtraWithout)} & \NLsFourCRealN & \NLnFourCRealDec & \NLnFourCRealN & \NLnFourCRealTauWith{} / \NLnFourCRealTauWithout\,{\scriptsize(+\NLnFourCRealExtraWith/\NLnFourCRealExtraWithout)} & \NLnFourCRealN \\
\bottomrule
\end{tabular}

%% file: sections/discussion.tex
\section{Discussion, limitations and conclusion}
\label{sec:discussion}

\paragraph{Recommendations for registries.} (1) Run the arms concurrently on the same variants and never compare against a stored baseline. (2) Register the stopping rule and use a test that is valid at every stopping time: the betting test with a futility margin in general, or the discordant-pair SPRT for pass/fail scores at $\delta=0$. (3) Treat admission as online multiple testing: give each exam an e-LOND level when it starts, and spend alpha over a skill's lineage so that resubmission does not pay. (4) Generate hidden variants from a committed seed; this controls leakage and also supplies the hundreds of paired units that certification needs. (5) Lead credentials with the anytime-valid lower bound, not the stopped estimate. (6) Publish both arms' transcripts and the registered parameters so the decision can be recomputed.

\paragraph{What the guarantee assumes.} Two kinds of assumption carry the guarantees, and they differ in strength. Within an exam, validity needs the conditional null of Eq.~\eqref{eq:H}, which holds under the design of Section~\ref{sec:setup}. These departures break it: state shared across variants (agent memory, response caching, batched calls that couple variants); a provider change during an exam that affects one arm more than the other; repeated seeds, which turn one variant's outcome into several; exclusion rules that depend on the arm or on the outcome; and any exam-level latent state that is not fixed before the exam, as in the coin example of Section~\ref{sec:setup}. Concurrent arms, fresh seeds, stateless calls and arm-blind exclusion rules registered in $\theta$ protect against each. E21 measures each departure for a useless skill (Table~\ref{tab:nullstress}, Appendix~\ref{app:more}). Mild versions leave the deployed grid gate near its target: a model change that affects both arms, 1 to 5\% repeated seeds or excluded skill-arm losses, and small shared effects give at most \NNsMildMax\%, against \NNsControl\% under the design. Strong versions do not: a model change that hits only the skill arm gives \NNsSkillArm\%, shared effects over blocks of ten variants \NNsBlock\%, 20\% repeats or exclusions \NNsRepeat\% and \NNsExclude\%, and an exam-level shift of the lift by $\pm0.05$ gives \NNsCoin\%. The fixed-$n$ test fails in the same cells, so these are failures of the design, not of sequential testing; each is prevented by the corresponding protocol rule. Across exams the conditions are much weaker: e-LOND and lineage spending need only that each exam's stopped wealth is an e-value given the level fixed at its start, and they hold under arbitrary dependence between exams.

\begin{table}[t]
\centering
\small
\caption{Threat model. \emph{Prevented}: the protocol makes the attack fail. \emph{Detectable}: a verifier can see it from the published record. \emph{Assumed}: the guarantee relies on an honest party. \emph{External}: needs enforcement outside the protocol.}
\label{tab:threats}
\begin{tabular}{p{0.36\linewidth}lp{0.42\linewidth}}
\toprule
Threat & Status & Mechanism or requirement \\
\midrule
Skill tailored to the hidden variants & Prevented & Skill digest logged before the commitment; seed mixes a later beacon \\
Proctor picks favourable variants & Prevented & Commitment precedes the beacon value \\
Stopping at a lucky moment & Prevented & Test valid at every stopping time (Proposition~\ref{prop:valid}) \\
Abandoning unfavourable exams & Prevented & Commitments stay visible, and the registry counts every committed exam, finished or not, in e-LOND's index and the lineage's attempt count; abandoning only removes admissions \\
Resubmitting until a pass & Prevented if linked & Lineage spending (Proposition~\ref{prop:lineage}) \\
Arm- or outcome-dependent exclusions & Detectable & Exclusion rule registered in $\theta$; excluded runs published \\
Skill copies the generator or its solver & Detectable in part & Coverage statistic; reviewing the skill text (Section~\ref{sec:res-leak}) \\
Choice of generator and margin & Assumed & Registry chooses them before the skill is seen; the credential certifies lift on that generator only \\
Proctor leaks the seed & Assumed & Threshold proctors or seed release inside trusted hardware \\
Transcripts not produced by the stated model & External & Attested execution; not provided by the protocol \\
Sybil authors evading lineage linking & External & Identity verification or stake \\
\bottomrule
\end{tabular}
\end{table}

Table~\ref{tab:threats} summarizes what the protocol protects. ``Verifiable'' in this paper means that a third party can recompute the decision from the published transcripts and commitments; it does not mean attested execution.

\paragraph{Limitations.} The entropy price of Theorem~\ref{thm:price} is the cost of the lineage guarantee alone; under the deployed level, in a registry past its first few dozen exams, online FDR control is the binding constraint and the price is small (E19). The validity results are theorems; the cost results depend on an outcome model whose nuisance parameters have no published values. We sweep them and fit them to the released SkillsBench moments, and the cost claims move within the ranges of Section~\ref{sec:res-cost}; the resampling estimates there assume that variants behave like the released configurations, which we cannot check, and they are optimistic because the skills were selected on observed lift. The reanalyses use benchmarks whose units (requirement batches, configurations) are not the hidden parametric variants we propose, and the SkillsBench registry lacks the self-generated condition. The generated-variant studies use two models; the benchmark study uses one. Nemotron's answers reached the token limit in \NLnCutPct\% of calls in studies 2 and 3 and \NLnFiveCutPct\% in study 5, which lowers the pass rates the gate sees; a reasoning model's budget is part of what a credential certifies. On the benchmark most tasks were excluded for harness failures, and studies 2 to 4 use task families that we designed. On the one family built by others (study 5) no published skill helped; the one text that did help, our harmful skill with Nemotron, was not certified within the cap. So we have shown the gate refusing on an external family but not admitting on one. When exams can be batched in windows and their data are independent, stopped e-BH \citep{wang2025stopped} is more powerful than e-LOND in our simulations (E20); our gate does not use it because a registry decides exams one at a time, and the lineage theory needs levels fixed at the start of each exam. The moment-matched outcome model's fitted correlation lies at the edge of the searched grid. None of the live studies tests whether any particular way of producing skills, such as peer synthesis, passes the gate more often. The verification algorithm establishes that the decision follows from the published transcripts, not that the transcripts were produced by the stated model; remote model APIs are not attestable end to end, and trusted execution has known side channels. A proctor who leaks the seed to an author defeats commit and reveal; threshold seeding by independent proctors is the natural fix and is not analysed here. Lineage alpha-spending is only as good as the registry's ability to link resubmissions, and the leakage experiment uses one synthetic task family.

\paragraph{Conclusion.} Skills are becoming the unit in which agent capability is shared, and registries already decide which ones to list. Those decisions are sequential, multiple and repeatable, and the per-skill evidence that benchmarks collect today is enough for a one-off fixed-sample analysis of some skills but certifies only a few once the analysis must survive sequential testing, a growing library and resubmission. A gate built from a paired exam on committed hidden variants, a betting test with a futility margin, e-LOND with lineage alpha-spending, and a recomputable dual-arm credential holds its error rates in every condition we tested that meets the design of Section~\ref{sec:setup}, and its cost can be planned: if hidden variants behave like the released configurations, from about fifty to a few hundred variants per certified skill.

%% file: sections/impact.tex
The gate is meant to make skill registries harder to game, and the paper describes how they can be gamed: by re-checking results, shopping across configurations, resubmitting, embedding encoded answer tables, or comparing against a stale baseline. These attacks are simple and already possible against registries that publish with-and-without verdicts; describing them alongside the defences should help registries more than attackers. Published transcripts can contain data from task environments, so task classes should use synthetic or licensed data, and redaction should happen before hashing under a registered policy. Logged records are permanent, so no personal data should be written to the log. A certified skill raises an agent's capability on a task class, which can include misuse; the registry's choice of task classes, not the statistics, is where that is controlled.

%% file: sections/appendix_notation.tex
\section{Notation}
\label{app:notation}

\begin{table}[H]
\centering\small
\caption{Main notation.}
\label{tab:notation}
\begin{tabular}{@{}l>{\raggedright\arraybackslash}p{0.62\linewidth}@{}}
\toprule
Symbol & Meaning \\
\midrule
$i$, $D_i$ & variant index; paired difference, pass with skill minus pass without, in $[-1,1]$ \\
$X_i=(1+D_i)/2$, $m_0=(1+\delta)/2$ & rescaled difference and its null mean \\
$\mathcal F_{i-1}$ & registry history at exam start, skill, comparator, commitment and earlier transcripts \\
$\Delta_i=\mathbb E[D_i\mid\mathcal F_{i-1}]$, $\Delta$ & conditional and average lift \\
$\delta$, $\delta_f$ & admission margin (0 throughout); futility margin \\
$\alpha$, $\alpha_f$ & registry error level; futility level \\
$a$ & level of one exam; admit when wealth reaches $1/a$ (in Section~\ref{sec:setup}, also the agent) \\
$\lambda_i$, $K_n$ & bet and wealth (e-process) of an exam \\
$\tau$, $N_{\max}$ & stopping time of an exam; cap on variants \\
$t$, $R_{t-1}$, $\gamma_t$ & registry exam index; earlier admissions; e-LOND sequence $t^{-1.6}/\zeta(1.6)$ \\
$\alpha_t=\alpha\gamma_t(R_{t-1}+1)$ & e-LOND level of exam $t$ \\
$k$, $q_k$ ($\gamma^{\text{lin}}_k$) & attempt index within a lineage; lineage schedule \\
$\min(\alpha_t,\alpha q_k)$ & deployed level \\
$J$, $\pi$ & attempt at which an honest author's first helpful version arrives; its law \\
$H(\pi)$, $\mathrm{KL}(\pi\|q)$ & entropy of $\pi$; mismatch between $\pi$ and the schedule \\
$K_{\inf}(P,m_0)$ & information per variant against the closest null \\
$C$ & expected variants an honest lineage spends on its helpful exam \\
$\beta$ & type-II error of an exam \\
$L=\log(1/a)$, $\rho=L/(L+1)$ & level in nats; grid ratio of the level-tuned bet \\
$\Lambda$, $w_j$, $c_u$ & distance of the best bet from the cap; grid weights; their normalizer \\
\bottomrule
\end{tabular}
\end{table}

%% file: sections/appendix_related.tex
\section{Extended related work}
\label{app:related}

\paragraph{Evaluating skills.} SkillsBench evaluates 87 tasks with and without curated skills across 18 model-harness configurations, three trials per cell, in pinned containers with deterministic verifiers, and reports Wald intervals for pass-rate differences \citep{li2026skillsbench}; its self-generated condition ran on 3 of the 18 configurations as a diagnostic. SWE-Skills-Bench runs paired with-and-without evaluations of 98 software-engineering skills and classifies each skill as golden, ineffective or harmful with an exact paired sign-flip test on ten requirement-level differences at $p<0.05$ \citep{han2026sweskillsbench}. This is a per-skill decision rule with a fixed sample size and no multiplicity correction. WebDev-Skills-Bench adds length-matched irrelevant-skill controls and leave-one-out ablations \citep{yang2026webdevskills}. ACES evaluates skills rather than agents with paired trials in which prerequisite and decoy skills are held fixed, a composite lift metric and confidence intervals, and leaves formal hypothesis tests to future work \citep{kevin2026aces}. SkillTester combines paired baseline and with-skill runs with a security probe suite \citep{wang2026skilltester}; SkillAudit, AEVAL and the framework of \citet{shaposhnikov2026framework} address skill-centred assessment, determinism and scale \citep{yu2026skillaudit,anand2026aeval}; the last reports lifts from 5.5 to 22 points across 19 configurations and finds that the Nemotron~3 models barely benefit. MASA finds that a skill that helps one backbone can harm another and adapts skills to the model \citep{yu2026masa}, and WebDev-Skills-Bench finds that skill rankings transfer weakly across models \citep{yang2026webdevskills}. The open-source skillcheck tool generates fresh tasks, grades blind and issues HELPS, PLACEBO or HARMS verdicts from bootstrap intervals \citep{sx4im2026skillcheck}. NVIDIA's trust pipeline gates signing on a scan and a with-and-without verdict \citep{nvidia2026trustpipeline,nvidia2026skillspector,nvidia2026skillevaluator}. Studies of the skill files themselves report widespread defects \citep{zhang2026reusable}, revision effects \citep{wang2026skillrevisions} and supply-chain attacks on registries \citep{saha2026skillmd}. We take the paired design from this literature and study what a registry decision built on it must satisfy.

\paragraph{Anytime-valid evaluation of LLM systems.} \citet{yin2026enough} build an anytime-valid detector on per-item paired outcomes to decide when a self-evolving LLM system, which updates skills and prompts, has stopped improving. This is the closest statistical neighbour; it targets a single system's evolution, not admission across a library with resubmissions. BB-EDGE gives anytime-valid leaderboards with e-processes and family-wise control over pairwise model comparisons \citep{gao2026bbedge}; AV-AIVAT applies anytime-valid stopping to agent-versus-agent evaluation \citep{li2026avaivat}; other work treats paired LLM evaluation as hypothesis testing with power analysis \citep{kotawala2026resolution} or monitors evaluation drift with a betting e-process \citep{li2026whodrifted}. A beta-binomial mixture e-process gives anytime-valid pairwise comparisons on incrementally revealed benchmarks \citep{kotawala2026pairwise}, CELEUS builds anytime-valid confidence sequences for benchmark accuracy with adaptive sampling \citep{zhou2026celeus}, and E-valuator turns agent verifier scores into sequential tests of trajectory success \citep{sadhuka2025evaluator}. None of these tests whether a skill helps, and none controls error across resubmissions. Our statistical tools come from the same literature: betting confidence sequences \citep{waudbysmith2024betting}, time-uniform bounds \citep{howard2021timeuniform}, e-values and safe testing \citep{ramdas2023gametheoretic,grunwald2024safe}, e-BH \citep{wang2022ebh}, online multiple testing with e-values \citep{xu2024online}, always-valid A/B testing \citep{johari2022always}, Ville's inequality \citep{ville1939etude} and Wald's SPRT \citep{wald1945sequential}.

\paragraph{Verifiable records.} Certificate Transparency \citep{laurie2013ct,laurie2021ctv2} and Sigstore's Rekor \citep{newman2022sigstore,sigstore2026rekor} provide append-only, publicly auditable logs; drand provides a public randomness beacon \citep{drand2026beacon}. For agents, \citet{bromme2026blackbox} anchors agent evidence on a blockchain and identifies capture authenticity as open; NovaFabric, notarized receipts and proof-of-behavior drafts give tamper-evident logs \citep{seyedkazemi2026novafabric,figuera2026notarized,dembowski2026agentledger}; ERC-8004 defines agent identity, reputation and validation registries \citep{derossi2025erc8004}. Zero-knowledge proofs can verify evaluation claims about hidden models \citep{south2024zksnarks}, and decentralized benchmark verification has been proposed \citep{pardasani2026whoverifies}. Patent filings describe credentialing an agent against a rubric with per-test hashes on a ledger \citep{drivehealth2026credentialing,drivehealth2026targeted}, comparing agent fingerprints to ledger hashes \citep{mysore2026citibank}, provenance of model assessments \citep{chen2023credo}, and committing to a model before revealing test data \citep{dayama2022ibm}. Our credential differs in binding both arms of a paired test and the sequential decision, so that the decision itself can be recomputed.

\paragraph{Skill synthesis.} Single-agent skill libraries admit skills by self-verification \citep{wang2023voyager,zhao2024expel,shinn2023reflexion,zheng2025skillweaver}; multi-agent and evolutionary methods distill or evolve skills from trajectories \citep{ni2026trace2skill,du2026multievoskill,li2026mindskill,ma2026skillv,ma2026skillneo,liu2026skillvine,hu2026selm,liu2026rethinking}. Any of these could supply candidates to the gate studied here.

%% file: sections/appendix_results.tex
\section{Additional results}
\label{app:more}

\begin{figure}[!htb]
\centering
\includegraphics[width=0.64\linewidth]{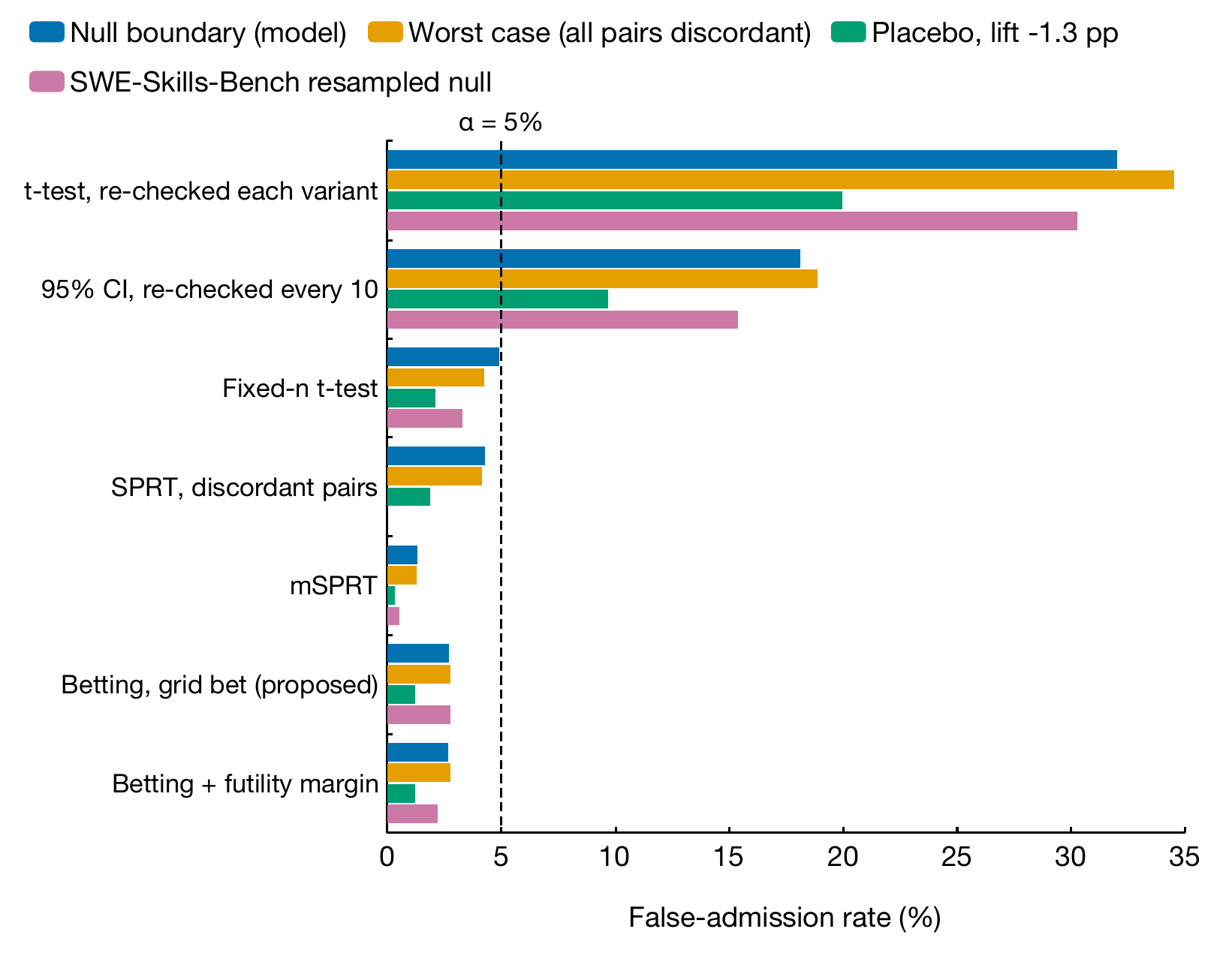}
\caption{False-admission rate under nulls (E1, E8). Model nulls use \PReps{} replicates each; the SWE-Skills-Bench resampled null draws differences from the released data with random signs. The SPRT is defined only for pass/fail data and is not run on the batch pass rates of SWE-Skills-Bench. Dashed line: $\alpha=5\%$.}
\label{fig:false}
\end{figure}

\paragraph{Validity checks.} Validity also holds for strong and weak common random numbers, for $r=3$ attempts per arm, and for a margin $\delta=0.05$. The released SWE-Skills-Bench data are tie-heavy, which makes the exact sign-flip test conservative; on continuous data the per-configuration rate would be near 5\% and the any-of-four rate near 19\% (\NShopRatio-fold on the released data). A one-sided version of the source rule marks \NSweOneSided{} cells.

\paragraph{Departures from the conditional null (E21).} E21 simulates exams of a useless skill ($\Delta=0$ before any departure, $\delta=0$) under each departure listed in Section~\ref{sec:discussion}, with 20{,}000 exams per cell (Monte Carlo standard error at most 0.4 points) and the default outcome model. The exam-level coin fixes the lift of a whole exam at $\pm s$; a model change at a uniform variant index shifts both arms' logits by $\pm1$, or the skill arm's by $\pm0.5$; shared effects add a normal logit effect to the skill arm, common to blocks of variants, with the base lift re-solved so that the marginal lift stays zero; repeated seeds copy an earlier variant's outcome; and exclusion drops variants the skill arm lost, as if they were harness failures. Table~\ref{tab:nullstress} gives the false-admission rates. The fixed-$n$ $t$-test admits nothing under the hard coin because its statistic is undefined when every difference is equal.
\begin{table}[!htb]
\centering\small
\caption{False-admission rate (\%) of a useless skill under departures from the design (E21), target 5\%. Grid and PrPl use the futility margin.}
\label{tab:nullstress}
\input{tables/t_null_stress.tex}
\end{table}

\paragraph{Total lineage cost (E19).} Theorem~\ref{thm:price} prices the helpful exam of a lineage under the lineage term alone. E19 measures the whole cost under the deployed level $\min(\alpha_t,\alpha q_k)$, with the useless versions, the cap $N_{\max}=600$, futility, and e-LOND levels for a lineage that opens at registry index 50, 200, 500 or 2{,}000, submits one version per ten registry exams, and finds one earlier admission per ten exams. An honest lineage submits $J-1$ useless versions and then a helpful one, $J\sim\pi$ geometric with mean 3, Zipf with exponent 2, or uniform on $1,\ldots,20$; schedules $q$ are $\zeta(1.6)$, the universal code, uniform on 20 and the oracle $q=\pi$; 2{,}000 lineages per cell, grid bet. Three results. (i) The lineage term sets the level in at most \NLcBindMax\% of exams, so the four schedules give the same costs to within \NLcSchedSpread\%; the e-LOND level, not the schedule, sets the price, and it grows with the registry: the helpful exam at lift 0.10 costs \NLcHelpAgeFifty{} variants at index 50 and \NLcHelpAgeTwoK{} at index 2{,}000, and its power falls from \NLcAdmAgeFifty\% to \NLcAdmAgeTwoK\%. (ii) The cost of the helpful exam is predicted by an intercept plus a slope times $\mathbb{E}\log(1/\text{level}_J)$, fitted once per lift, with mean error \NLcPredErrMean\% and maximum \NLcPredErrMax\% over 192 cells, so the $\log(1/\text{level})$ form of the theorem holds for the deployed gate with its cap and futility. (iii) Useless versions cost more than the helpful one: with $J$ geometric, \NLcGeoUseless{} variants against \NLcGeoHelp{} (lift 0.10, index 500, $\zeta(1.6)$), and with $J$ uniform on 20, \NLcUniUseless{} against \NLcUniHelp. A useless version at $\Delta=-0.05$ stops for futility much sooner than one at $\Delta=0$ (\NLcUselessNeg{} against \NLcUselessZero{} variants per lineage on average). No null lineage was admitted more than \NLcNullMax\% of the time, and no honest lineage had a useless version admitted more than \NLcHonFalseMax\% of the time.

\paragraph{The unit of error (E22).} E22 simulates 400 registry histories of 200 exams (ten windows of 20) under the convention of Section~\ref{sec:gate}. Sixty percent of lineages are null, with one to eight useless versions. The rest are honest: some useless versions, then a helpful one ($\Delta\sim U(0.05,0.25)$), then up to two replacement edits tested against the admitted version, each either useless or a further improvement ($\Delta\sim U(0.05,0.15)$). Every history contains lineages admitted twice, once for the helpful version and once for an improvement. Under the deployed level, with the grid bet, the FDR over exams is \NFuFdr\% (\NFuAdm{} admissions per history), a null lineage is admitted with probability \NFuNullLin\%, and an honest lineage has one of its useless versions, before or after its admission, admitted with probability \NFuHonFalse\%, while \NFuHonAdm\% of honest lineages have a helpful version admitted. In the first history, for example, lineage 0 had its helpful version admitted at attempt 1 (level 0.022, 272 variants); its two useless edits were tested against it at levels 0.0010 and 0.00046 and rejected, one by futility after 394 variants and one at $N_{\max}$.

\paragraph{Grid against PrPl, and other e-BH variants (E2, E20).} On the E3 histories with the grid bet, e-BH on adjusted suprema matched e-LOND in power (\NEbhGridSupPow\%) at \NEbhGridSupN{} variants per exam, and running every exam to $N_{\max}$ before e-BH gave \NEbhGridNmaxPow\% at \NEbhGridNmaxN{}; with PrPl the order is the same, but adjusted suprema lose more power (\NEbhPrplSupPow\%). The grid bet is faster at moderate and large lifts (at the curated-skill lift it needs \NGbCurGrid{} variants against \NGbCurPrpl), has less power at small lifts (\NETwoPowBetHalf\% against \NETwoPowBetHalfPrpl\% at a lift of 0.05) and admits a useless skill somewhat more often, while staying below $\alpha$ (\NETwoNullBet\% against \NETwoNullBetPrpl\% at the boundary).

\paragraph{Cost details (E2, E3).} Capping the betting gate at $N_{\max}=300$ in E2's runs gives \NETwoPowBetThree\% power at the design lift for \NETwoRunsBetThree{} agent runs, against \NETwoRunsFixed{} for the fixed design. With the futility margin a skill with $\Delta=-0.05$ runs \NETwoNBetFutNeg{} variants; the SPRT runs \NETwoNSPRTZero{} for a useless skill. In the SWE-like library each truly helpful admission costs \NEFiveSweRunsBet{} agent runs with the per-exam betting gate. In E3, e-LOND with the slowly decaying default sequence reaches \NEThreePowELond\% power (E20's separate histories give \NEbhGridElondPow\%). With a re-checked interval, a resubmitted useless skill is admitted within five attempts with probability \NResubFivePeekCI\%.

\begin{table}[!htb]
\centering\small
\caption{Calibrated library scenarios (E5). False share: fraction of admitted skills with $\Delta\le\delta$ (here $\delta=0$). Runs count both arms. Betting rows use the grid bet.}
\label{tab:scen}
\resizebox{\linewidth}{!}{\input{tables/t_scenarios.tex}}
\end{table}

\begin{figure}[!htb]
\centering
\includegraphics[width=0.48\linewidth]{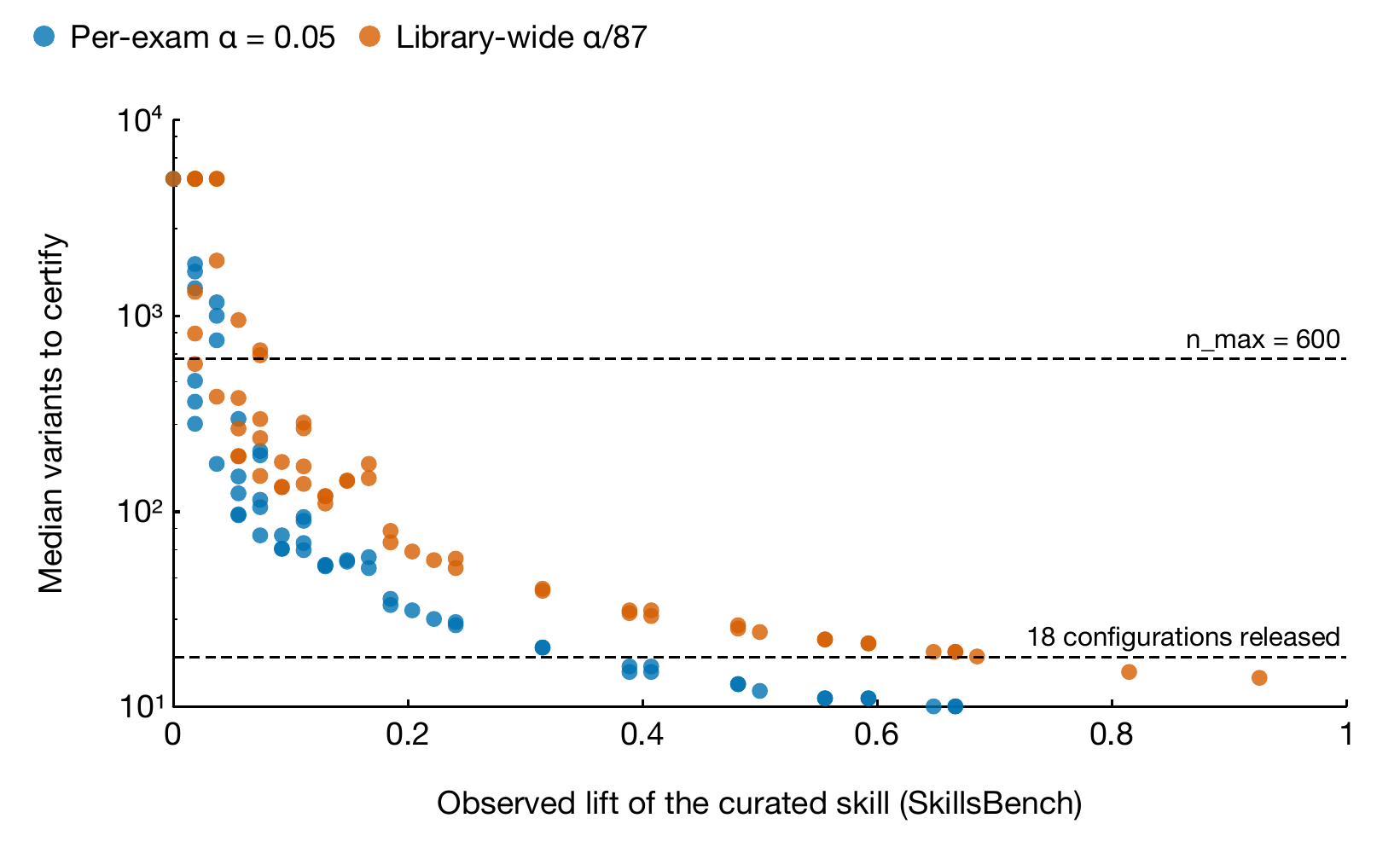}
\caption{Median variants needed to certify each SkillsBench curated skill with positive observed lift (\NSbPos{} skills), by resampling its released per-configuration differences (E9; 1{,}000 resamples per skill). Points at the top did not certify within 2{,}000 variants.}
\label{fig:need}
\end{figure}

\begin{figure}[!htb]
\centering
\includegraphics[width=0.48\linewidth]{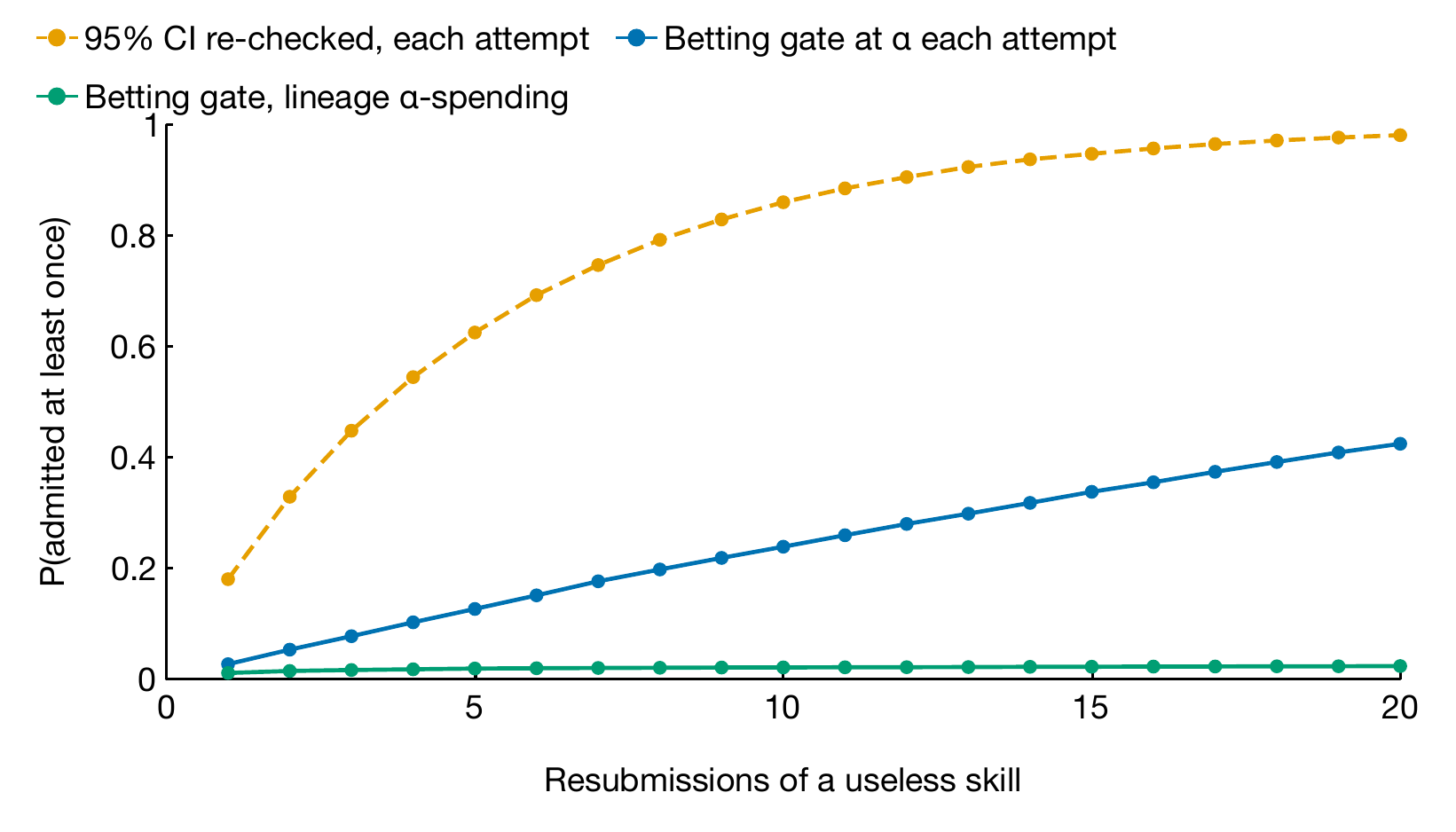}
\caption{Probability that a useless skill ($\Delta=0$) has been admitted after $k$ resubmissions, each a fresh exam (E3).}
\label{fig:resub}
\end{figure}

\paragraph{Ten units (E8).} With the PrPl cap $c$ the wealth after $n$ units is at most $(1+c/2)^n$, so reaching $1/\alpha=20$ needs 14 units at $c=1/2$ (ten give at most \NSweMaxWealth) and all ten differences equal to 1 at $c=3/4$ (wealth \NSweMaxWealthHi; the largest any cell reached was \NSweObsWealth). With the grid bet seven straight wins suffice, and the largest wealth any cell reached was \NSweObsWealthGrid.

\paragraph{Fixed-sample library procedures and the theory checks (E10--E12).} E10 applies Benjamini-Hochberg, Benjamini-Yekutieli and Bonferroni at $\alpha=0.05$ to one-sided paired $t$-test and exact sign-flip $p$-values for the \NSbTasks{} SkillsBench skills, on all 18 configurations and on 200 random half-splits. On all 18 configurations, exact sign-flip tests certify \NFnSfBH{} skills under Benjamini-Hochberg, \NFnSfBY{} under Benjamini-Yekutieli and \NFnSfBonf{} under Bonferroni ($t$-tests: \NFnTBH, \NFnTBY{} and \NFnTBonf). Decided on nine configurations, Benjamini-Yekutieli certifies \NFnSplitTBY{} skills with $t$-tests and none with sign-flip tests, because with nine pairs the smallest exact sign-flip $p$-value, $1/512$, exceeds the Bonferroni threshold $0.05/87$; sign-flip tests with Benjamini-Hochberg certify \NFnSplitSfBH{} skills on average from nine configurations. E11 takes each of the \NThHelpful{} skills with positive released lift, uses its 18 per-configuration differences as $P$, computes $K_{\inf}$ by maximizing $g_P$, builds $Q^*$, and resamples 400 exams per skill at $\alpha=0.05$ without truncation or futility for the Kelly bet, PrPl with $c=1/2$, and aGRAPA with $c=1/2$ and $c=0.9$ (median $\NThAgrapaHiRatio\times$ the lower bound). The lineage checks resample 2000 exams of the median skill (\NThMedTask) at levels $\alpha q_k$ for three schedules. Attempt-count laws are truncated at 200 attempts. E12 checks Section~\ref{sec:theory}'s schedule results. It computes worst-case excesses over $\Pi_N$ for $N$ from 5 to 1{,}000, and certified intervals for $R(\Pi_N^\downarrow)$ for $N$ from 10 to 3{,}000 (width below $10^{-3}$ nats). It estimates $\mathbb E\,\mathrm{KL}(\pi\|\hat q)$ for the learned schedule of Theorem~\ref{thm:learned} from 4{,}000 draws of the counts, with $N=20$, $\epsilon=0.05$, three laws $\pi$ (geometric with mean 3, Zipf with exponent 2, uniform), $m$ from 0 to 1{,}000 and 0, 10 or 50 poisoned counts at the least likely attempt; every estimate is below the bound of part (b). For variants, it simulates 2000 exams of the median skill with the Kelly bet at 17 levels; expected variants grow linearly in $\log(1/a)$ with slope \NScSlope, against $1/K_{\inf}=\NScOneOverG$, and lineage costs are read off this curve.

\paragraph{Details of E14 and E18.} The matched-power runs (E18) use the default model at $r=1$, \PRepsPower{} replicates per cell, the SPRT with type-II error $\beta\in\{0.20,0.10,0.05,0.01\}$, the betting gate with futility margin truncated at $N_{\max}\in\{150,200,300,600\}$, and the fixed-$n$ test planned for 80\% to 99\% power. The moment fit searches a grid of $s_\theta$ from 0.5 to 5, $s_g$ from 0 to 3 and $\rho$ from 0 to 0.95, then a finer grid around the best point, matching each candidate's moments on $10^5$ simulated differences at the released lift and comparator rate with a relative squared-error loss. The fitted $\rho$ sits at the edge of the grid. E23 widens the grid ($s_g$ up to 5, $\rho$ up to 0.999): the fit keeps $s_\theta=s_g=2.25$, now in the interior, and moves $\rho$ to 0.999, again at the edge, with a loss of 0.0149 against 0.0160. The fitted-model costs move by at most 4\% (for example 166 against 170 expected variants at lift 0.10, $r=1$, $\kappa=0$), and the range over all sensitivity cells is 37 to 389 variants against 37 to 392, so no cost conclusion depends on where the grid stops. The correlated model draws $g_i=g+s_g(\kappa z_i+\sqrt{1-\kappa^2}z_i')$ with $\theta_i=s_\theta z_i$ and solves for $g$ so that $\Delta$ is exact; each cost cell uses 3{,}000 replicates. For the library procedure, each of the 87 skills gets 400 resampled exams of up to 2{,}000 variants at each of 25 levels from $\alpha$ down to $10^{-6}\alpha$; an exam at an e-LOND level uses the largest grid level not above it, which can only overstate the cost. Each of 1{,}000 random arrival orders updates the e-LOND rejection count with admissions within 600 variants. On the same grid, the Bonferroni level gives a median of \NRvLibBonf{} variants and the per-exam level \NRvLibPerExam, close to the \NSbMedNBonf{} and \NSbMedNAlpha{} of E9.

\paragraph{Level-tuned grid (E13).} For each of the \NThHelpful{} skills, E13 runs the grid exam of Theorem~\ref{thm:adaptive}, the Kelly bet, the same grid with $\rho=1/2$, the PrPl bet ($c=1/2$) and aGRAPA on 1{,}000 shared resampled sequences at the levels $\alpha q_1$ and $\alpha$, with the grid truncated at $\lambda_j\ge10^{-4}/m_0$ (truncation drops weight and so keeps validity). At level $\alpha q_1$ the grid needs fewer variants than aGRAPA for \NAgGridBeatsAgrapa\% of the skills, and every simulated cost lies below the bound. For the median skill, 1{,}500 sequences per level give $K_{\inf}\mathbb E[\tau]-L$ for 17 levels; the Kelly bet's excess is at most \NAgKellyExMax{} nats. Validity is checked with 4{,}000 sequences of 2{,}000 draws from each skill's least-favourable null $Q^*$: the mean rejection rate is \NAgMeanReject\%, at level \NAgLevelQone\%.

\paragraph{Reported lift.} The stopped estimate overstates the lift by \NEFourRelBiasFive\% at $\Delta=0.05$ and \NEFourRelBiasTen\% at $\Delta=0.10$. A fixed-$n$ test reported only when significant has the same problem (\NEFourBiasFixedFive\% at $\Delta=0.05$). At admission the anytime-valid lower bound averages at most \NEFourLcbMax{} points.

\begin{figure}[!htb]
\centering
\includegraphics[width=0.48\linewidth]{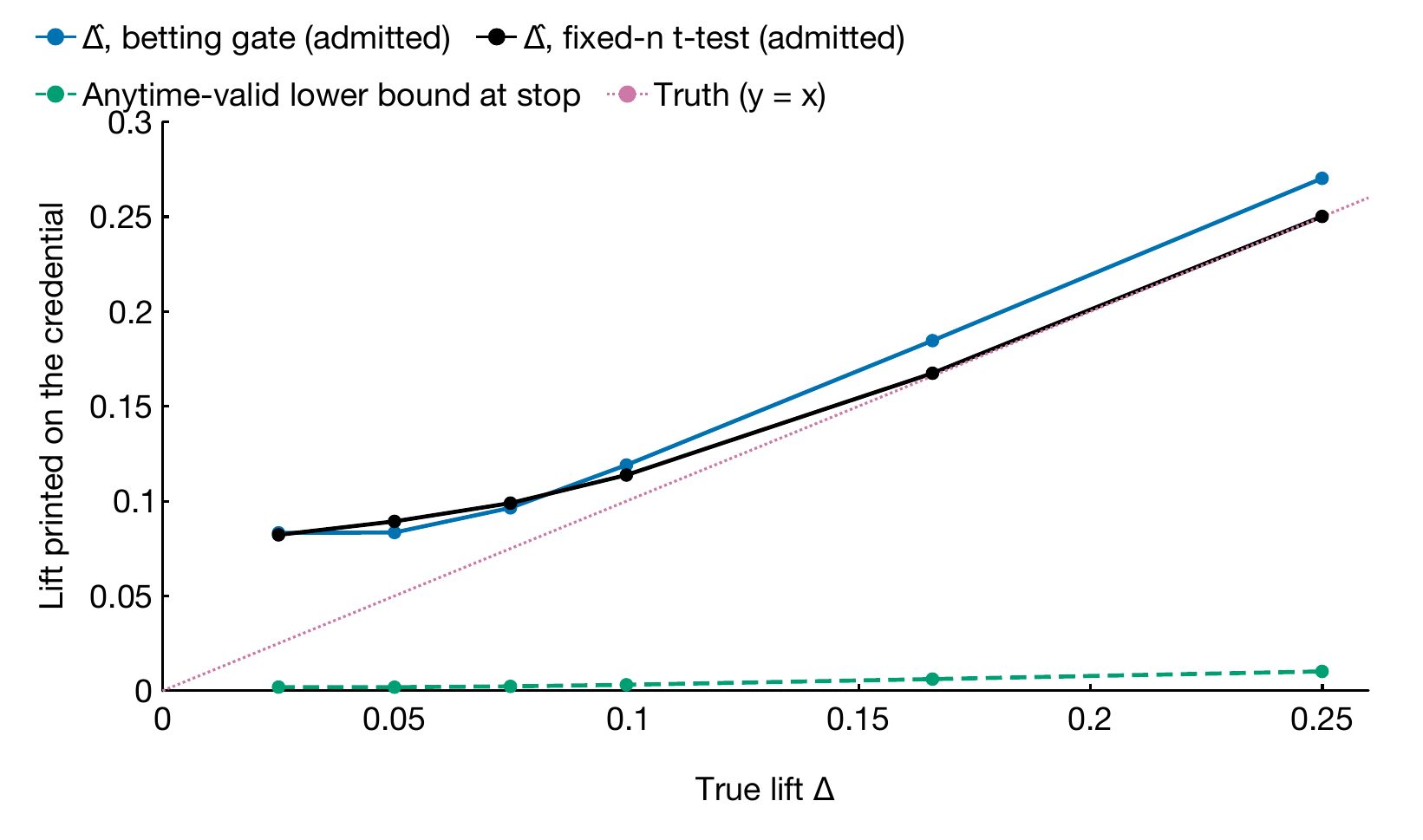}
\caption{Reported lift for admitted skills against the true lift (E4, 3{,}000 replicates per point).}
\label{fig:wc}
\end{figure}

\paragraph{Leakage and verification.} Tables~\ref{tab:leak} and~\ref{tab:verify} plant lookup tables covering 5\% of a \PGenSpace-instance generator space and flag a skill when its estimated coverage exceeds \PCoverMax\%. Normalizing separators and quotes catches spaced and split literals, and decoding base64 and hex runs catches encoded tables. On clean skills the false-positive rate was \NEsixFPLiteral\% for every detector; short numbers in honest text, such as status codes, gave a mean estimated coverage of \NEsixCleanCover\%. Our clean skills are templates, so this rate is a lower bound for real skills. For \NESevenN{} variants with \PTranscriptKB{}~kB transcripts (\NESevenMB{}~MB) the proctor's bookkeeping takes \NESevenProctor{}~s.

\begin{table}[!htb]
\centering\small
\caption{Leakage-check recall (\%) on planted tables, and false-positive rate on clean skills (E6).}
\label{tab:leak}
\input{tables/t_leakage.tex}
\end{table}

\begin{table}[!htb]
\centering\small
\caption{Credential cost (E7), median of three runs.}
\label{tab:verify}
\input{tables/t_verification.tex}
\end{table}

\paragraph{Live pilot.} We ran the full protocol once on a real agent: OpenHands with Claude Haiku~5.5 on \NPilotTasks{} SkillsBench tasks drawn by the committed rule from drand round \NPilotRound, three trials per arm, with both arms running concurrently. The model was changed by a logged amendment before any run; the amendment drew fresh secrets and a new beacon round. Two tasks errored in both arms (an agent crash and a sandbox that would not start) and were reported, not replaced, as pre-registered. On the \NPilotScored{} scored tasks the skills changed no pass count (mean $D=\NPilotDbar$, final wealth \NPilotWealth), so nothing was admitted. A separate run of the verifier opened both commitments, re-derived the task sample from the beacon, recomputed the decision from the logged transcripts and checked the signature; changing one recorded reward made it fail. The pilot cost about US\$2.5. % pilot/jobs/*/summary.json: 2.54
It demonstrates the protocol, not the effect of skills: four paired units are far too few (Section~\ref{sec:res-cost}).

\paragraph{Security screen (E15).} Helpfulness and safety are separate admission requirements \citep{wang2026skilltester}. We ran a static scanner, SkillSpector \citep{nvidia2026skillspector}, offline and without a language model, on the \NSecSkills{} skills shipped with the 87 SkillsBench tasks. It rated \NSecSafe{} safe, \NSecCaution{} caution and \NSecDNI{} do-not-install, the last in \NSecTasksFlag{} tasks, \NSecFlagHelpful{} of which have positive released lift. Risk score and released lift are unrelated (Spearman $\rho_S=\NSecRho$, $p=\NSecP$, 87 tasks). We scanned a clean export of the registered commit: a first scan of our working copy also flagged a skill for compiled bytecode that our own runs had left there. Each flagged skill was read three times by language models, each reading blind to the others and made from the skill files and scanner output alone: twice with Claude and once with GPT-5.5 (\texttt{gpt-5.5-2026-04-23}). All three readings found the same \NSecGenuine{} real concern among the \NSecFlagN{} flags, and none found another: a penetration-testing skill that instructs the agent to brute-force logins, collect private keys and add its own key to \texttt{authorized\_keys}, none of which its task needs. The readings rated the rest false positives or the result of files the scanner could not parse; the two Claude readings agreed on that category for \NSecCatAgree{} of \NSecFlagN{}, and GPT-5.5 called all of them false positives. A helpfulness certificate is therefore no evidence of safety, and the credential should carry the scan (tool version, rules, score) as a separate field.

\subsection{Live studies 1 to 3 in detail (benchmark run, logs family, orders family)}
\label{app:live-detail}
\paragraph{On the benchmark.} We first ran the committed protocol on all \NLoTasks{} SkillsBench tasks with OpenHands and Claude Haiku~5.5 (API model identifier \texttt{claude-haiku-5-5}, runs on 15 and 16 November 2026), three trials per arm, in an order derived from drand round \NLoRound. The test asks whether the curated skills, taken as one library, help on average: $D_i$ is the per-task difference in passes divided by three, tested with the PrPl betting gate (cap $1/2$), as committed, at $\alpha$ in the committed order. As committed, a task with an errored run in either arm was reported and excluded. \NLoExcluded{} tasks were excluded, for sandbox start-up refusals by the harness, agent-protocol and image failures, and a provider-side interruption during which model requests were rejected; \NLoMissing{} were never run because their environments request access to the host, which we did not grant. One logged amendment raised the budget after the first batch, whose results we had seen; a second reran two batches that the harness had refused as a whole. Continuing a betting test after seeing its path does not affect its validity (Proposition~\ref{prop:valid}), and both amendments are entries in the hash-chained ledger. On the \NLoScored{} scored tasks the mean difference was $\bar D=\NLoDbar$ (\NLoPos{} positive, \NLoNeg{} negative), the final wealth was \NLoWealth, short of $1/\alpha=20$, and the anytime-valid lower bound was \NLoLcb, so nothing was admitted. The verifier opened the commitment, re-derived the order from the beacon, recomputed the statistics from the \NLoRuns{} logged runs and checked the signature. It cost about US\$\NLoCost. The exclusions are not random: tasks whose images the harness refuses differ from those it runs.

\paragraph{On generated variants.} In the first family a variant is an access log of 60 to 90 lines generated from a 32-byte seed, and the task asks which client had the most server errors and how many. The logs follow two house conventions that cannot be inferred without documentation: the first field names the edge node that wrote the line, with the client in a trailing field, and status 599 records a client disconnect rather than a server error. The generator guarantees a unique answer, and in 95\% of variants the naive reading gives a wrong one.

Four skills were registered and hashed before the beacon round: a helpful skill that documents the conventions; a placebo of similar length with generic log-reading advice; a planted leak, which adds to the placebo a table of answers for 300 public example seeds; and a harmful skill that states wrong conventions. The model was Claude Haiku~5.5 (\texttt{claude-haiku-5-5}), called through the API with default settings on 16 November 2026. For each variant both arms answered the same log concurrently, and a pass is an exact match of client and count. The four exams ran in a committed order that includes a resubmission: the placebo, the leak and the harmful skill opened lineages A, B and C, and the helpful skill was submitted as the second attempt of lineage A. Exam $t$ ran at level $\min(\alpha_t,\alpha\gamma^{\text{lin}}_k)$ with e-LOND levels $\alpha_t$ (Proposition~\ref{prop:lineage}), with the PrPl betting gate (cap $1/2$), as committed, and the futility margin and $N_{\max}$ of Section~\ref{sec:gate}. Variant $j$ of exam $t$ used a seed derived from drand round \NLsRound, which did not exist when the commitment was logged. Before committing we ran five public variants per skill to check the harness (US\$\NLsCalCost); these were not scored, and we saw their results.

The leak table gave no lift (\NLsLeakWith{} passes with the skill, \NLsLeakWithout{} without): the exam variants came from seeds that did not exist when the skill was registered, so the table covered none of them. The harmful skill could not be told apart from the placebo, because the model rarely solved the task unaided (\NLsBasePct\% of variants without a skill) and so had little room to get worse; showing harm needs a family with a higher baseline. During the placebo exam a provider-side interruption rejected requests for a period; the \NLsPlacExcl{} affected pairs were reported and excluded, as committed, and the exam resumed at the next variant index. The study used \NLsPairs{} paired variants and cost US\$\NLsCost{} in API fees.

\paragraph{A second family and a resubmission attack.} We therefore ran a second committed study (drand round \NLsThreeRound) as a new registry history. Its family is an order export in CSV in which a channel code marks internal test orders that must be excluded from revenue, a convention the data do not reveal. The code changes the answer in about half the variants, so the model can succeed without documentation about half the time. A first version of this family, in which a numeric code marked refunds, was abandoned after calibration because the model guessed the code; both calibrations are in the ledger. Three skills were registered: a helpful skill that documents the code, a placebo of similar length, and a harmful skill that states a plausible but false pricing rule. The helpful skill was again submitted as the second attempt of the placebo's lineage. Finally, the placebo of the first study was resubmitted unchanged five times in one lineage, to exercise the level bookkeeping on a live model.

Without a skill the model passed \NLsBBasePct\% of the family's variants. The gate admitted the helpful skill after \NLsBHelpN{} variants (\NLsBHelpWith{} passes with the skill, \NLsBHelpWithout{} without, over \NLsBHelpRun{} variants run) and stopped the placebo for futility after \NLsBPlacN. The harmful skill was stopped for futility after only \NLsBHarmN{} variants (passes at the decision: \NLsBHarmTauWith{} with the skill, \NLsBHarmTauWithout{} without); the skill lowered the pass rate in every discordant variant (12 of 12; exact one-sided sign test after the fact on the \NLsBHarmRun{} variants run, $p=0.0002$). At \NLsBHarmN{} variants the anytime-valid upper bound on $\Delta$ was still \NLsBHarmUcb, above zero, so the gate refused the skill without certifying that it harms; a registry that wants to flag harmful skills needs a separate test with its own error budget. The five resubmissions ran at levels falling from \NLsResubLevelFirst{} to \NLsResubLevelLast{}, set in every case by the e-LOND term rather than the lineage term, as in most exams of E3b; \NLsResubAdmit{} were admitted, and at a per-exam level $\alpha$ without lineage spending \NLsResubAdmitPer{} would have been. With and without the placebo the model gave the same pass or fail on every variant, so each exam saw $D=0$ throughout and stopped for futility after \NLsResubNmin{} variants. Because the two arms never differed, any gate would have refused these attempts, so this part of the study is a check of the implementation, not evidence for lineage spending, whose error rate is a probability over histories (Section~\ref{sec:res-library}). The study cost US\$\NLsThreeCost, and its credential verified as in the first.

\paragraph{A second model.} Both studies used one closed model. To check that the gate's behaviour does not depend on it, we repeated them with an open-weight model from another family, Nemotron~3.5 Super VL (NVIDIA, released 15 October 2026; OpenRouter identifier \texttt{\NLnModel}), as two new registry histories committed before drand round \NLnRound{} on 17 November 2026. Everything but the model was kept: the skills, generators, prompts, graders, exam orders, levels and gate, all fixed by hash in the commitment. The model was called through OpenRouter with the provider pinned to the only endpoint that served the unquantized (bf16) weights and with fallbacks disabled, so that a response from any other provider would have counted as an error. Nemotron reasons at length before it answers, and its reasoning shares the token budget with the answer; in a smoke test on public variants, 6 of 20 answers were cut off at 8,000 tokens, so we committed a limit of \NLnMaxTok{} tokens for this model. Even so, \NLnCut{} of \NLnCalls{} calls (\NLnCutPct\%) reached the limit without a final answer and were graded as fails, as the pass rule requires. This is part of the model's measured behaviour under the committed settings; it lowers the pass rates the gate sees but not the validity of its decisions.

With Nemotron, every decision matched the one made with Haiku. In the second family the harmful skill lowered the pass rate in all \NLnBHarmDisc{} discordant variants (exact one-sided sign test after the fact, $p=\NLnBHarmSignP$), but the anytime-valid upper bound on $\Delta$ at the decision was still \NLnBHarmUcb, so the gate again refused the skill without certifying harm. Unlike Haiku, Nemotron did not always give the same result with and without the resubmitted placebo: \NLnResubDisc{} variants across the five attempts were discordant, so these exams saw nonzero differences. All five still stopped for futility, after \NLnResubNmin{} to \NLnResubNmax{} variants, and none would have been admitted at a per-exam level $\alpha$. Since a skill that helps one model can fail to help another \citep{yu2026masa,yang2026webdevskills}, a credential states the model under which the skill was admitted. The two studies used \NLnPairs{} paired variants; \NLnFailed{} calls failed and their pairs were excluded as committed. They cost US\$\NLnTotalCost{} in API fees, and both credentials verified.

\subsection{Study 4: published skills, selection and design}
\label{app:live-extra}
\paragraph{Shortlist and selection.} We searched public skill repositories for instruction-only skills (no bundled scripts) by authors other than the model providers, under an open licence, whose rules a program can check exactly. Three were shortlisted and ranked before any call, each pinned to a commit and a SHA-256 hash checked on load: \texttt{newsroom-style} by Joe Amditis (AP style; MIT), \texttt{commit} by Sentry (commit messages; Apache-2.0) and \texttt{apa7} by Kalaris Labs (APA~7; MIT). The selection rule, the window $[0.05,0.40]$ and the limit of one widening of the shortlist were fixed in the design document before calibration. Table~\ref{tab:s4cal} gives every calibration; their transcripts are in the ledger.

\begin{table}[tb]
\centering\small
\caption{Public calibrations for study 4 (20 variants per arm, seeds public, not scored). Passes with / without the skill. Selection used the Haiku lift only.}
\label{tab:s4cal}
\begin{tabular}{lllrrr}
\toprule
Skill & Model & Role & Published skill & Placebo & Lift \\
\midrule
AP style & Haiku & study 4b & \NCalFourBRealWith{} / \NCalFourBRealWithout & \NCalFourBPlacWith{} / \NCalFourBPlacWithout & \NCalFourBLift \\
Sentry commit & Haiku & study 4c & \NCalFourCRealWith{} / \NCalFourCRealWithout & \NCalFourCPlacWith{} / \NCalFourCPlacWithout & \NCalFourCLift \\
APA~7 & Haiku & study 4 (selected) & \NCalFourRealWith{} / \NCalFourRealWithout & \NCalFourPlacWith{} / \NCalFourPlacWithout & \NCalFourLift \\
APA~7 & Nemotron & study 4 & \NCalNFourRealWith{} / \NCalNFourRealWithout & \NCalNFourPlacWith{} / \NCalNFourPlacWithout & \\
\bottomrule
\end{tabular}
\end{table}

\paragraph{Task families.} In the AP family a variant is a set of reporter's notes written in non-AP form, and a brief passes if every graded item (dates, times, counts, ages, money, ``over'', attribution verbs) takes the form the skill prescribes. In the commit family a variant is a set of change notes, and a message passes if its header, type, subject, line lengths, breaking-change marker, issue footer and omission of a customer's contact details follow the skill. In the APA family a variant is a journal-article record with 1 to 30 authors, a title in title case and a DOI in a non-URL form; a pass requires the exact reference-list entry (including sentence-case title, en dash in the page range, \texttt{https://doi.org/} form and the rule for 21 or more authors, with the APA manual's punctuation ``nineteenth author, . . . final author'', which the skill states only as ``an ellipsis'') and the exact parenthetical citation, after a fixed normalization of whitespace, markdown emphasis and quotation marks. Each placebo is general advice on the same activity (news writing, commits, academic writing) of matched length, with none of the graded rules. In the AP family that was not enough to make the placebo inert. With it Haiku wrote ``over'' for ``more than'' in \NLsFourBPlacOverWith{} of \NLsFourBPlacOverN{} briefs that needed it (\NLsFourBPlacOverWithout{} without) and money in AP form in \NLsFourBPlacMoneyWith{} of \NLsFourBPlacMoneyN{} (\NLsFourBPlacMoneyWithout{} without). On Nemotron the placebo moved the same rules: ``over'' in \NLnFourBPlacOverWith{} of \NLnFourBPlacOverN{} briefs (\NLnFourBPlacOverWithout{} without) and money in \NLnFourBPlacMoneyWith{} of \NLnFourBPlacMoneyN{} (\NLnFourBPlacMoneyWithout{} without). Without the placebo the models almost never produced these forms. A control that cannot help must come from an unrelated activity; a placebo from the same activity tests whether the specific rules add to general advice. The commit and APA placebos keep the published skill's name in their header (\texttt{commit}, \texttt{apa7}); the AP placebo is named \texttt{newsroom-craft}, so the model saw a different name in that study only. Titles in the APA family are filled from fixed templates and are sometimes ungrammatical (``a experimental study''); both arms see the same record.

\paragraph{Power plan.} The plan simulated the grid-bet exam at the paired outcome law estimated from the Haiku calibration and at a conservative law, and chose the first design (placebo first, then real skill first; $N_{\max}$ from 600 to 1800) with conservative power at least $0.80$. The first version of the conservative law lowered the lift by one standard error but ignored variants in which the skill turns a pass into a fail; the placebo calibration showed such variants occur, so before commitment we amended it to add the placebo's discordance rate (\NPowNoise{} per side) to both sides. Conservative power fell from \NPowConsVone{} to \NPowCons, the design was unchanged, and both versions are in the record. Studies 4b and 4c, with large calibration lifts, used the design of studies 2 and 3 without a simulation.

\paragraph{Excluded pairs.} Only the Nemotron arm of study 4 lost pairs: all \NLnFourPlacExcl{} exclusions in its placebo exam (of \NLnFourPlacRun{} pairs run) and all \NLnFourRealExcl{} in its skill exam (of \NLnFourRealRun) were caused by rate-limit errors (HTTP 429) from the pinned provider. Failed calls were \NLnFourPlacFailWith{} with the skill and \NLnFourPlacFailWithout{} without it in the placebo exam, and \NLnFourRealFailWith{} and \NLnFourRealFailWithout{} in the skill exam. The committed rule excludes a pair whose call fails in either arm after retries. A rate-limit error is returned before the model produces anything, so exclusion cannot depend on the outcome the failed call would have had, and the decision uses the remaining pairs in their original order. In the same arm \NLnFourCut{} answers reached the token limit and were graded as fails.

\paragraph{Deviations from the protocol, and why they do not affect the decisions.} Three things did not go as written; each is in the ledger with the logs of the failed attempts.

\emph{(i) Wrong labels.} A substitution error in the script that generated the runners of studies 4b and 4c made the descriptive label of their Nemotron arms read ``4bbN'' and ``4ccN''. The label is text only: the ledger entry kinds, seeds, transcript folders and all hashed objects are correct. We left it as committed, because changing it would change the hash the verifier checks.

\emph{(ii) A run that stopped before it started.} The first attempt to run the Nemotron arms of 4b and 4c stopped before writing any transcript, because the transcript folder did not exist. The rerun regenerated the same seeds; no outcome of the first attempt was recorded or seen.

\emph{(iii) An early reveal.} In the Nemotron arm of study 4 the exams were first started before the committed beacon round existed, and a build step run by mistake afterwards logged the reveal of the committed secret at 16:05:45 UTC, nine seconds before beacon round 34071530 was published and before any exam ran. The secret keeps the hidden variants unknown both to whoever chooses the skills, generator and grader and to a submitter during the exam, on which resubmission-proofness relies. Neither role was compromised: the variant seeds hash the secret with the beacon, so they could not be computed before the round existed; both attempts of the lineage and the generator, prompt, grader and plan were fixed by hash in a commitment logged two minutes earlier; and the ledger was private. The exams then ran under the same commitment, and each decision follows mechanically from the transcripts. The ledger shows both, identical, reveal entries. The verifier's checks do not test that a reveal follows the close of the exams, so a passing verification does not attest it. Our runners did not refuse an early reveal; the released code will, and the committed runners stay unchanged so that their hashes verify.

\subsection{Study 5: an externally built family (E25)}
\label{app:study5}
 The generator is Google DeepMind's \texttt{mathematics\_dataset} 1.0.1 \citep{saxton2019analysing} (Apache-2.0), installed from PyPI and called unmodified in a pinned environment. A variant is a (module, regime, seed) triple; only modules whose draws do not depend on earlier draws were eligible (35 training modules in the hardest regime and 14 extrapolation modules). The selection rule, written before any model call on the family, ran 16 public-seed questions per eligible module through Haiku without a skill and kept the modules with a pass rate in $[0.30,0.60]$, at most one per area, closest to $0.45$ first. Two qualified, from different areas: primality testing (7 of 16) and division of large numbers (9 of 16). A variant's module is chosen by its seed with probability $1/2$ each. The model must put its answer inside \texttt{<answer>} tags; a pass is an exact match after normalisation, or equality as exact rationals.

The three published skills, \texttt{Math.Skill}, \texttt{math-teacher} and \texttt{math-help}, come from three public repositories, are MIT-licensed, and were used byte for byte at the commits pinned in the release, which names the repositories. We wrote the placebo (general work habits, no mathematical method) and the harmful skill (a primality screen that divides only by 2, 3, 5, 7 and 11, with composite numbers given as prime examples, and division answers as six-significant-figure decimals where the generator asks for exact fractions), both matched in length. The paired calibration ran 20 public-seed variants per skill with Haiku; every calibrated lift was negative (from $-0.05$ for the placebo to $-0.30$ for the harmful skill), so the power plan found no cap that reaches its target and the rule fixed in advance set $N_{\max}=600$. The Nemotron study used the same plan and the same exam order, levels and seed derivation, in its own registry history (drand rounds \NLsFiveRound{} and \NLnFiveRound). Nemotron ran with a 16{,}000-token limit and reached it in \NLnFiveCutPct\% of its answers over the study; Haiku reached its limit in \NLsFiveCutPct\% of its answers, most often in the \texttt{Math.Skill} exam with the skill (\NLsFiveSkillCutWith{} of \NLsFiveSkillRun). Four pairs of the Nemotron harmful exam were excluded for API failures after retries, under the committed rule.

\emph{Deviations.} (i) \emph{Grader.} The grader frozen before calibration (v1) failed mathematically correct answers whose terms were in another order and multiple-choice answers that gave the right letter with its value. We changed it (v2: SymPy equality for arithmetic expressions in integers and square roots, and the leading letter for multiple choice) after the no-skill calibration and before any skill was run, so the change cannot favour a skill, and re-graded the calibration offline with no new calls; both versions are in the release. Module selection used the v2 rates. (ii) \emph{Restarts.} The first Haiku run stopped with HTTP 425 because the committed beacon round had not yet been published, before any variant was drawn. The first Nemotron run failed on its first chunk because its transcript folder did not exist; nothing was written or logged, and the rerun started exam 1 from pair 0 with the same committed seeds, as for studies 4bN and 4cN. (iii) \emph{Labels.} The docstring of the committed runner was carried over from study 4 and names the old commands and budget; the values that apply are in the code and the committed parameters. Every decision recomputes from the ledger and transcripts with the released verifier.

%% file: tables/t_null_stress.tex
\begin{tabular}{lrrrrr}
\toprule
Departure from the design & Grid & PrPl & mSPRT & SPRT & Fixed-$n$ \\
\midrule
None (design of Section~\ref{sec:setup}) & 2.9 & 1.5 & 1.3 & 4.4 & 5.1 \\
Exam-level coin, $D_i\equiv\pm1$ & 49.4 & 49.4 & 49.4 & 49.4 & 0.0 \\
Exam-level coin, lift $\pm0.05$ & 22.2 & 26.0 & 23.6 & 20.2 & 17.8 \\
Exam-level coin, lift $\pm0.10$ & 49.3 & 49.5 & 49.5 & 41.7 & 40.1 \\
Model change mid-exam, both arms & 2.5 & 1.3 & 1.2 & 4.3 & 4.9 \\
Model change mid-exam, skill arm only & 15.5 & 14.4 & 17.4 & 7.7 & 7.3 \\
Shared effect, blocks of 5, sd 0.5 & 3.6 & 2.1 & 2.0 & 5.5 & 6.1 \\
Shared effect, blocks of 10, sd 0.5 & 5.1 & 3.1 & 3.0 & 7.0 & 6.9 \\
Shared effect, blocks of 10, sd 1.0 & 11.5 & 7.6 & 8.4 & 12.2 & 10.7 \\
Repeated seeds, 1\% & 2.9 & 1.6 & 1.4 & 4.5 & 5.0 \\
Repeated seeds, 5\% & 3.5 & 2.1 & 1.9 & 5.4 & 5.7 \\
Repeated seeds, 20\% & 7.4 & 5.2 & 4.7 & 9.7 & 9.4 \\
Excluding skill-arm losses, 1\% & 3.0 & 1.8 & 1.6 & 4.4 & 5.1 \\
Excluding skill-arm losses, 5\% & 3.7 & 2.4 & 2.1 & 5.9 & 6.4 \\
Excluding skill-arm losses, 20\% & 9.9 & 10.3 & 8.0 & 14.3 & 13.5 \\
\bottomrule
\end{tabular}

%% file: tables/t_scenarios.tex
\begin{tabular}{llrrrr}
\toprule
Scenario & Gate & Admitted (\%) & False share (\%) & Runs/candidate & Runs/true admission \\
\midrule
Curated-like & Fixed-$n$ $t$ & 86.6 & 0 & 274 & 317 \\
 & CI re-checked & 96.0 & 0 & 158 & 165 \\
 & SPRT & 86.8 & 0 & 127 & 147 \\
 & Betting + margin & 91.9 & 0 & 238 & 259 \\
 & Betting + e-LOND & 83.5 & 0 & 475 & 569 \\
\addlinespace
Self-generated-like & Fixed-$n$ $t$ & 3.1 & 8 & 274 & 9{,}448 \\
 & CI re-checked & 7.5 & 19 & 1131 & 18{,}548 \\
 & SPRT & 3.4 & 10 & 70 & 2{,}329 \\
 & Betting + margin & 3.3 & 5 & 250 & 7{,}934 \\
 & Betting + e-LOND & 0.3 & 0 & 274 & 84{,}454 \\
\addlinespace
SWE-Skills-like & Fixed-$n$ $t$ & 9.9 & 52 & 274 & 5{,}768 \\
 & CI re-checked & 23.8 & 75 & 977 & 16{,}217 \\
 & SPRT & 9.4 & 48 & 159 & 3{,}269 \\
 & Betting + margin & 8.5 & 35 & 709 & 12{,}956 \\
 & Betting + e-LOND & 3.2 & 0 & 770 & 23{,}702 \\
\addlinespace
WebDev-like & Fixed-$n$ $t$ & 14.3 & 3 & 274 & 1{,}978 \\
 & CI re-checked & 28.1 & 8 & 947 & 3{,}662 \\
 & SPRT & 14.8 & 3 & 122 & 851 \\
 & Betting + margin & 16.0 & 2 & 444 & 2{,}825 \\
 & Betting + e-LOND & 5.8 & 0 & 544 & 9{,}384 \\
\bottomrule
\end{tabular}

%% file: tables/t_leakage.tex
\begin{tabular}{lrrr}
\toprule
Skill (table covers 5\% of the space) & Literal & Normalized & Normalized + decoded \\
\midrule
Plain table & 100 & 100 & 100 \\
Spaced characters & 10 & 100 & 100 \\
Split string literals & 24 & 100 & 100 \\
Base64 blob & 0 & 0 & 100 \\
Hex blob & 0 & 0 & 100 \\
Generator code copied & 0 & 0 & 0 \\
Clean skills (false-positive rate) & 0 & 0 & 0 \\
\bottomrule
\end{tabular}

%% file: tables/t_verification.tex
\begin{tabular}{rrrrr}
\toprule
Variants $N$ & Transcripts (MB) & Proctor bookkeeping (s) & Verify (s) & Log bytes \\
\midrule
50 & 4 & 0.03 & 0.00 & 936 \\
100 & 8 & 0.05 & 0.01 & 936 \\
200 & 16 & 0.10 & 0.02 & 936 \\
400 & 32 & 0.21 & 0.04 & 936 \\
600 & 48 & 0.32 & 0.05 & 936 \\
\bottomrule
\end{tabular}

%% file: sections/appendix_theory.tex
\section{Proofs for Section~\ref{sec:theory}}\label{app:theory}

\paragraph{Proposition~\ref{thm:lineage}.}
\emph{(a)} Let $N_k$ be the event that exam $k$ does not admit. Given $\mathcal H_{K-1}$ on the event $N_1\cap\dots\cap N_{K-1}$, exam $K$ is valid at level $a_K$ by Definition~\ref{def:condnull}. Hence $\Pr(N_1\cap\dots\cap N_K)=\mathbb E[\mathbf 1\{N_1\cap\dots\cap N_{K-1}\}\Pr(N_K\mid\mathcal H_{K-1})]\ge(1-a_K)\Pr(N_1\cap\dots\cap N_{K-1})$. Induction gives $\Pr(\bigcap_{k\le K}N_k)\ge\prod_{k\le K}(1-a_k)$; let $K\to\infty$, and use $1-\prod_k(1-a_k)\le\sum_ka_k$.
\emph{(b)} Each exam's rule is fixed in advance, its variants are drawn independently of earlier exams, and the version's score law $Q$ does not change, so the outcome of exam $k$ is a fixed function of its own variants and the events $N_k$ are independent with $\Pr(N_k)=1-\sigma_k(Q)$, and $\Pr(\text{ever admitted})=1-\prod_k(1-\sigma_k(Q))$. Control at $\alpha$ requires $\prod_k(1-\sigma_k(Q))\ge1-\alpha$, that is, $\sum_k-\log(1-\sigma_k(Q))\le\bar\alpha$, and $-\log(1-s)\ge s$. With $\sigma_k(Q)\ge\sigma$ for all $k$ the error after $K$ attempts is at least $1-(1-\sigma)^K$.

\paragraph{Proposition~\ref{thm:onexam}.}
\emph{(a)} If $\mathbb E_P[\tau]=\infty$ there is nothing to prove. Take any null $Q$ with $\mathrm{KL}(P\|Q)<\infty$ and let $A$ be the admission event. Under $Q$ the exam may never stop, so we work with the bounded stopping time $\tau_n=\tau\wedge n$ and the event $A_n=A\cap\{\tau\le n\}$, which is $\mathcal F_{\tau_n}$-measurable. The change-of-measure inequality \citep[Lemma~1]{kaufmann2016complexity} gives $\mathbb E_P[\tau_n]\,\mathrm{KL}(P\|Q)\ge\mathrm{kl}(P(A_n),Q(A_n))$, where $\mathrm{kl}$ is the Bernoulli divergence. Write $p_n=P(A_n)$ and $q_n=Q(A_n)\le s$. Then $\mathrm{kl}(p_n,q_n)=p_n\log(1/q_n)+(1-p_n)\log\{1/(1-q_n)\}-h(p_n)\ge p_n\log(1/s)-\log2$, since the middle term is nonnegative and the binary entropy satisfies $h\le\log2$. Let $n\to\infty$: $\mathbb E_P[\tau_n]\uparrow\mathbb E_P[\tau]$ and $p_n\uparrow P(A)\ge1-\beta$. Divide, and take the infimum over $Q$.
\emph{(b)} Validity is Ville's inequality, as in Proposition~\ref{prop:valid}; the factors $1+\lambda(X_i-m_0)$ are nonnegative for every $\lambda\le1/m_0$, the boundary included. The increments $Y_i=\log(1+\lambda(X_i-m_0))$ are i.i.d.\ with $Y_i\le\log(1+\lambda(1-m_0))\le\log(1/m_0)$ and mean $g_P(\lambda)>0$; finiteness of $g_P(\lambda)$ makes them integrable even at $\lambda=1/m_0$, where they are not bounded below. Write $S_n=\sum_{i\le n}Y_i$. Wald's identity for the bounded stopping time $\tau\wedge n$ gives $g_P(\lambda)\,\mathbb E[\tau\wedge n]=\mathbb E[S_{\tau\wedge n}]$. Before $\tau$ the walk is below $\log(1/a)$, and the step that crosses adds at most $\log(1/m_0)$, so $S_{\tau\wedge n}<\log(1/a)+\log(1/m_0)$ (the overshoot bound). Hence $g_P(\lambda)\,\mathbb E[\tau\wedge n]<\log(1/a)+\log(1/m_0)$ for every $n$, and monotone convergence gives the stated bound for $\mathbb E[\tau]$. This covers the boundary case $\lambda^*=1/m_0$ of Theorem~\ref{thm:price}(b) and Eq.~\eqref{eq:sandwich}: if $\lambda^*=1/m_0$ maximizes $g_P$, then $g_P(1/m_0)=K_{\inf}$ is finite and positive, which forces $P(X=0)=0$. Power one follows from the strong law.
\emph{(c)} \emph{Weak duality.} Let $Q$ be a null law with $P\ll Q$ and let $0\le\lambda\le1/m_0$. By Jensen's inequality,
\begin{align*}
g_P(\lambda)-\mathrm{KL}(P\|Q)&=\mathbb E_P\log\Big\{(1+\lambda(X-m_0))\frac{dQ}{dP}\Big\}\le\log\mathbb E_Q[1+\lambda(X-m_0)]\\
&\le\log\{1+\lambda(\mathbb E_QX-m_0)\}\le0.
\end{align*}
\emph{Attainment.} The function $g_P$ is concave on $[0,1/m_0]$; let $\lambda^*$ maximize it. It is positive because $g_P'(0)=\mathbb E_PX-m_0>0$. Define $Q^*(x)=P(x)/(1+\lambda^*(x-m_0))$ on the support of $P$ and put the leftover mass $1-Q^*(\operatorname{supp}P)$ at $x=0$. Since $1/(1+\lambda u)=1-\lambda u/(1+\lambda u)$, the total mass of the first part is $1-\lambda^*\mathbb E_P[(X-m_0)/(1+\lambda^*(X-m_0))]=1-\lambda^*g_P'(\lambda^*)$.
\begin{itemize}
\item If $\lambda^*<1/m_0$, then $g_P'(\lambda^*)=0$. The first part has mass $1$, there is no leftover, and $\mathbb E_{Q^*}[X-m_0]=g_P'(\lambda^*)=0$.
\item If $\lambda^*=1/m_0$, then $g_P$ is finite only if $P(X=0)=0$, and $1+\lambda^*(x-m_0)=x/m_0$. Optimality at the right endpoint gives $g_P'(1/m_0)\ge0$, so the leftover mass $1-m_0\mathbb E_P[1/X]$ is nonnegative. The mean is $\mathbb E_{Q^*}[X-m_0]=(m_0-m_0^2\mathbb E_P[1/X])-m_0(1-m_0\mathbb E_P[1/X])=0$.
\end{itemize}
In both cases $Q^*$ is a null law with $\mathrm{KL}(P\|Q^*)=\mathbb E_P\log\{P(X)/Q^*(X)\}=g_P(\lambda^*)$. Combined with weak duality this gives $\max_\lambda g_P=K_{\inf}(P,m_0)$, attained at $Q^*$. Concavity makes the supremum over $[0,1/m_0)$ equal to the maximum over $[0,1/m_0]$. For laws with infinite support the same argument applies with sums replaced by integrals.

\paragraph{Theorem~\ref{thm:price}.}
\emph{(a)} Take the null law $Q^*$ of Proposition~\ref{thm:onexam}(c), for which $\mathrm{KL}(P\|Q^*)=K_{\inf}<\infty$, and let $A_k$ be the event that exam $k$ admits, so $Q^*(A_k)=\bar\alpha q_k$. Proposition~\ref{thm:lineage}(b) gives $\sum_kq_k\le1$. Follow the proof of Proposition~\ref{thm:onexam}(a) with $Q=Q^*$, but keep the binary entropy: $\mathbb E_P[\tau_{k,n}]K_{\inf}\ge\mathrm{kl}(p_n,s_n)\ge p_n\log(1/(\bar\alpha q_k))-h(p_n)$, where $p_n=P(A_k\cap\{\tau_k\le n\})$ and $s_n\le\bar\alpha q_k$. Let $n\to\infty$. Then $p_n\uparrow p=P(A_k)\ge1-\beta\ge1/2$, $h$ is continuous, and $h(p)\le h(1-\beta)=h(\beta)$ because $h$ decreases on $[1/2,1]$. So $\mathbb E_P[\tau_k]K_{\inf}\ge(1-\beta)\log\{1/(\bar\alpha q_k)\}-h(\beta)$. Exam $k$ is fixed in advance and runs on variants revealed after $\{J=k\}$ is determined, which are i.i.d.\ $P$ given that event and the earlier history, so $\mathbb E[\tau_J\mid J=k]=\mathbb E_P[\tau_k]$. Average over $k\sim\pi$, the marginal law of $J$, and write $\sum_k\pi_k\log(1/q_k)=H(\pi)+\mathrm{KL}(\pi\|q)$. Finally $\mathrm{KL}(\pi\|q)\ge-\log\sum_kq_k\ge0$ by Gibbs' inequality for a subprobability.
\emph{(b)} The gate is lineage-valid by Proposition~\ref{thm:lineage}(a), since $\sum_k\alpha q_k\le\alpha$. Proposition~\ref{thm:onexam}(b) at level $\alpha q_k$ gives $\mathbb E[\tau_k]\le\{\log(1/\alpha)+\log(1/q_k)+\log(1/m_0)\}/g_P(\lambda)$. As in part (a), $\mathbb E[\tau_J\mid J=k]=\mathbb E_P[\tau_k]$. Average over $\pi$, and use $\sum_k\pi_k\log(1/q_k)=H(\pi)+\mathrm{KL}(\pi\|q)$.

\paragraph{Eq.~\eqref{eq:sandwich}.} The right inequality is part (b) with $\lambda=\lambda^*$, since $g_P(\lambda^*)=K_{\inf}$ by Proposition~\ref{thm:onexam}(c). For the left one, the exams have power one by Proposition~\ref{thm:onexam}(b), so $p=1$ and $h(p)=0$ in the proof of part (a). Exam $k$ is valid at level $\alpha q_k$, so $Q^*(A_k)\le\alpha q_k$, and the proof of part (a) gives $\mathbb E_P[\tau_k]K_{\inf}\ge\log\{1/Q^*(A_k)\}\ge\log\{1/(\alpha q_k)\}$. Average over $\pi$.

\paragraph{Corollary~\ref{cor:universal}.}
If $\log(1/q_k)\le\log k+\log\log k$ for all large $k$, then $q_k\ge1/(k\log k)$ for all large $k$, and $\sum_kq_k=\infty$ because $\sum_k1/(k\log k)$ diverges. For the second claim, $c=\sum_k1/\{(k+1)\log^2(k+1)\}$ is finite, and $q_k=c^{-1}/\{(k+1)\log^2(k+1)\}$ gives $\log(1/q_k)=\log(k+1)+2\log\log(k+1)+\log c$.

\paragraph{Proposition~\ref{prop:minimax}.}
\emph{(a)} Let $q$ be the gate's schedule under $Q^*$, as in Theorem~\ref{thm:price}(a), with total mass $t=\sum_kq_k\le1$ and normalization $\tilde q=q/t$. Then $\mathrm{KL}(\pi\|q)=\mathrm{KL}(\pi\|\tilde q)+\log(1/t)\ge\mathrm{KL}(\pi\|\tilde q)$. Theorem~\ref{thm:price}(a) gives $K_{\inf}C(\pi)-(1-\beta)\{\log(1/\bar\alpha)+H(\pi)\}\ge(1-\beta)\mathrm{KL}(\pi\|\tilde q)-h(\beta)$, and $\sup_{\pi\in\Pi}\mathrm{KL}(\pi\|\tilde q)\ge R(\Pi)$.
\emph{(b)} Take the supremum over $\pi$ of the right inequality of Eq.~\eqref{eq:sandwich}.
\emph{(c)} For any $q$, expanding $\log(\pi_{jk}/q_k)=\log(\pi_{jk}/\bar\pi_{w,k})+\log(\bar\pi_{w,k}/q_k)$ gives
$\sup_{\pi\in\Pi}\mathrm{KL}(\pi\|q)\ge\sum_jw_j\mathrm{KL}(\pi_j\|q)=\sum_jw_j\mathrm{KL}(\pi_j\|\bar\pi_w)+\mathrm{KL}(\bar\pi_w\|q)\ge\sum_jw_j\mathrm{KL}(\pi_j\|\bar\pi_w)$.
Take the infimum over $q$.
\emph{(d)} Since $\sum_{k\le N}q_k\le1$, some $k\le N$ has $q_k\le1/N$, and the point mass at $k$ has $\mathrm{KL}=\log(1/q_k)\ge\log N$. For the uniform $u$, $\mathrm{KL}(\pi\|u)=\log N-H(\pi)\le\log N$. If $\pi$ is non-increasing, $k\pi_k\le\sum_{j\le k}\pi_j\le1$, so $\log\{\pi_k/q_k\}=\log(k\pi_kH_N)\le\log H_N$ for $q_k=1/(kH_N)$.
\emph{Certified interval in E12.} Let $U_m$ be uniform on $\{1,\dots,m\}$. Every $\pi\in\Pi_N^{\downarrow}$ is the mixture $\sum_{m\le N}m(\pi_m-\pi_{m+1})U_m$, with $\pi_{N+1}=0$; the weights are nonnegative and sum to $\sum_k\pi_k=1$. Since $\mathrm{KL}(\pi\|q)$ is convex in $\pi$, $\sup_{\pi\in\Pi_N^\downarrow}\mathrm{KL}(\pi\|q)=\max_m\mathrm{KL}(U_m\|q)$ for every $q$. Hence, for any weights $w$ and any $q$, part (c) gives $\sum_mw_m\mathrm{KL}(U_m\|\bar U_w)\le R(\Pi_N^\downarrow)\le\max_m\mathrm{KL}(U_m\|q)$. E12 evaluates both ends at the prior and output law of the Blahut-Arimoto iteration, so the interval does not depend on its convergence.

\paragraph{Theorem~\ref{thm:learned}.}
\emph{(a)} The counts, and hence $\hat q^{(i)}$, are fixed when lineage $i$ opens. They are part of the history $\mathcal H_0$ of lineage $i$, whatever produced them. Also $\sum_k\hat q^{(i)}_k=(1-\epsilon)+\epsilon=1$. The proof of Proposition~\ref{thm:lineage}(a), conditional on $\mathcal H_0$, gives $\Pr(\text{ever admitted}\mid\mathcal H_0)\le\sum_k\alpha\hat q^{(i)}_k=\alpha$.
\emph{(b)} The counted lineages are chosen without regard to their attempt indices, so the honest counts at $k$ are $b_k\sim\mathrm{Bin}(m_h,\pi_k)$. (If lineages were counted because they had closed, short lineages would be over-represented and this step would fail.) Then $c_k\ge b_k$ and $m\le m_h+r$, so $\hat q^{(i)}_k\ge(1-\epsilon)(b_k+1)/(m_h+r+N)$ on every sample. Hence
$\mathbb E\log(\pi_k/\hat q^{(i)}_k)\le\log\frac1{1-\epsilon}+\log\pi_k+\log(m_h+r+N)+\mathbb E\log\frac1{b_k+1}$.
By Jensen's inequality, $\mathbb E\log\{1/(b_k+1)\}\le\log\mathbb E\{1/(b_k+1)\}$, and for $b\sim\mathrm{Bin}(m,p)$,
$\mathbb E\{1/(b+1)\}=\frac1{(m+1)p}\sum_{c=0}^m\binom{m+1}{c+1}p^{c+1}(1-p)^{m-c}=\frac{1-(1-p)^{m+1}}{(m+1)p}\le\frac1{(m+1)p}$.
So $\mathbb E\log(\pi_k/\hat q^{(i)}_k)\le\log\frac1{1-\epsilon}+\log\frac{m_h+r+N}{m_h+1}$. Average over $k\sim\pi$. The pathwise bound follows from $\hat q^{(i)}_k\ge\epsilon/N$.
\emph{(c)} For $i>D+1$, lineage $i$ counts the $m_h=i-1-D$ lineages $1,\dots,i-1-D$. All are closed when lineage $i$ opens, by assumption, and the set is fixed by the indices alone, so their attempt indices are i.i.d.\ $\pi$ and (b) applies with $r=0$. For $i\le D+1$, bound the logarithm in (b) by $\log N$, its value at $m_h=0$. For $i>D+1$, use $\log(1+x)\le x$ and $m_h+1\ge i-D$: $\sum_{i=D+2}^n(N-1)/(i-D)\le(N-1)\sum_{j=2}^n1/j\le(N-1)\log n$. For the converse, let $J_1,\dots,J_n$ be i.i.d.\ $\pi$ and let $q^{(i)}$ depend on $J_1,\dots,J_{i-1}$; a rule with a delay is a special case. Then $M(j_1,\dots,j_n)=\prod_iq^{(i)}_{j_i}(j_1,\dots,j_{i-1})$ is a probability law on $\{1,\dots,N\}^n$, or a subprobability, which only increases the divergence. By the chain rule, $\sum_i\mathbb E\,\mathrm{KL}(\pi\|q^{(i)})=\mathrm{KL}(\pi^{\otimes n}\|M)$. Its supremum over $\pi\in\Pi_N$ is at least $R_n(N)=\inf_M\sup_\pi\mathrm{KL}(\pi^{\otimes n}\|M)$, whose expansion is that of \citet{xie1997minimax} \citep[see also][Theorem~2]{boucheron2009coding}. For the cost, apply the proof of Theorem~\ref{thm:price}(a) to lineage $i$ given $J_1,\dots,J_{i-1}$, with $\beta=0$, then sum and use the chain rule as above.

\emph{Capped exams.} Let $c_k$ count the counted lineages admitted at attempt $k$, and let $\tilde\pi$ be the law of the admitting attempt given admission. Lineages are independent and counted without regard to their attempt indices, so given their number the admitted honest counts are multinomial with probabilities $\tilde\pi$, and the proofs of (b) and (c) apply verbatim with $\tilde\pi$ for $\pi$ and $m_h$ the number of admitted honest lineages. For the total-variation bound, let $\hat J$ be the admitting attempt and $A$ the event that no useless version is admitted before attempt $J$ and the helpful exam at attempt $J\le N$ admits before the cap. Then $\Pr(A^c)\le\alpha+\beta_{\max}=\varepsilon$, by lineage control for the useless versions and the power assumption for exam $J$, and $\hat J=J$ on $A\subseteq\{\text{admitted}\}$. For any set $S$, $|\Pr(J\in S\mid A)-\Pr(J\in S)|\le\Pr(A^c)\le\varepsilon$. The law of $\hat J$ given admission mixes its law given $A$, with weight $\Pr(A)/\Pr(\text{admitted})$, and its law on the rest of the admission event, whose weight is at most $\varepsilon/(1-\varepsilon)$. Hence $\mathrm{TV}(\tilde\pi,\pi)\le\varepsilon+\varepsilon/(1-\varepsilon)$. Lineages are taken independent, as in part (b). No assumption on resubmission after a miss is needed; the cap of $N$ attempts is what makes a uniform $\beta_{\max}$ possible, since an exam truncated at $N_{\max}$ cannot admit at a level below $m_0^{N_{\max}}$.

\paragraph{Theorem~\ref{thm:adaptive}.} \emph{(a)} Since $\lambda_j\le\rho/m_0<1/m_0$, each factor $1+\lambda_j(X_i-m_0)\ge1-\rho>0$, and under Definition~\ref{def:condnull} each wealth $W_n(\lambda_j)=\prod_{i\le n}\{1+\lambda_j(X_i-m_0)\}$ is a nonnegative supermartingale with $W_0=1$. By monotone convergence so is $M_n=\sum_jw_jW_n(\lambda_j)$, with $M_0=\sum_{j\ge0}w_j=c_u\sum_{k\ge1}1/\{(k+1)\log^2(k+1)\}=1$. Ville's inequality gives $\Pr(\sup_nM_n\ge1/a)\le a$. Nothing in the exam depends on $P$. Power one follows from (b), since $\mathbb E_P[\tau]<\infty$.
\emph{(b)} The maximizer exists: $g_P$ is concave on $[0,1/m_0]$ and upper semicontinuous, since $\log\{1+\lambda(x-m_0)\}\le\log(1/m_0)$ allows the reverse Fatou lemma; its maximum is $K_{\inf}$ by Proposition~\ref{thm:onexam}(c), and $\lambda^*>0$ because $g_P'(0)=\mathbb E_PX-m_0>0$. Let $j=j^*$. Then $j\le\Lambda/\log(1/\rho)<j+1$, so $\rho^{j+1}<m_0\lambda^*\le\rho^{j}$, and $\rho\lambda^*\le\lambda_j\le\lambda^*$. Concavity with $g_P(0)=0$ gives $g_P(\lambda_j)\ge(\lambda_j/\lambda^*)g_P(\lambda^*)\ge\rho K_{\inf}$. Since $M_n\ge w_jW_n(\lambda_j)$, the exam stops no later than the single-bet exam with bet $\lambda_j$ at level $aw_j$, and Proposition~\ref{thm:onexam}(b) bounds the latter's expected stopping time by $\{\log(1/a)+\ell(j)+\log(1/m_0)\}/g_P(\lambda_j)$. Divide by $\rho K_{\inf}$, and use $1/\rho=1+1/L$. For the explicit form, $\log(1/\rho)=\log(1+1/L)\ge1/(L+1)$ gives $j^*\le(L+1)\Lambda$, so $j^*+2\le(L+1)(2+\Lambda)$. Since $(j+2)\log^2(j+2)$ increases in $j$, $\ell(j^*)=\log(1/c_u)+\log(j^*+2)+2\log\log(j^*+2)\le\log(1/c_u)+u+2\log u$. Expand $(1+1/L)L=L+1$.
\emph{(c)} Validity is Proposition~\ref{thm:lineage}(a) with (a). Apply (b) to exam $k$ with $L_k=\log(1/\alpha)+\log(1/q_k)\ge\log(1/\alpha)$, so $1+1/L_k\le1+1/\log(1/\alpha)$; this replacement multiplies a nonnegative term, since $m_0\le1$, $c_u<1$ and $u_k=\log\{(L_k+1)(2+\Lambda)\}\ge\log\{2(1+\log(1/\alpha))\}\ge1$ for $\alpha\le0.5$. Then average over $k\sim\pi$ as in Theorem~\ref{thm:price}, using $\mathbb E[\tau_J\mid J=k]=\mathbb E_P[\tau_k]$. The functions $L\mapsto\log(L+1)$ and $L\mapsto\log\log\{(L+1)(2+\Lambda)\}$ are concave and increasing (the second is the logarithm of a positive concave function), so Jensen's inequality bounds their averages by their values at $\sum_k\pi_kL_k=\bar L$. The lower bound is Theorem~\ref{thm:price}(a) with $\beta=0$.

\paragraph{Scope.} Theorem~\ref{thm:price} counts only the helpful version's exam, and assumes untruncated exams with the stated power. With $N_{\max}$ truncation and futility, $\beta>0$ enters through $(1-\beta)$, and the useless earlier versions also spend variants. The theorems treat the lineage component of the deployed level $\min(\alpha_t,\alpha\gamma^{\text{lin}}_k)$, not its interaction with e-LOND. The cost bounds use i.i.d.\ variants within an exam. The validity statements use only the conditional null.

%% file: sections/appendix.tex
\section{Proofs}
\label{app:proofs}

\paragraph{Proposition~\ref{prop:valid}.} The argument holds for any predictable bets $\lambda_i\in[0,1/m_0]$, which covers both caps of Eq.~\eqref{eq:wealth} and the aGRAPA bet with $\lambda_i\le c/m_0<1/m_0$; the grid bet is an average of such processes with fixed bets, and an average of nonnegative supermartingales with $K_0=1$ is one. Each factor $1+\lambda_i(X_i-m_0)$ is nonnegative because $X_i\ge0$ and $\lambda_i m_0\le1$. Since $\lambda_i$ is $\mathcal F_{i-1}$-measurable and $\mathbb{E}[X_i\mid\mathcal F_{i-1}]\le m_0$, $\mathbb{E}[K_i\mid\mathcal F_{i-1}]=K_{i-1}(1+\lambda_i(\mathbb{E}[X_i\mid\mathcal F_{i-1}]-m_0))\le K_{i-1}$, so $(K_n)$ is a nonnegative supermartingale with $K_0=1$. Ville's inequality gives $\Pr(\sup_nK_n\ge1/\alpha)\le\alpha$. Admission requires $K_n\ge1/\alpha$ at some $n$, whatever else stops the exam, so $\Pr(\text{admit})\le\alpha$. For any stopping time $\tau$, optional stopping for nonnegative supermartingales gives $\mathbb{E}[K_\tau]\le1$. The futility process uses its own predictable bets and only ends exams, so it cannot add admissions. Nothing here requires the $X_i$ to be identically distributed.

\paragraph{Remark~\ref{prop:ebh}.} At the stopping time $\tau$, $K_{\tau-1}<1/\alpha$ and $K_\tau=K_{\tau-1}(1+\lambda_\tau(X_\tau-m_0))\le K_{\tau-1}(1+\bar\lambda(1-m_0))<(1+\bar\lambda(1-m_0))/\alpha$, because $X_\tau\le1$; exams that never cross end below $1/\alpha$. e-BH admits the $k$ largest e-values only if $e_{(k)}\ge m/(\alpha k)$, which needs $(1+\bar\lambda(1-m_0))/\alpha>m/(\alpha k)$, that is $k>m/(1+\bar\lambda(1-m_0))$, and the $k$ largest must all have crossed. At $m_0=1/2$: $\bar\lambda=1/2$ gives $k>0.8m$, $\bar\lambda=3/4$ gives $k>0.727m$, and $\bar\lambda=1$ (aGRAPA, $c=1/2$) gives $k>2m/3$. Since $\bar\lambda<1/m_0$, $1+\bar\lambda(1-m_0)<1/m_0$, so $k>m_0m$ for every admissible bet.

\paragraph{Proposition~\ref{prop:lineage}.} The level of the $k$-th exam is fixed when it starts. Let $\mathcal H_{k-1}$ be the history before exam $k$, including version $k$ and its comparator, and let $\nu_k\in\{0,1\}$ indicate that version $k$ is null; $\nu_k$ is $\mathcal H_{k-1}$-measurable. By Proposition~\ref{prop:valid}, on $\{\nu_k=1\}$ exam $k$ admits with probability at most its level given $\mathcal H_{k-1}$, which is at most $\alpha\gamma^{\text{lin}}_k$. A union bound over the lineage gives $\Pr(\text{some null version admitted})\le\sum_k\mathbb E[\nu_k\Pr(\text{admit}_k\mid\mathcal H_{k-1})]\le\sum_k\alpha\gamma^{\text{lin}}_k\le\alpha$, whatever the non-null versions do. Each exam's level is at most its e-LOND level, and e-LOND's FDR bound holds for any predictable levels not exceeding $\alpha\gamma_t(R_{t-1}+1)$ \citep{xu2024online}. In one line: let $\alpha'_t\le\alpha\gamma_t(R_{t-1}+1)$ be fixed when exam $t$ starts, let $e_t$ be its wealth at the stopping time, so that $\mathbb E[e_t\mid\mathcal F_{t-1}]\le1$ for a null exam by optional stopping, where $\mathcal F_{t-1}$ is the registry's history when exam $t$ starts and the null holds conditionally on it (Section~\ref{sec:setup}), and let $R\ge R_{t-1}+1$ be the total number of admissions when exam $t$ admits. Then $\mathbf 1\{\text{admit}_t\}/R\le\alpha'_t e_t/(R_{t-1}+1)\le\alpha\gamma_t e_t$, and summing over null exams gives $\mathrm{FDR}\le\alpha\sum_t\gamma_t\le\alpha$. Using fewer admissions than $R_{t-1}$, for example only those known when a window starts, lowers $\alpha'_t$ and keeps the bound.

\section{Implementation details}
\label{app:impl}
\paragraph{Credential construction.} Each transcript is serialized as canonical JSON \citep{rundgren2020jcs} and hashed into one Merkle tree per arm with RFC~6962 domain separation \citep{laurie2013ct,merkle1988digital}. The credential body binds the agent identity, model and harness versions, $H(s)$, the commitment and reveal, $\theta$, both arms' roots, the leakage-report hash, and the statistics $(N,\hat\Delta,\mathrm{LCB},\text{decision})$. Its digest $\chi$ is signed and logged.

Predictable plug-in bets follow \citet{waudbysmith2024betting}: $\hat\mu_t=(1/2+\sum_{i\le t}X_i)/(t+1)$, $\hat v_t=(1/4+\sum_{i\le t}(X_i-\hat\mu_i)^2)/(t+1)$, and $\lambda_t$ in Eq.~\eqref{eq:wealth} uses $\hat v_{t-1}$. The aGRAPA bet is $\lambda_t=\min\{c/m_0,\max\{0,(\hat\mu_{t-1}-m_0)/(\hat v_{t-1}+(\hat\mu_{t-1}-m_0)^2)\}\}$. The lower confidence bound inverts the PrPl wealth over a grid of 2{,}001 values of $m$ with a running intersection; with the PrPl bet an exam admits exactly when its bound exceeds $\delta$, and with the grid bet the bound is reported but does not decide. e-LOND uses $\gamma_t=t^{-1.6}/\zeta(1.6)$, or $\gamma_t=1/T$ over a planned horizon $T$. Signatures are Ed25519. The reference implementation passes 79 tests, including Ville's bound at the null boundary and under the worst-case two-point null, time-uniform coverage of the bound, a check that the fixed-$n$ sample size of Section~\ref{sec:eval} reaches its nominal power, and detection of tampered transcripts, skills, decisions, reveals, signatures and hidden early stopping.

\section{Credential fields}
\label{app:schema}
The credential body contains: version and status; agent identity, controller key, model and harness versions, configuration digest; skill digest (Merkle root over canonically ordered files); comparator kind and digest; commitment, registered parameters $\theta$ and their hash, registration, commitment and beacon heights, and the reveal $(\sigma_p,\rho)$; for each arm the Merkle root, leaf count and leaf hashes; the statistics $(N,\hat\Delta,\mathrm{LCB},\text{decision},\text{test})$; the security-scan and leakage-report hashes; and, optionally, an attestation quote over $H(R^{(0)}\|R^{(1)})$. The digest $\chi$ is computed over the body without the signature and log fields, which are attached afterwards.

\begin{algorithm}[H]
\caption{VerifyCredential$(\chi)$}\label{alg:verify}
\begin{algorithmic}[1]
\State fetch body $B$; check $H(\mathrm{JCS}(B))=\chi$, the signature, and that $\chi$ is logged and not revoked
\State check log order: registration of $(\mathrm{id}(a),H(s))$ at $b_0<$ commitment $c$ at $b_1<b_2<$ reveal at $b_3$
\State check $H(\sigma_p\|\rho\|\theta)=c$ and the skill digest; set $\sigma\gets H(\sigma_p\|\beta_{b_2})$
\State fetch both arms' transcripts; check their Merkle roots against $B$
\For{$i=1,\dots,N$}
\State check each transcript names variant $i$, attempt $\ell$ and digest $H(G(\sigma,i))$, with the stated model and harness
\State recompute every score from the recorded final state with the verifier; check it equals the recorded score
\EndFor
\State recompute $D_{1:N}$ and the wealth paths under $\theta$; check that no boundary was crossed before $N$, that $N=N_{\max}$ if no boundary was crossed, and that the decision and $\hat\Delta$ match $B$
\end{algorithmic}
\end{algorithm}

\section{Pre-registration plan for the peer-synthesis study}
\label{app:campus}
\paragraph{Question.} Do skills distilled by a mentor agent from the shared traces of a study group of $k$ agents pass the gate more often, and with larger certified lift, than skills an agent writes from its own traces?
\paragraph{Design.} Task classes with parametric practice and exam generators and deterministic verifiers (candidates: log forensics, data reconciliation, spreadsheet automation). Conditions: no skill (comparator), curated, self-generated, peer-synthesized with $k\in\{2,4,8\}$, and controls (length-matched irrelevant skills, span-shuffled skills, planted leaks). Each candidate gets a paired exam on a fresh examinee agent under the gate of Section~\ref{sec:gate}, with $\theta$ committed before the run. Primary outcome: certified admission and the lower confidence bound against no skill. Primary comparison: peer-synthesized against self-generated on matched classes. Contributor credit, reputation and version carry-over, which the original design included, are secondary and are not part of the confirmatory analysis.
\paragraph{Parameters to fix and hash before any run.} $\alpha$, $\alpha_{\mathrm f}$, $\delta$, $\delta_{\mathrm f}$, $r$, $N_{\max}$, the bet rule and cap, the e-LOND sequence, the lineage rule, leakage thresholds, models and harnesses with exact versions, and the analysis-code hash.

\section{External data}
\label{app:data}
We use two released files, unmodified; their SHA-256 hashes are in the code repository (\texttt{data/external/PROVENANCE.md}).
\begin{itemize}
\item SWE-Skills-Bench (MIT license), file \path{results/final_results/ALL_combined_cells.csv} at commit \texttt{4d3450d}, from\newline \url{https://github.com/ControllableModel/SWE-Skills-Bench}.
\item SkillsBench (AGPL-3.0), commit \texttt{3748490}, file\newline \path{website-data/results-registry.json}, from\newline {\small\url{https://github.com/benchflow-ai/skillsbench-trajectories}}.
\end{itemize}